\documentclass[review,3p,times]{elsarticle}

\usepackage{graphicx}
\usepackage{subfigure}
\usepackage{caption}

\usepackage{amssymb}
\usepackage{amsmath}
\usepackage{xfrac}
\usepackage{booktabs}
\usepackage{mathtools}

\usepackage{url}
\usepackage[dvipsnames,table]{xcolor}

\makeatletter
\newif\if@restonecol
\makeatother

\usepackage{bm}
\usepackage{multirow}
\usepackage{mathrsfs}

\usepackage[ruled,linesnumbered,lined]{algorithm2e}

\usepackage{boolexpr,pdftexcmds,trace}
\usepackage{array} 
\usepackage{pdflscape}

\newcommand{\R}{\ensuremath{\mathbb{R}}}

\newcommand{\fs}[2]{\fontsize{#1 pt}{#2 pt}\selectfont}

\allowdisplaybreaks

\newcommand{\CMMsha}{$\text{CMM}_\textit{sha}$\xspace}
\newcommand{\CMMsep}{$\text{CMM}_\textit{sep}$\xspace}

\begin{document}

\begin{frontmatter}



\title{Semi-Supervised Active Learning for Support Vector Machines: A Novel Approach that Exploits Structure Information in Data}
\author{Tobias Reitmaier}
\author{Adrian Calma}
\author{Bernhard Sick}

\address{Intelligent Embedded Systems Lab, University of Kassel\\
Wilhelmsh\"oher Allee 71 -- 73, 34121 Kassel, Germany\\ 
(e-mail: $\{\text{tobias.reitmaier,adrian.calma,bsick}\}\text{@uni-kassel.de}$)}

\begin{abstract}
In our today's information society more and more data emerges, e.g.~in social networks, technical applications, or business applications. Companies try to commercialize these data using data mining or machine learning methods. For this purpose, the data are categorized or classified, but often at high (monetary or temporal) costs. An effective approach to reduce these costs is to apply any kind of active learning (AL) methods, as AL controls the training process of a classifier by specific querying individual data points (samples), which are then labeled (e.g., provided with class memberships) by a domain expert. However, an analysis of current AL research shows that AL still has some shortcomings. In particular, the structure information given by the spatial pattern of the (un)labeled data in the input space of a classification model (e.g.,~cluster information), is used in an insufficient way. In addition, many existing AL techniques pay too little attention to their practical applicability. To meet these challenges, this article presents several techniques that together build a new approach for combining AL and semi-supervised learning (SSL) for support vector machines (SVM) in classification tasks. Structure information is captured by means of probabilistic models that are iteratively improved at runtime when label information becomes available. The probabilistic models are considered in a selection strategy based on distance, density, diversity, and distribution (4DS strategy) information for AL and in a kernel function (Responsibility Weighted Mahalanobis kernel) for SVM. The approach fuses generative and discriminative modeling techniques. With $20$ benchmark data sets and with the MNIST data set it is shown that our new solution yields significantly better results than state-of-the-art methods. 

\end{abstract}


\end{frontmatter}

\section{Introduction}
\label{sec:intro}
Machine learning is based on sample data. Sometimes, these data are labeled and, thus, models to solve a certain task (e.g., a classification or regression task) can be built using targets assigned to the input data of a classification or regression model. In other cases, data are unlabeled (e.g., for clustering) or only partially labeled. Correspondingly, we distinguish the areas of \textit{supervised}, \textit{unsupervised}, and \textit{semi-supervised} learning. In many application areas (e.g., industrial quality monitoring processes~\cite{Sic98}, intrusion detection in computer networks~\cite{HSS03}, speech recognition~\cite{FHYA12}, or drug discovery~\cite{MME14}) it is rather easy to collect unlabeled data, but quite difficult, time-consuming, or expensive to gather the corresponding targets. That is, labeling is in principal possible, but the costs are enormous. An effective approach to reduce these costs is to apply \textit{active learning} (AL) methods, as AL controls the training process by specific querying of individual samples (also called examples, data points, or observations), which are then labeled (e.g., provided with class memberships) by a domain expert. In this article, we focus on classification problems.

Pool-based AL typically assumes that at the beginning of a training process a large set $U$ of unlabeled samples and a small set $L$ of labeled samples are available. Then, AL iteratively increases the number of labeled samples ($\mathbf{x} \in L$) by asking the ``right'' questions with the aim to train a classifier with the highest possible generalization performance and, at the same time, the smallest possible number of labeled samples. Generally, a \textit{selection strategy} is used that selects informative samples from $U$ by considering the  current ``knowledge'' of the classifier that shall be trained actively. These samples are labeled by an oracle or a domain expert, added to the training set $L$, and the classifier is updated. The AL process stops as soon as a predefined stopping criterion is met.

An analysis of current research in the field of AL \cite{Cawlay11, GCDL11, Settles11} shows that AL still has some shortcomings. In particular, the structure information, which is given by the spatial arrangement of the (un)labeled data in the input space of a classifier, is used in an insufficient way. Furthermore, many existing AL techniques pay too little attention to their practical applicability, e.g., regarding the number of initially labeled samples or the number of adjustable parameters which should both be as low as possible. To meet these challenges we present in this article several techniques that build together a new approach that combines AL and semi-supervised learning (SSL) for support vector machines (SVM).

In general, machine learning techniques can be applied whenever ``patterns'' or ``regularities'' in a set of sample data can be recognized and, thus, be exploited. For a classification tasks this often means that the data build clusters of arbitrary shapes, whereby such \textit{structures} shall be revealed and modeled in order to consider them for the active sample selection and for the classifier training. Starting from this point of view we propose several techniques in this article to combine AL and SSL for SVM: First, we claim that in a real application an AL process has to start ``from scratch'' without any label information. Therefore, we use probabilistic mixture models (i.e., generative models) to capture the structure information of the unlabeled samples. These models are determined \textit{offline} (before the AL process starts) with the help of \textit{variational Bayesian inference} (VI) techniques in an unsupervised manner. Second, during the AL process, when more and more label information becomes available, we use these class labels to revise these density models. For this, we introduce a \textit{transductive learner} into the standard PAL process that adapts the mixture model with the help of local modifications such that model components preferably model clusters of samples that belong to the same class. Third, the data sets $\textit{U}_{\textit{i}}$ and $\textit{L}_{\textit{i}}$ in each iteration $i$ ($i>1$) of the AL process can be seen as a sparsely labeled data set that can be used to train a classifier in a semi-supervised manner. Based on the iteratively improved parametric density models, e.g., Gaussian mixture models in the case of a continuous (real-valued) input space, we derive a new data-dependent kernel, called responsibility weighted Mahalanobis (RWM) kernel. Basically, this kernel is based on Mahalanobis distances being part of the Gaussians but it reinforces the impact of model components from which any two samples that are compared are assumed to originate. Fourth, a selection strategy for AL has to fulfill several tasks, for example: At an early stage of the AL process, samples have to be chosen in all regions of the input space covered by data (\textit{exploration phase}). At a late stage of the AL process, a fine-tuning of the decision boundary of the classifier has to be realized by choosing samples close to the (current) decision boundary (\textit{exploitation phase}). Thus, ``asking the right question'' (i.e., choosing samples for a query) is a multi-faceted problem. To solve this problem we adopt a self-adaptive, multi-criteria strategy called 4DS that considers structure in the data for its sample selection. Fifth, many works in the field of AL assume that at the beginning of the AL process a relatively large number of labeled samples are available, but in a real application this is usually not true. In addition, these data have a great impact on the learning performance of the actively trained classifier. Therefore, we present an extended version of the standard PAL cycle that starts with an empty labeled training set $L_0$ and determines the first labeled set ($L_1$) within an \textit{initialization round} using structure information.

The innovative aspects of the work presented in this article are:
\begin{itemize}
	\item Our new approach starts with an empty initial training set. 
	This means that the ``knowledge'' of the actively trained classifier cannot be used for sample selection in the first query round. 
	Hence, a density based strategy is used to find informative samples.
	\item Data structure is captured with help of generative, probabilistic mixture models, that are initially trained with VI in an unsupervised fashion.
	These data models are adapted (revised) during the AL process with help of class information that becomes available. 
	Thus, we aim to model data clusters of samples that are assumed to belong to the same class.
	\item Structure information is considered (1) for the active sample selection by a self-adaptive, multi-criteria selection strategy (4DS) and (2) for SVM training (finding the support vectors) with help of a data-dependent (RWM) kernel.
	\item Bringing all together results in a practical, effective, and efficient approach that combines AL and SSL for SVM.
\end{itemize}

The AL approach presented here is based on a complex interplay of several building blocks presented in previous issues of this journal that are combined and evaluated here for the very first time (the selection strategy 4DS \cite{RS13}, the transductive learning scheme \cite{RCS14}, and the RWM kernel for SVM \cite{RS15}).

The remainder of this article is structured as follows: Section~\ref{sec:example} illustrates the potential of AL and the properties of our new approach. Section~\ref{sec:overview}  gives an overview of related work. Section~\ref{sec:foundations} sketches the probabilistic mixture model that we use to gather structure in data and defines the RWM kernel and the selection strategy 4DS to train an SVM in a semi-supervised active fashion. In addition, Section~\ref{sec:foundations} explains our extension of the standard PAL cycle that makes it possible to start the AL process ``from scratch'' (without any labeled samples) and to refine the mixture model as soon as label information becomes available. Results of a large number of simulation experiments are set out in Section~\ref{sec:results}. Finally, Section~\ref{sec:conclusion} summarizes the key findings and gives an outlook to our future work.

\section{Motivating Example}
\label{sec:example}
\begin{figure}[htbp!]
	\begin{center}
		\subfigure[Random sampling (Random).]{\label{fig:example_alA}\includegraphics[width=0.318\textwidth]{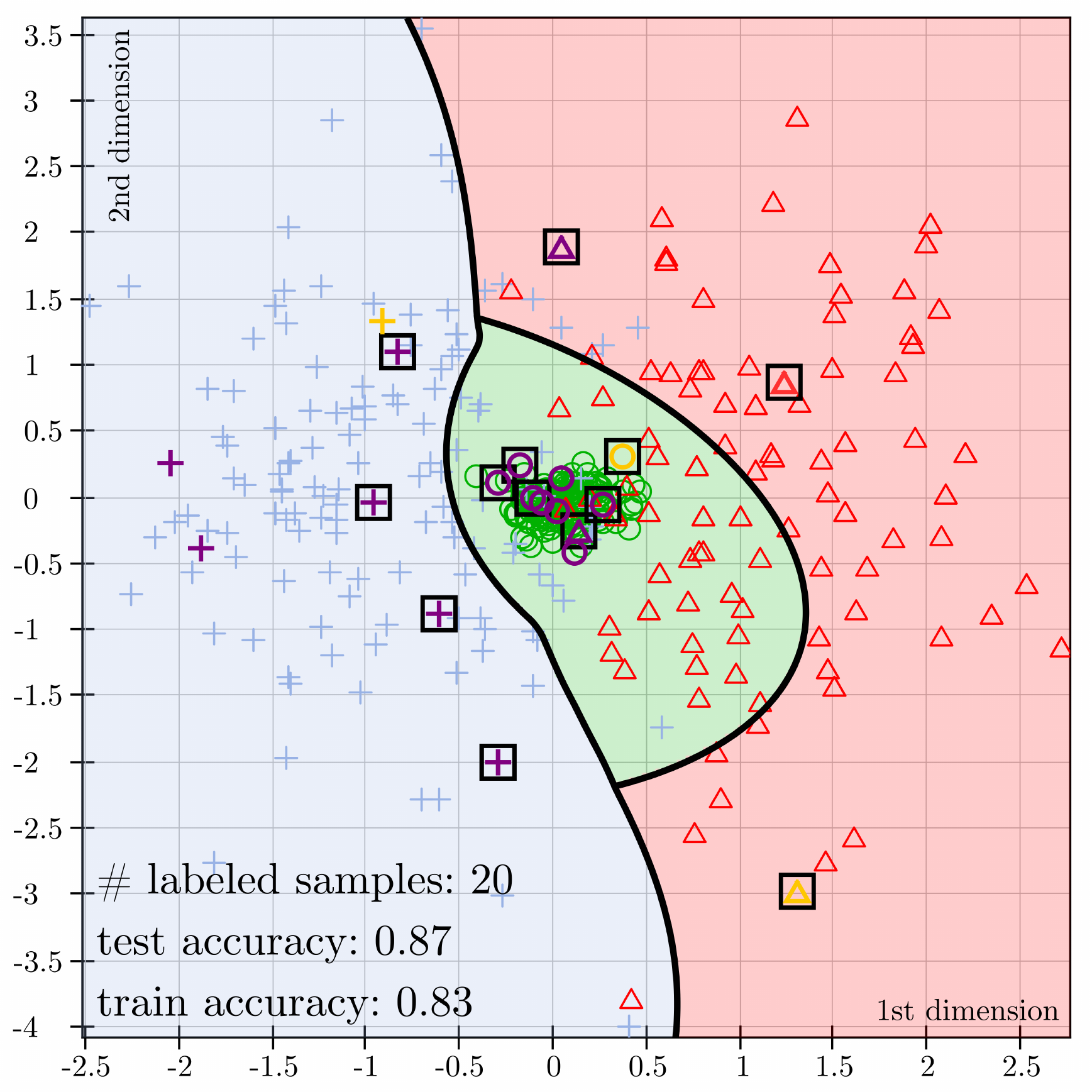}}\quad
		\subfigure[Uncertainty sampling (US).]{\label{fig:example_alB}\includegraphics[width=0.318\textwidth]{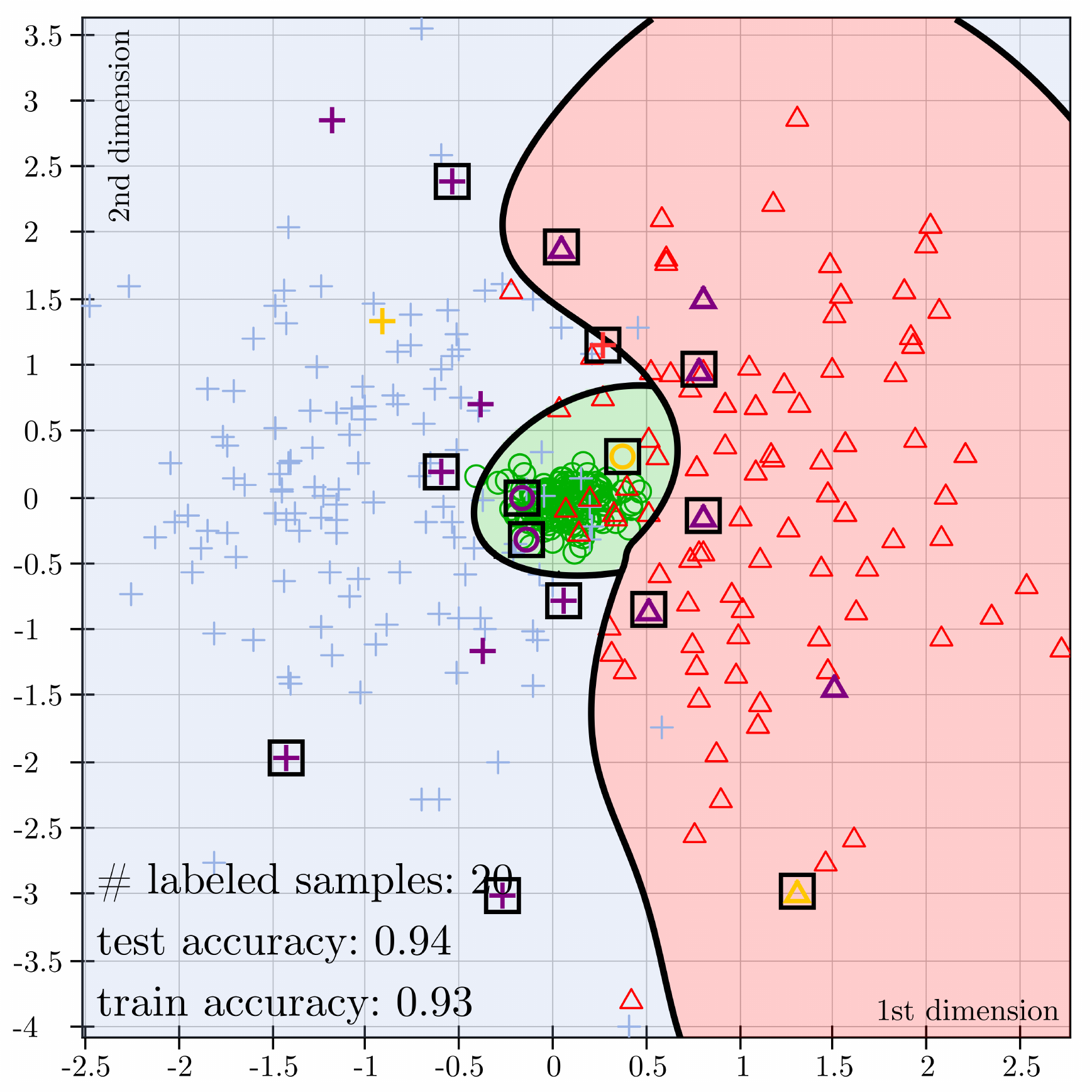}}\quad
		\subfigure[Using structure information (4DS and RWM kernel).]{\label{fig:example_alC}\includegraphics[width=0.318\textwidth]{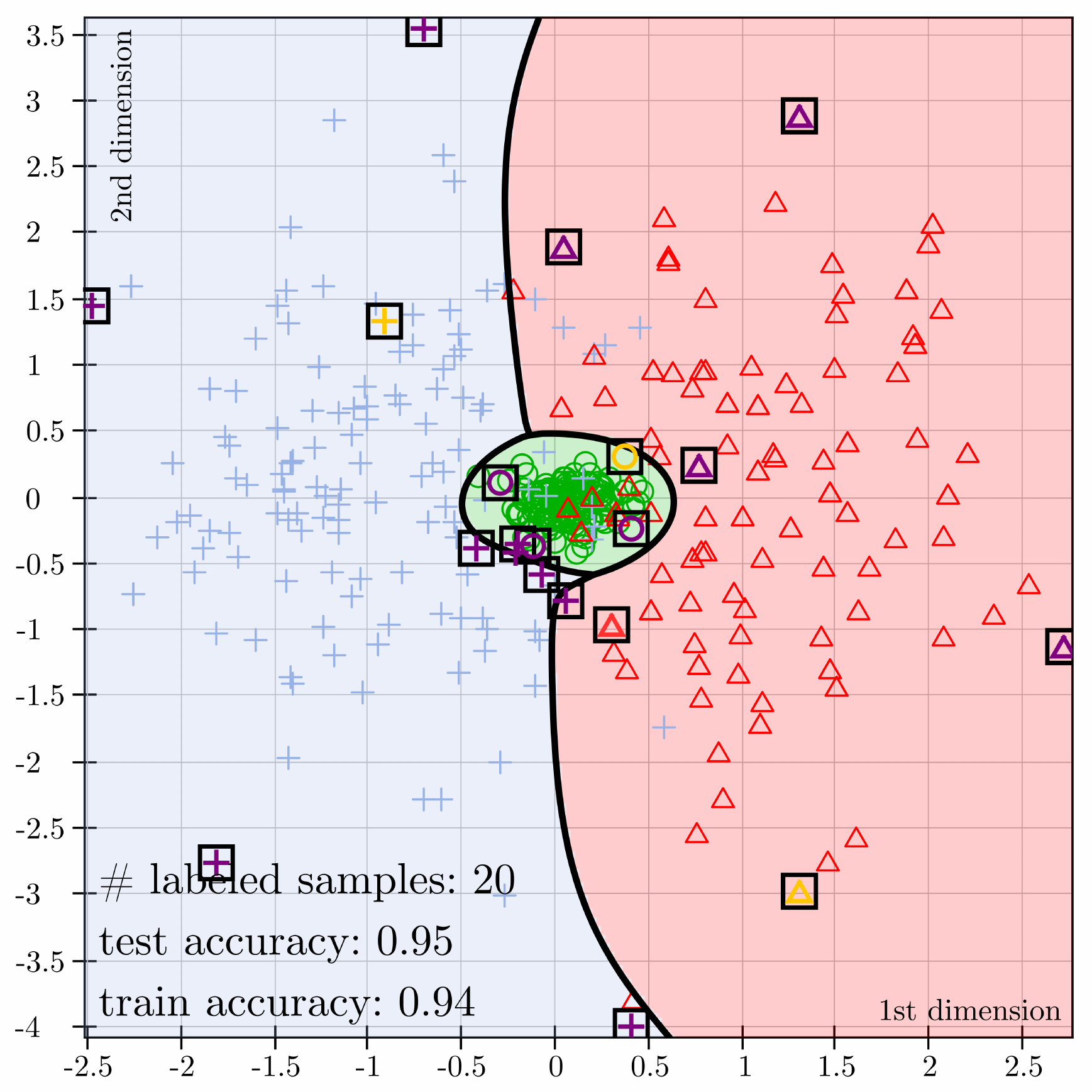}}
		\hfill
		\subfigure[Classification accuracy vs.~learning cycle (Random).]{\label{fig:example_alD}\includegraphics[width=0.317\textwidth]{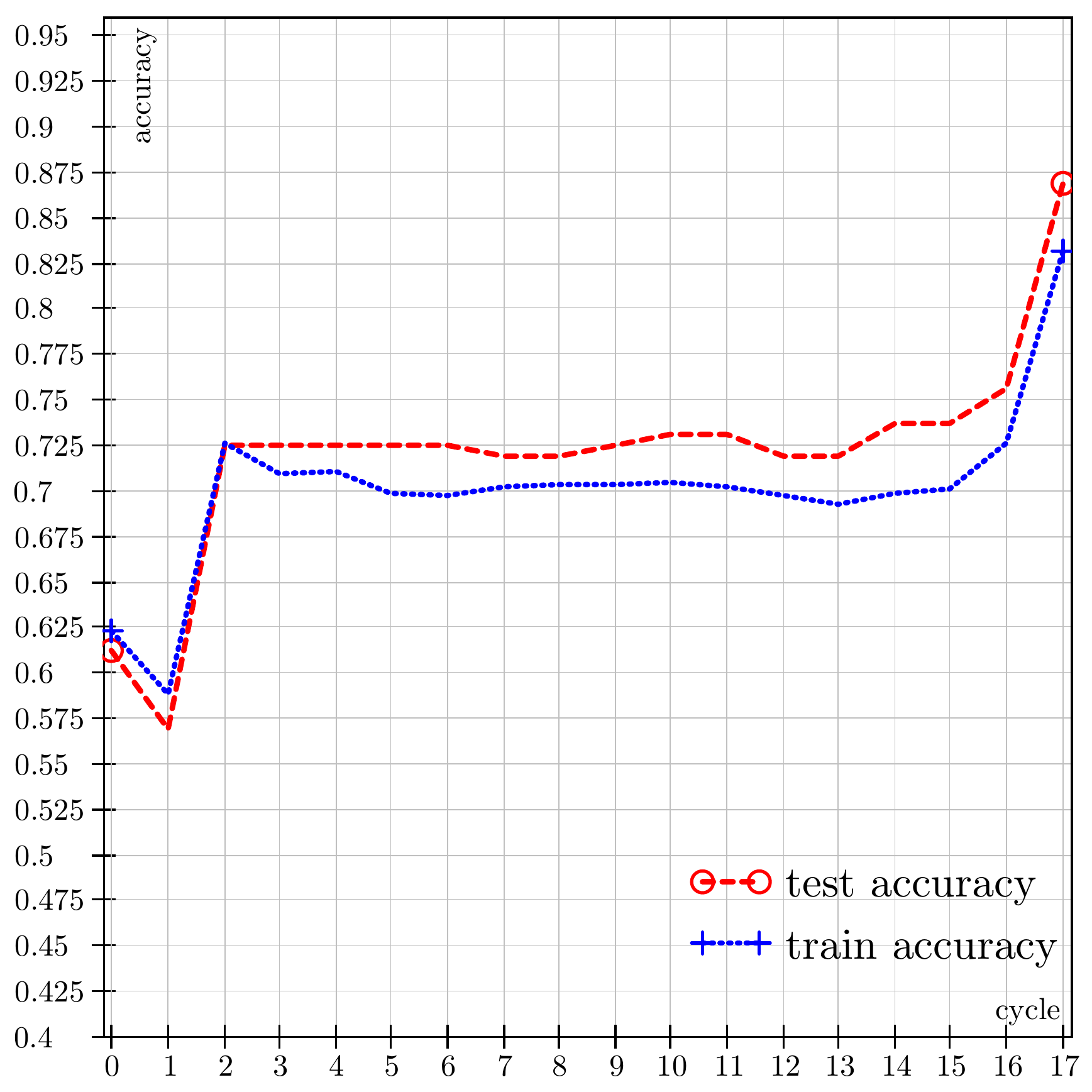}}\quad
		\subfigure[Classification accuracy vs.~learning cycle (US).]{\label{fig:example_alE}\includegraphics[width=0.317\textwidth]{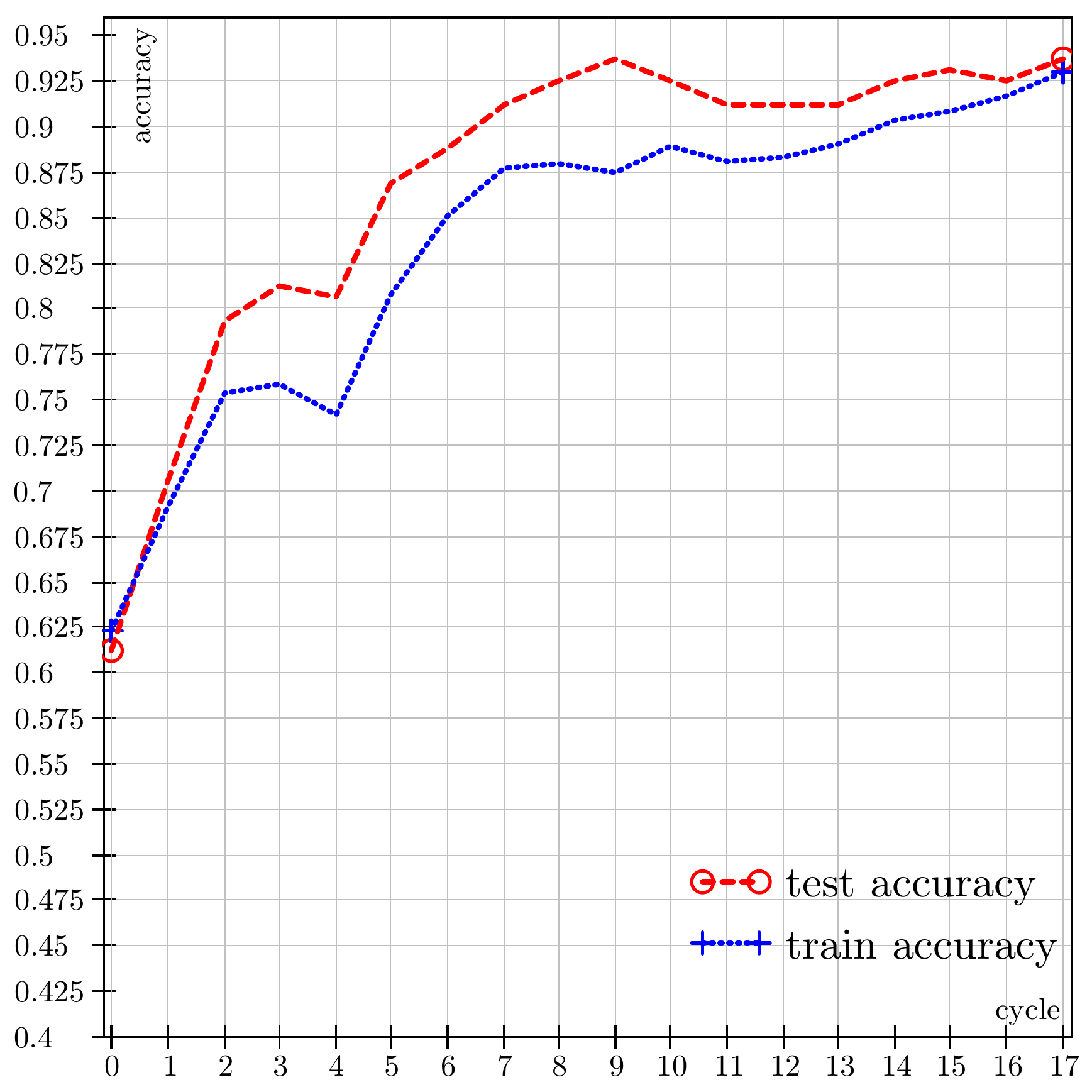}}\quad
		\subfigure[Classification accuracy vs.~learning cycle (4DS and RWM kernel).]{\label{fig:example_alF}\includegraphics[width=0.317\textwidth]{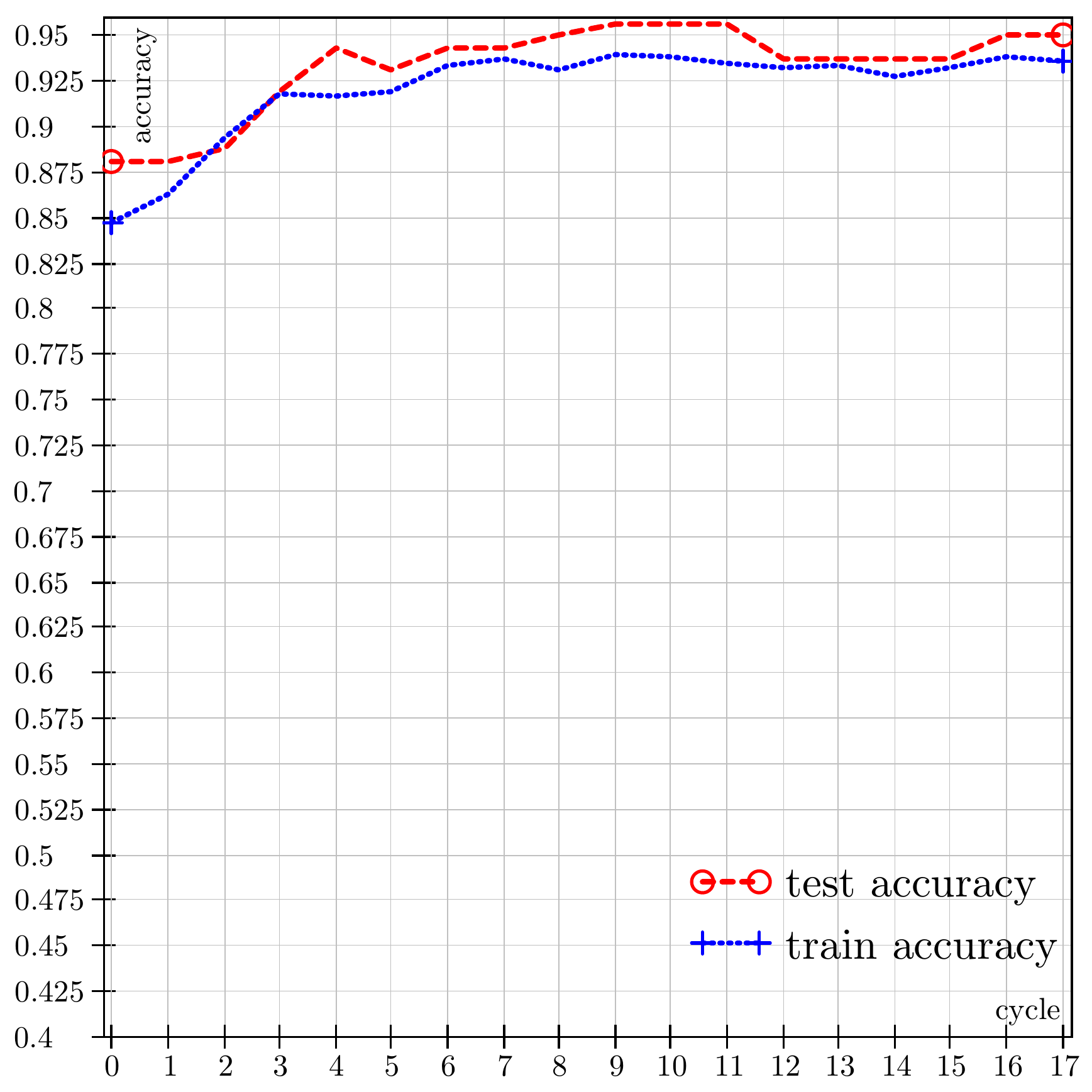}}
		\caption{Active training of SVM based on an artificial generated three class problem. Figs.~\ref{fig:example_alA}--\ref{fig:example_alC} show SVM, which were actively trained based on $20$ labeled samples (depicted in orange, violet or red color). Figs.~\ref{fig:example_alA}--\ref{fig:example_alC} show the corresponding learning curves (classification accuracies on the test and training data). Fig.~\ref{fig:example_alA} display an SVM with RBF kernel, which was trained on randomly selected samples, i.e.,~without considering any knowledge (test accuracy of $87\%$). Fig.~\ref{fig:example_alB} presents an SVM with RBF kernel, which was actively trained with US ($94\%$). Fig.~\ref{fig:example_alC} shows an SVM with RWM kernel, where the samples were actively selected with 4DS ($95\%$).}
		\label{fig:example_al}
	\end{center}
\end{figure}

In the following, we will illustrate with a simple example (1) the potential of AL and (2) the properties of our new approach to train an SVM in a semi-supervised active fashion (see Fig.~\ref{fig:example_al}).

Assume we observe three processes producing data in a two-dimensional input space (small blue plus signs, green circles and red triangles), that shall be classified by an SVM. For our example, we generated a set of $500$ samples using a Gaussian mixture model (GMM) with three components ($\boldsymbol{\mu}_1 = (-1.00, 0.00)^{\text{T}}$, $\boldsymbol{\mu}_2 = (0.00,0.00)^{\text{T}}$, $\boldsymbol{\mu}_3 = (1.00,0.00)^{\text{T}}$, $\boldsymbol{\Sigma}_1 =\left(\begin{smallmatrix}0.50 & 0.00 \\ 0.00 & 1.50 \end{smallmatrix}\right)$, $\boldsymbol{\Sigma}_2 =\left(\begin{smallmatrix}0.04 & 0.00 \\ 0.00 & 0.03 \end{smallmatrix}\right)$, $\boldsymbol{\Sigma}_3 =\left(\begin{smallmatrix}0.50 & 0.00 \\ 0.00 & 0.50 \end{smallmatrix}\right)$, $\pi_{1}=\pi_{2}=\pi_{3}=0.33$) and split them into training and test data according to a 5-fold cross-validation. Fig.~\ref{fig:example_al} shows only the training data of the first cross-validation fold, that corresponds to the pool of unlabeled samples $U$ and the initially labeled samples $L$, where we assume that label information is only given for three randomly selected samples, one for each class (shown in orange color). After that, in each query round (learning cycle) only one sample is actively selected, labeled by a domain expert, and then considered for SVM training. Here, the samples are shown in boldface and colored red, if they were selected in the current query round and otherwise violet. The samples that are marked with rectangles correspond to the support vectors of the respective SVM. Fig.~\ref{fig:example_alA} shows an SVM with RBF kernel, that was actively trained based on $20$ selected and thus labeled samples. Here, the ``knowledge'' of the SVM was not considered for the sample selection since the samples were selected randomly, a strategy also called random sampling (Random). Fig.~\ref{fig:example_alD} depicts the corresponding learning process of the SVM with RBF kernel and here we see that the SVM reaches an accuracy of about $87\%$ on the test data after the execution of the last query (i.e., with a set of 20 labeled samples). Fig.~\ref{fig:example_alB} also shows the AL process of an SVM with RBF kernel, but this time the samples are actively selected with an uncertainty sampling (US) strategy. This means that in each AL cycle the sample lying nearest to the current decision boundary of the SVM is selected. We can see, that the SVM reaches an test accuracy of $94\%$, if we recognize the knowledge of the SVM for active sample selection with US. In addition, Fig.~\ref{fig:example_alE} shows that the SVM reaches this accuracy already with only $12$ labeled samples. In Fig.~\ref{fig:example_alC} we see an SVM with RWM kernel that was also actively trained based on $20$ labeled samples, but $17$ of these samples (not the first three) were selected with the selection strategy 4DS. Here, the SVM yields the highest test accuracy of $95\%$. Moreover, Fig.~\ref{fig:example_alF} shows that the SVM with RWM kernel trained with only three labeled samples (i.e., with $0.75\%$ of the overall set of available training samples) reaches a significantly higher test accuracy ($87.5\%$) compared to the SVM with RBF kernel (slightly more than $61\%$). Further, after the active selection of three additional samples, the SVM with RWM kernel always yields a test accuracy of more than $92.5\%$ throughout the AL process.

This example shows on the one hand, that AL obtains respectable results if the active sample selection is done deliberately (US vs. Random). On the other hand it can be seen that considering structure information may result in a more efficient and effective AL process of a discriminative classifier such as SVM.

\section{Related Work}
\label{sec:overview}
This section discusses related techniques that are specially developed to train SVM  by combining AL and SSL. For a detailed overview in condensed form about the current state-of-the-art in the field of AL, SSL, and techniques that combine these two approaches not only for SVM we refer to \cite{RS13, RS15, RCS14}. In \cite{RS13}, the main five categories of selection strategies  are explained, that are used in the field of (pool-based) AL to find informative samples: \textit{uncertainty sampling, density weighting, estimated error reduction, AUC maximization}, and \textit{diversity sampling}. In addition, we refer to \cite{Settles09, JH09} for a more detailed overview on this topic. Related work in the field of SSL, that make also use of the unlabeled data to train discriminative classifiers, such SVM for instance, is given in \cite{RS15}. In \cite{RCS14}, related work is presented that combines AL and SSL to train (generative) classifiers actively in a more effective way, e.g., either by using modeling techniques for capturing and refining structure information, so that only the most informative samples are selected, or by querying a ``second'' paradigm instead of an human expert to reduce the labeling costs.

SSL is based on the idea that a classifier with high generalization capability is trained using information of labeled and unlabeled samples. In combination with AL, this means that the active selection of outliers shall be avoided. An SVM approach that tries to solve this problem is called \textit{representative sampling}~\cite{XYTXW03}. An SVM based on an initially labeled set of samples $\mathit{L}$ is trained. Then, in each iteration of the AL process the clustering algorithm $c$-means is used to cluster all unlabeled samples $\mathbf{x} \in \mathit{U}$ that lie within the current margin of the SVM. From each of the resulting $k$ clusters the sample with minimal average distance to all samples that belong to the same cluster is queried. While only unlabeled samples within the margin of the SVM are presented to an oracle or human expert, representative sampling selects on the one hand samples for which the SVM is most uncertain regarding their class assignments and on the other hand samples that are representative regarding $\mathit{U}$. The main drawback of representative sampling is that clustering techniques are in general computationally expensive. An approach that avoids this these efforts and still makes use of the unlabeled samples, is called \textit{hinted sampling}~\cite{LFL12}. Hinted sampling selects also samples, that are both representative and uncertain (i.e., the SVM is most uncertain about their class membership), but doing so the unlabeled samples are neither clustered nor assigned with a class label (as with \textit{transductive SVM} (TSVM)~\cite{Joachims99}). Rather, the samples in $\mathit{U}$ are used as \textit{hints} for finding a decision boundary, that on the one hand leads to a small classification error regarding the labeled training samples ($\mathbf{x} \in \mathit{L}$) and on the other hand takes course close to (or through) the set of unlabeled samples (hints set). Since in each query round the informative samples are selected with help of uncertainty sampling (US), the selected ones are located close to the decision boundary of the SVM and are representative, too. In general, the \textit{HintSVM} can be regarded as an SSL approach~\cite{CSZ06}, but HintSVM differ significantly from typical SSL techniques as the \textit{semi-supervised SVM} (S$^3$VM)~\cite{CSK08,RSM11,FGQZ11}, since these approaches try to find decision boundaries that are located as distant as possible to the samples in $\textit{U}$ (in low density regions).

TSVM assume that the labeled and unlabeled samples are independent and identically distributed (i.i.d.). But this assumption is not guaranteed in AL. This problem is also related to \textit{transfer learning}~\cite{PY10}, because the distributions of the training and test data may differ. For this reason, TSVM are not well suited for AL, since an active sample selection (e.g.,~with US) easily results in an overfitting of the TSVM towards the already queried samples~\cite{WYZ11}. Therefore, in~\cite{CC10} an approach is presented, that focuses more on SSL than on AL by using AL techniques only to find a good initialization to train a TSVM or related SSL approaches. Another approach, published in~\cite{SYH11} and used to extract protein sequences, combines AL and SSL in the same way, too. For this, an SVM is initially trained regarding a small set of labeled samples $\mathit{L}$ and then used to iteratively query samples that the SVM would assign to classes with high uncertainty. These samples are then labeled by an expert and the SVM is retrained. If $\mathit{L}$ reaches a certain number of labeled samples, the AL process stops and a \textit{deterministic annealing SVM} (DA-SVM)~\cite{SKC06}, that corresponds to an implementation variant of TSVM, is used to label the yet unlabeled samples ($\mathbf{x} \in \mathit{U}$). In contrast,~\cite{LXQ13}~focuses on AL and uses SSL only as additional help to train an SVM as efficiently as possible, i.e.,~with the smallest possible number of expert queries. For this purpose, \cite{LXQ13} combines \textit{self-training} \cite{Scu65} with US to select on the one hand samples that are representative and can, therefore, be labeled by the SVM itself, and on the other hand ``uncertain'' samples, i.e.,~samples that are difficult to label, and, therefore, labeled by an human expert. Again, at the beginning of the learning process an SVM is trained based on a small initially labeled data set. Then, the following two steps are alternately executed: In the first step, $m$ samples are iteratively selected by means of US. The SVM is retrained once a new sample is selected, labeled by an domain expert, and added to the training set $\mathit{L}$. In each of these $m$ query rounds, the unlabeled samples $\mathbf{x} \in \mathit{U}$ are also labeled by the SVM and, based on these class assignments for each $\mathbf{x}$, a rate of class change $q_{\mathbf{x}}$ is determined or updated. Then, in the second step, the unlabeled samples $\mathbf{x} \in \mathit{U}$, that have a rate of class change $q_{\mathbf{x}}$ equal to zero, are divided into $|\mathit{C}|$ subsets according to the class label $c \in \mathit{C}$, assigned by the SVM in the first step. These subsets are then used to select one representative sample for each class $c$. Here, a sample is the more representative for the class $c \in \mathit{C}$, the more its distance to the current decision boundary of the SVM corresponds to the median of the distances of all samples $\textbf{x}$ belonging to the $c$-th subset. The representative samples are subsequently labeled with the class label of the subset which they belong and added to $\mathit{L}$. If no stopping condition is met, the first step is executed again. The main problem of this approach is, that in the presence of data with bimodal or multimodal class distributions the representative samples are assigned ``untrusted'' class labels, resulting in a deterioration of the classification performance of the actively trained SVM~\cite{KPI14}.

Other works \cite{HJZL09, WMZL05} combine AL and SSL to create an efficient approach for content-based image retrieval. Here, several images (samples) are selected iteratively regarding class assignments for which the actively trained \textit{Laplacian SVM} (LapSVM) \cite{BNS06, WL09, NCW15} is most uncertain. The LapSVM follows the principle of manifold regularization, integrating an ``intrinsic regularizer'' \cite{MB11} into the SVM training. This regularizer is estimated based on the labeled and unlabeled samples with help of a Laplacian graph which can be seen as a non-parametric density estimator. To avoid the selection of images with ``redundant'' content a ``min-max'' approach \cite{HJZL09} is used that ensures the diversity of queried images. In \cite{WMZL05}, an extension of the LapSVM for image retrieval is presented, where multiple Laplacian graphs for heterogeneous data, such as content, text, and links are used. Both approaches show that the LapSVM is very well suited for AL.

What do we intend to make better or in a different way? An actively trained SVM with our new RWM kernel uses a parametric density estimation to capture structure information, because a parametric estimation approach often leads to a more ``robust'' estimate than a non-parametric approach (used by the LapSVM (LAP kernel), for instance). An SVM with RWM kernel can be actively trained with help of any selection strategy and the RWM kernel can be used in combination with any standard implementation of SVM (e.g.,~LIBSVM \cite{libsvm}) or with any solver for the SVM optimization problem (e.g.,~SMO) without any additional algorithmic adjustments or extensions. In addition, the data modeling  can be conducted once, i.e.,~\textit{offline}, based on the unlabeled samples before the AL process starts (i.e., unsupervised). Moreover, it is also possible to refine the density model based on class information becoming known during the AL process (\textit{online}), so that the model captures the structure information of the unlabeled and labeled samples. Here, local adaptions of the model components are used for an efficient model refinement.     

\section{Theoretical and methodical foundations}
\label{sec:foundations}

In this section we will (1) describe our (unsupervised) approach to capture structure information in data which is based on probabilistic mixture models. Next, we show (2) how class information can be used either to extend these models to classifiers or to refine the density model.  
Based on that we (3) define a new similarity measure that allows us to exploit structure in data for SVM training, i.e.,~to determine the support vectors. (4) We define the multi-criteria selection strategy 4DS, that uses structure information to select informative samples actively. Bringing all together, we (5) extend the traditional PAL cycle so that it starts with an empty training set and iteratively refines the data modeling based on the known classes to train an SVM in an improved semi-supervised active fashion. 

\subsection{Capturing Structure in Data with Probabilistic Mixture Models}
\label{subsec:capturing}

Assume we have a set $X$ of samples, where each sample $\mathbf{x} \in X$ can be described with a $\mathcal{D}$-dimensional vector, then we capture the structure information contained in $X$ with help of the mixture model 
\begin{equation}
p(\mathbf{x}) = \sum_{\textit{j}=1}^{\textit{J}}p(\mathbf{x}|j)p(j),	
\end{equation}
consisting of $\textit{J}$ densities with $\textit{j} \in \left\{1,\dots, \textit{J}\right\}$ In this probabilistic context $\mathbf{x}$ is seen as a random variable. Here, the conditional densities $p(\mathbf{x}|j)$ are the model \textit{components} and the~$p(j)$ are multinomial distributions with parameters $\pi_\textit{j}$ (\textit{mixing coefficients}). 

On the basis of $p(\mathbf{x})$ we can determine for a particular sample $\mathbf{x}'$ and a component $j$ the value 
\begin{equation}
\rho_{\mathbf{x}',\textit{j}} = p(j|\mathbf{x}') = \frac{p(\mathbf{x}'|j)\pi_{\textit{j}}}{\sum_{\textit{j}'=1}^{\textit{J}}p(\mathbf{x}'|j')\pi_{\textit{j}'}}.	
\end{equation}
These values are called \textit{responsibilities} of the $\textit{J}$ components for the generation of the sample $\mathbf{x}'$. That is, a responsibility of a component is an estimate for the value of a latent variable that describes from which process in the real world a sample originates. The idea behind this approach is that components model processes in the real world from which the samples we observe are assumed to originate (or: to be ``generated'').

Which kind of density function can be used for the components? In general, a $\mathcal{D}$-dimensional sample $\mathbf{x}$ may have~$\mathcal{D}_{\text{cont}}$ (i.e., real-valued) dimensions (attributes) and $\mathcal{D}_{\text{cat}}=\mathcal{D}-\mathcal{D}_{\text{cont}}$ categorical ones. Without loss of generality we arrange these dimensions such that 
\begin{equation}
	\mathbf{x}=\left(x_1, \cdots, x_{\mathcal{D}_\text{cont}},x_{\mathcal{D}_\text{cont}+1},\cdots, x_{\mathcal{D}}\right)^{\text{T}.}
\end{equation}
Note that we italicize $x$ when we refer to single dimensions. The continuous part of this vector $\mathbf{x}^\text{cont} = (x_1,\dots,x_{\mathcal{D}_\text{cont}})$
with $x_d \in \mathbb{R}~$ for all $d \in \{1,\dots,\mathcal{D}_\text{cont}\}$ is modeled with a multivariate \emph{normal} (i.e., Gaussian) distribution with
\textit{center} (expectation) $\boldsymbol{\mu}$ and \textit{covariance matrix} $\boldsymbol{\Sigma}$. With $\text{det}(\cdot)$ denoting the determinant of a matrix we use the
model
\begin{equation}
{\cal N}(\mathbf{x}^\text{cont}|\boldsymbol{\mu},\boldsymbol{\Sigma}) =   
\frac{1}{(2\pi)^{\frac{\mathcal{D}_\text{cont}}{2}} \mathrm{det}(\boldsymbol{\Sigma})^{\frac{1}{2}}} \exp\left(-0.5 \, \left(\Delta_{\boldsymbol{\Sigma}}(\mathbf{x}^\text{cont},\boldsymbol{\mu})\right)^2\right)
\end{equation} 

with the distance measure 
$\Delta_{\boldsymbol{\Sigma}}(\mathbf{a}, \mathbf{b})$ given by 
\begin{equation}
\label{eq:maha}
\Delta_{\boldsymbol{\Sigma}}(\mathbf{a}, \mathbf{b}) = \sqrt{(\mathbf{a} - \mathbf{b})^\text{T} \boldsymbol{\Sigma}^{-1} (\mathbf{a} - \mathbf{b})}.
\end{equation}
$\Delta_{\boldsymbol{\Sigma}}(\mathbf{a}, \mathbf{b})$ defines the \emph{Mahalanobis distance} of vectors $\mathbf{a}, \mathbf{b} \in \mathbb{R}^{\mathcal{D}_\text{cont}}$ based on a $\mathcal{D}_\text{cont} \times \mathcal{D}_\text{cont}$ covariance
matrix $\boldsymbol{\Sigma}$. For many practical applications, the use of Gaussian components can be motivated by the generalized \textit{central limit theorem} which roughly states that the sum of independent samples from any distribution with finite mean and variance converges to a normal distribution as the size of the data set goes to infinity 
(cf.,~for example~\cite{DHS01}).

For categorical dimensions we use a 1-of-$K_d$ coding scheme where $K_d$ is the number of possible categories of attribute $\boldsymbol{x}_d$ ($d \in \{\mathcal{D}_{\text{cont}}+1,\dots,\mathcal{D}\}$).
The value of such an attribute is represented by a vector
$
\boldsymbol{x}_d = (x_{d_1},\dots,x_{d_{K_d}})
$
with $x_{d_k} = 1$ if $\boldsymbol{x}_d$ belongs to category $k$ and $x_{d_k} = 0$ otherwise. The categorical dimensions are modeled by means of special cases of \emph{multinomial} distributions. That is, for an input dimension
(attribute) $\boldsymbol{x}_d \in \{\boldsymbol{x}_{\mathcal{D}_{\text{cont}}+1},\dots,\boldsymbol{x}_\mathcal{D}\}$ we use
\begin{equation}
{\cal M}(\boldsymbol{x}_d|\boldsymbol{\delta}_d) = \prod_{k=1}^{K_d} (\delta_{d_k})^{x_{d_k}}
\end{equation}
with $\boldsymbol{\delta}_d = (\delta_{d_1},\dots,\delta_{d_{K_d}})$ and the restrictions $\delta_{d_k} \geq 0$ and $\sum_{k=1}^{K_d}\delta_{d_k} = 1$. The
rationale for using such distributions for categorical variables is obvious as any given distribution of a categorical variable can
perfectly be modeled.

We assume that the categorical dimensions are mutually independent and that there are no dependencies between the categorical and the continuous dimensions. Then, the component densities $p(\mathbf{x}|j)$ are defined by
\begin{equation}
p(\mathbf{x}|\textit{j}) = {\cal N}(\mathbf{x}^\textrm{cont}|\boldsymbol{\mu}_\textit{j}, 
\boldsymbol{\Sigma}_\textit{j}) \cdot \prod_{\mathclap{d=\mathcal{D}_\text{cont}+1}}^{\mathcal{D}} 
{\cal M}(\mathbf{x}_d|\boldsymbol{\delta}_{\textit{j},d}).
\end{equation}

How can the various parameters of the density model $p(\mathbf{x})$ be determined? Assuming that the samples $\mathbf x \in X$ are independent and identically distributed (i.i.d.), we perform the parameter estimation by means of \textit{variational Bayesian inference} (VI) which realizes the Bayesian idea of regarding the model parameters as random values whose distributions have to be trained \cite{FS09,Bis06}. This approach has two important advantages: First, the estimation process is more robust, i.e., it avoids ``collapsing'' components, so-called singularities, and second it optimizes the number of components by its own. That is, the training process starts with a large number of components and prunes components automatically until an appropriate number $\textit{J}$ is reached. For a more details of VI see \cite{FS09,Bis06}.

\subsection{Iterative Refinement of Models Capturing Structure in Data}
\label{subsec:refinement}

If we take class information $c \in \{1,\dots, C\}$ into account, we can extend our estimated probabilistic mixture model $p(\mathbf{x})$ to a generative classifier called CMM (classifier based on mixture models).
However, to obtain high classification performance it is important to recognize ``overlapping'' processes that either generate samples belonging to different classes or cannot be modeled perfectly. For this reason, we distinguish between two different training techniques: 

The first one, called \textit{shared-components classifier} (\CMMsha), captures  structure information in an \textit{unsupervised way} (as described in Section~\ref{subsec:capturing}) with a shared-components density model and subsequently extends this model to a classifier using class labels. That is, to minimize the risk of classification errors we compute for an input sample $\mathbf{x}'$ the posterior distribution $p(c|\mathbf{x}')$ and select then (according to the \textit{winner-takes-all} principle) the class $c$ with the highest posterior probability. In this case, the distribution $p(c|\mathbf{x})$ is decomposed as follows: 
\begin{equation}
	p(c|\mathbf{x})
= \sum_{\textit{j}=1}^{\textit{J}} {p(c|j)} \cdot {p(j|\mathbf{x})}. 	
\end{equation}
Thus, the \CMMsha classifier is based on a \textit{single mixture model} $p(\mathbf{x}) = \sum_{\textit{j}'=1}^{\textit{J}} p(\mathbf{x}|j') p(j')$ (i.e., ``shared'' by all classes), where the $p(c|j)$ are multinomial distributions with parameters $\xi_{\textit{j},c}$. These parameters can be estimated in a \textit{supervised} step for samples with given class labels using the responsibilities: 
\begin{equation}
\label{eq_cmm_shared}
\xi_{\textit{j},c}  =  \frac{1}{\textit{N}_\textit{\textit{j}}} \sum_{\mathbf{x} \in \mathit{X}_c} \rho_{\mathbf{x},\textit{j}}
\end{equation}
with $\textit{N}_\textit{j} = \sum_{\mathbf{x}' \in X} \rho_{\mathbf{x}',j}$ being the ``effective'' number of samples ``generated'' by component $\textit{j}$ (see \cite{FS09} for more details) and $\mathit{X}_c$ corresponds to the subset of all samples $\mathbf{x} \in X$ for which $c$ is the assigned target class. 

The second one, called \textit{separate-components classifier} (\CMMsep), also uses class information to assign the model components to classes but (in contrast to \CMMsha) it builds a separate-components density model, so that the data modeling is performed in a \textit{supervised} way, too. In this case the distribution $p(c|\mathbf{x})$ is decomposed in another way:
\begin{equation}
\label{eq_cmm_separate}
	    p(c|\mathbf{x})
    = \frac{p(c) \sum_{\textit{j}=1}^{\textit{J}_c} p(\mathbf{x}|c,\textit{j}) p(j|c)}{p(\mathbf{x})}.
\end{equation}
Therefore, the \CMMsep classifier is based on a number of $C$ \textit{mixture density models} $p(\mathbf{x}|c) = \sum_{j=1}^{J_c}p(\mathbf{x}|c,\textit{j})p(j|c)$ (one for each class), so that $p(\mathbf{x}) = \sum_{c'=1}^{C} \left(p(c') \sum_{\textit{j}=1}^{J_{c'}} p(\mathbf{x}|c',\textit{j}) p(j|c')\right)$. Here, the conditional densities $p(\mathbf{x}|c,\textit{j})$ ($c\in \{1,\dots,C\}, j \in \{1, \ldots, J_c\}$) are the model \textit{components}, the $p(j|c)$ are multinomial distributions with parameters $\pi_{c,\textit{j}}$ (\textit{class dependent mixing coefficients}), and $p(c)$ is a multinomial distribution with parameters $\xi_{c}$ (\emph{class priors}). To evaluate Eq.~\eqref{eq_cmm_separate}, we have to exploit the fact that in the case of \CMMsep we treat all classes \textit{separately} and, thus, components are uniquely assigned to \textit{one} class, i.e., $p(c|j) \in \left\{0,1\right\}$. To train the \CMMsep we have to split the entire training set $X$ into $C$ subsets $X_c$ first, each containing all samples of the corresponding class $c$, i.e., $X_c = \{\mathbf{x}_n|\mathbf{x}_n \text{ belongs to class } c\}$ with $\sum_{c=1}^{C} | X_c| = |X|$, where $|\cdot|$ denotes the cardinality of a set. Then, for each $X_c$, a mixture model is trained separately by means of VI as sketched in Section~\ref{subsec:capturing}. After this, we have found parameter estimates for the $p(\mathbf{x}|c,j)$ and $p(j|c)$, cf.~Eq.~\eqref{eq_cmm_separate}. The parameters for the class priors $p(c)$ are estimated with 
\begin{equation}
 \xi_c = \frac{|\mathit{X}_c|}{|\mathit{X}|}.
\end{equation}

Comparing the two modeling approaches from the viewpoint of AL we can state that the main advantage of the \CMMsha is that underlying mixture density model can be trained completely unsupervised and only the gradual assignments of its components to classes have to be determined in a subsequent, supervised step. The key advantage of the \CMMsep -- for which we need a fully labeled data set already for the first modeling step -- is that label information may improve the data modeling, e.g., we may expect a better discrimination of highly overlapping densities belonging to different classes. But, if we assume that no class labels are available at the beginning of an AL process, we must start with a shared-components density model only.

Thus, the key idea of our AL technique is to start with the density model $p(\mathbf{x})$ being part of a \CMMsha to model structure in data. During the AL process, when more and more labeled data become available, we will iteratively revise the model using label information. This can roughly be seen as a kind of transformation from a \CMMsha towards a \CMMsep until a stopping criterion is met. To keep the computational costs for this transformation process as low as possible we introduce a \textit{transductive learner} into the AL process, that adapts the density mixture model by means of local modifications. This learner consists of several steps, and we look briefly (and in a simplifying way) at the most important ones: First, we realize that at least one model component models a set of samples that belong to different classes (i.e.,~labeled by the oracle or human expert) using an interestingness measure called \textit{uniqueness}. These components are denoted as \textit{disputed}. Second, we detect the (labeled and unlabeled) samples that are modeled by the disputed components. Third, the unlabeled ones of these samples are  labeled transductively by a sample-based classifier (related to a k-nearest-neighbor approach). Fourth, this set of (now fully labeled) samples is used to train a local separate-components model (\CMMsep), whose components are then fused or combined with all non-disputed model components (of the initial shared-components model) in the fifth step. For a more detailed description of the transductive learner see \cite{RCS14}.       

\subsection{Exploiting Structure in Data for Active Semi-Supervised SVM Training}
\label{subsec:exploitstructure}

As described above, we gather structure information by means of probabilistic mixture models and if necessary we revise these models as soon as class information becomes available during the AL process. But, how can we integrate this information into the active training process of a discriminative classifier (such as SVM) to optimize its learning performance? Here, we can distinguish between two possibilities: First, we can use structure information to train an SVM in each AL round in an semi-supervised fashion. For this, we define the new data-dependent RWM kernel, that is based on previously described mixture models. Second, we consider structure information for the active sample selection with help of a self-adaptive, multi-criteria selection strategy called 4DS. Both are described in the following.

\subsubsection{Data-Dependent Kernel for Semi-Supervised SVM Training}
\label{subsubsec:RWM}

In principal, generative classifiers such as CMM often perform worse than discriminative classifiers such as SVM in many applications. But, on the other hand no density information (or, more general, information concerning structure of data in the input space of the classifier) can be extracted from standard SVM to improve its training process or use this information for active sample selection. Thus, we developed a new data-dependent kernel for SVM, called \textit{responsibility weighted Mahalanobis (RWM}) kernel. This kernel assesses the similarity of any two samples by means of a parametric density model. Basically, this kernel emphasizes the influence of the model components from which two samples that are compared are assumed to originate (that is, the ``responsible'' model components).

With help of the Mahalanobis distance measure described in Section~\ref{subsec:capturing} we can determine the distance of any two samples in the $\mathcal{D}$-dimensional input space with respect to a process modeled by a single Gaussian component with given mean $\mu$ and covariance matrix $\boldsymbol{\Sigma}$. In general, however, we need a number of $J > 1$ components to model densities accurately. 
Assuming, a given set of samples $X$ with only real-valued dimensions. Then, we model $X$ only with a Gaussian mixture model (GMM) as described above. In order to consider the distances of two samples with respect to all components contained in the GMM we need a similarity measure that combines these distances by means of a linear combination. This leads us to the new RWM kernel that weights the Mahalanobis distance $\Delta_{\boldsymbol{\Sigma}_j}$ according to the responsibilities of the $j$-th Gaussian for the generation of the two considered samples $\mathbf{x}_n$, $\mathbf{x}_m \in X$:

\begin{equation}
\label{eq:rwm_kernel}
K_\mathbf{RWM}(\mathbf{x}_{n},\mathbf{x}_{m})=\text{exp}\left(-\gamma\left(\Delta_{\mathbf{RWM}}\left(\mathbf{x}_{n}-\mathbf{x}_{m}\right)\right)^2\right).
\end{equation}

Here, $\gamma = \frac{1}{2\sigma^2} \in \R^{+}$ is the kernel width and the similarity measure $\Delta_{\mathbf{RWM}}\left(\mathbf{x}_{n},\mathbf{x}_{m}\right)$ is defined as follows:
\begin{equation}
\Delta_{\mathbf{RWM}}(\mathbf{x}_{n},\mathbf{x}_{m}) = \sum_{j=1}^{J}\left(\frac{1}{2} \left(\rho_{\mathbf{x}_n,j} + \rho_{\mathbf{x}_m,j}\right) \Delta_{\mathbf{\boldsymbol{\Sigma}}_{j}}(\mathbf{x}_{n},\mathbf{x}_{m})  \right).
\end{equation}

The main advantages of the RWM kernel are (for more details see \cite{RS15}): (1) Standard training techniques such as SMO and standard implementations of SVM auch as \textsf{libsvm} \cite{libsvm} can be used with RWM kernels without any algorithmic adjustments or extensions as only the kernel matrices have to be provided. (2) In case of SSL this kernel outperforms some other kernels that capture structure in data such as the Laplacian kernel (Laplacian SVM) \cite{MB11} that can be regarded as being based on non-parametric density estimates. (3) C-SVM with RWM kernels can easily be parametrized using existing heuristics for RBF kernels relying on line search strategies in a two-dimensional parameter space. This does not hold for the Laplacian kernel, for example. 

In most classification problems we also have categorical (non-ordinal) input dimensions that typically cannot be handled as continuous ones. Assume we are given a set $X$ of samples where each $\mathcal{D}$-dimensional sample has $\mathcal{D}_{\text{cont}}$ continuous and $\mathcal{D}_\text{cat} = \mathcal{D}-\mathcal{D}_{\text{cont}}$ dimensions. Here, each categorical dimension $\boldsymbol{x}_{d}$ ($d \in \left\{\mathcal{D}_{\text{cont}}+1, \dots, \mathcal{D}\right\}$) has $K_{d}$ different categories for which we use a $1$-of-$K_d$ coding scheme (cf.~Section \ref{subsec:capturing}). 

Then, we extend the RWM kernel accordingly:

\begin{equation}
K_{\mathbf{RWM}}(\mathbf{x}_n,\mathbf{x}_m) = 
\exp\left(-\gamma \left(\alpha\left(\Delta_{\mathbf{RWM}}(\mathbf{x}_{n}',\mathbf{x}_{m}')\right)^2 + \beta\left(\Delta_{\mathbf{0/1}}(\mathbf{x}_{n}'',\mathbf{x}_{m}'')\right)^2\right) \right)
\end{equation}

with weighting factors $\alpha, \beta\in [0,1]$, $\mathbf{x}_n,\mathbf{x}_m \in X$ and $\mathbf{x}_n',\mathbf{x}_m' \in \R^{\mathcal{D}_\text{cont}}$, $\mathbf{x}_n'',\mathbf{x}_m'' \in \mathbb{B}^{\mathcal{D}_\text{cat}'}$ (with $\mathcal{D}_\text{cat}'=\sum_{d=1}^{\mathcal{D}_\text{cat}} K_{d}$) only containing the values of the respective continuous and (binary encoded) categorical dimensions, respectively. For the categorical dimensions, we define

\begin{equation}
 \Delta_{\mathbf{0/1}}(\mathbf{x}_{n}'',\mathbf{x}_{m}'') = \sum_{d = 1}^{\mathcal{D}_\text{cat}} (1 - {\delta_{d}}_{n,m}) 
 \end{equation}
with 
\begin{equation}
{\delta_{d}}_{n,m} = 
\begin{cases}
1 & \text{ for } \left(\boldsymbol{x}_d\right)_n = \left(\boldsymbol{x}_d\right)_m\\
0 & \text{ otherwise } \end{cases},
\end{equation}
i.e.,~simply by checking the values in the different dimensions for equality. If necessary, it is also possible to weight the categorical part and the continuous part differently by means of the parameters $\alpha, \beta \in [0,1]$. If $\alpha$ and $\beta$ are both set to $1$ and the covariance matrix of each model component corresponds to the identity matrix, then the RWM kernel behaves such as an RBF kernel with binary encoded categorical dimensions.
 
\subsubsection{Active Sample Selection Considering Structure in Data}
\label{subsubsec:4DS}

As already mentioned, the main goal of AL is to get the best possible classifier at the lowest possible labeling costs. Therefore, a selection strategy for informative samples must be able to detect all decision regions (exploration phase) and to fine-tune the decision boundary (exploitation phase), which means that an selection strategy has to find a trade-off between exploration and exploitation in order to train a classifier efficiently and effectively. Thus, our selection strategy 4DS is based on the two hypotheses: (1) A selection strategy has to consider various aspects and, thus, has to combine several criteria. (2) In different phases of the AL process, these criteria must be weighted differently. 

Consequently, 4DS considers the following four criteria:

\begin{itemize}
	\item The first criterion, the \textit{distance} of samples to the current decision boundary, corresponds to the idea to choose samples with high uncertainty concerning their correct class assignment. Here, for the active training of an SVM its decision function is used:
	\begin{equation}
		\text{distance}(\mathbf{x})= \left\vert \sum_{n=1}^{\left\vert L \right\vert} \alpha_{n}c_{n}K(\mathbf{x}_n,\mathbf{x}) + b \right\vert,
	\end{equation} with Lagrange coefficients $\alpha_{n} \in \R$, bias $b \in \R$, kernel function $\R^{D}: \R^{D} \times X \rightarrow \R^{+}$ and classes $c \in \left\{1, \dots, C\right\}$. Here, the set $L$ contains all actively labeled samples so far. That is, the nearer $\mathbf{x}$ is to the decision boundary, the more this distance criterion tends to zero. For generative classifiers this criterion is estimated using class posteriors for $\mathbf{x}$. For this, the \textit{entropy} \cite{DE95}, the \textit{smallest-margin} \cite{SDW01}, or the \textit{least coefficient} \cite{CM05} approaches can be used. This criterion is needed to fine-tune the decision boundary of the classifier and must, therefore, emphasized at later stages of the AL process.  
	\item The second criterion, the \textit{density} of regions where samples are selected, is simply calculated as follows:
	\begin{equation}
		\text{density}(\mathbf{x})=p(\mathbf{x}),
	\end{equation} 
	where $p(\mathbf{x})$ is the density model that we use to gather structure information (cf.~Sections \ref{subsec:capturing} and \ref{subsec:refinement}). Samples with higher likelihood values are preferred. This criterion is used to avoid the selection of outliers and to explore ``important'' regions of the input space that may be misclassified if they are neglected. In many applications, this criterion is important at early stages of the AL process.
	\item The third criterion, the (class) \textit{distribution} of samples, makes use of responsibility information to consider all model components of $p(\mathbf{x})$ according to their mixing coefficients and, implicitly, the unknown ``true'' class distribution of all samples (i.e., class priors) for  sample selection. For that purpose, each unlabeled sample is added temporarily to the set of all samples actively selected so far (this set corresponds to the union of the set of all labeled samples so far and the current query set $S$) and then the deviation between the responsibility distribution of this set and the distribution of the mixing coefficients is calculated as follows:
	\begin{equation}
		\text{distribution}(S \cup \left\{\mathbf{x}\right\}) = 1- \sum_{j=1}^{J} \left\{\text{max}\left\{0,\pi_j-\left(\frac{1}{|L|+|S|+1} \sum_{\mathbf{x}' \in L \cup S \cup \left\{\mathbf{x}\right\}} \rho_{\mathbf{x}',j}\right)\right\}\right\}
	\end{equation}
	Samples with high values are preferred as those samples highly contribute to reach the goal.
	\item The fourth criterion, the \textit{diversity} of samples in the query set, is needed to offer the possibility to select more than one sample in each query round. Otherwise samples are selected that can be regarded as being redundant from the viewpoint of the classifier training. The computation of the diversity criterion is based on the approach described in \cite{DRH06} but we estimate the empirical entropy with help of a parametric estimation approach for $p(\mathbf{x})$ instead of a Parzen window estimate:
	\begin{equation}
		\text{diversity}(S \cup \left\{\mathbf{x}\right\}) = -\frac{1}{|S|+1} \sum_{\mathbf{x}' \in S \cup \left\{\mathbf{x}\right\}} \text{log}(p(\mathbf{x}')).
	\end{equation}
	Note that the current query set $S$ is always non-empty here as we do not consider this measure for the selection of the first sample in each query round. 
\end{itemize}
These criteria can be weighted individually in a linear combination. However, 4DS uses a self-adaptation scheme to determine the weights of the first three criteria depending on the learning performance of the actively trained classifier in each AL round. Merely, the weight for the diversity criterion has to be set by the user. But if the selection of more than one sample per AL round is not necessary, 4DS can be regarded as parameter-free. 4DS adapts the weight according to the following idea: On the one hand 4DS focusses on the class distribution measure in the initial rounds of an AL process to explore the input space of the classifier. On the other hand, if the classification performance of the classifier deteriorates, the density criterion is favored for choosing representative samples. Otherwise the distance criterion
 is favored for fine-tuning the current decision boundary, and vice versa. For more details see \cite{RS13}. 
 
\subsection{Extended Pool-based Active Learning Cycle}
\label{subsec:extendedPAL}

The traditional pool-based AL (PAL) typically starts with a large pool $\textit{U}_0$ of unlabeled samples and a small set of labeled samples $\textit{L}_0$ (with $|\textit{L}_0| \ll |\textit{U}_0|$ and $X= \textit{U}_0 \cup \textit{L}_0$), and a \textit{classifier} $\mathbf{G}_0$ is trained initially based on $\textit{L}_0$. Then, in each query round $i$ a \textit{query set} $\textit{S}_i$ of unlabeled, informative samples is determined by means of a \textit{selection strategy} $\mathcal{Q}$, which takes into account the ``knowledge'' contained in $\mathbf{G}_i$, and presented to an \textit{oracle} $\mathcal{O}$ (or a human expert) in order to be labeled. Then, $S_{i,\textit{labeled}}$ is added to $\textit{L}_i$, removed from $\textit{U}_i$, and $\mathbf{G}_i$ is updated. If a given stopping criterion is met, PAL stops, otherwise the next query round $i+1$ starts.

To apply PAL successfully, it is necessary to choose the selection strategy $\mathcal{Q}$ carefully. Depending on this choice the initially labeled training data set $\textit{L}_0$ has also a decisive impact on the learning performance of the actively trained classifier. In general, the following rule can be applied: The ``simpler'' the selection strategy, the more labeled samples well distributed in the input space of the classifier have to be available at the beginning of the AL process, because these samples can be regarded as the ``initial knowledge'' contained in classifier $\mathbf{G}_0$. But in the literature of PAL the appropriate selection of an initial training data set $\textit{L}_0$ is largely ignored \cite{HND10}.

These challenges can be met if we consider structure information for the active sample selection and for the determination of the initially labeled samples. In the simplest case, structure information can be gathered with any kind of clustering technique and samples close to cluster centers are a good choice for the initial training set $\textit{L}_0$, for example. However, for real applications, AL should start ``from scratch'', i.e.,~without any initially labeled data. Therefore, we extend the traditional PAL cycle in the following ways: (1) The data structure is gathered with a probabilistic, generative mixture model $\mathbf{M}_{0}$, that is estimated based on $\textit{U}_0$ in an unsupervised way. (2) The initial set $\textit{L}_0$ is empty and, therefore, we determined in the first query round $i=1$ (\textit{initialization round}) the first labeled set $L_1$ with a \textit{density-based strategy}, that only considers structure information contained in $\mathbf{M}_0$. (3) A \textit{transductive learner} is used, that revises the generative model $\mathbf{M}_i$ in each cycle $i$ based on the class information of $\textit{L}_i$ (cf.~Section~\ref{subsec:refinement}).

Algorithm \ref{alg:poolBasedActiveWithSharedSeparateComponentModels} sketches our extended PAL approach and Algorithm \ref{alg:densitybasedstrategy} the density-based strategy, used to determine $\textit{S}_1$ in the initialization
 round. 

\begin{algorithm}[htb!]
\let\oldnl\nl
\newcommand{\nonl}{\renewcommand{\nl}{\let\nl\oldnl}}
\caption{Extended Pool-Based Active Learning}
\label{alg:poolBasedActiveWithSharedSeparateComponentModels}
\DontPrintSemicolon
\LinesNumbered
{

\LinesNumberedHidden
\medskip
\fs{9}{11}
\SetKwInOut{Input}{input}
\SetKwInOut{Output}{output}
\Input{$\cal O$ 						      \hspace{0.32cm}\tcp*[l]{\fs{7}{9}oracle, e.g., a human domain expert}\\
  	    \hspace{0.1cm}$n$ 				    \hspace{0.29cm}\tcp*[l]{\fs{7}{9}maximum   number of query samples (our stopping criterion, could be replaced)}\\
  	    \hspace{0.1cm}$k$ 				    \hspace{0.3cm}\tcp*[l]{\fs{7}{9}size of the query set $\mathit{S}_i$ (in each learning cycle $i>0$; $k \leq n$)}\\
  	    \hspace{0.1cm}$m$ 				    \hspace{0.23cm}\tcp*[l]{\fs{7}{9}size of the query set $\mathit{S}_0$ (initialization round; $m \leq n$)}\\
		\hspace{0.1cm}$U_0$ 			    \hspace{0.10cm}\tcp*[l]{\fs{7}{9}initial unlabeled pool with $|U_0| \geq n$}\\
		\hspace{0.1cm}$L_0$ 			    \hspace{0.15cm}\tcp*[l]{\fs{7}{9}initial labeled data set is $\emptyset$ here}\\
		\hspace{0.1cm}$\mathbf{M}_{0}$	\hspace{0.03cm}\tcp*[l]{\fs{7}{9}initial \CMMsha trained in an unsupervised way with VI using $U_0$}\\
		\hspace{0.1cm}$\cal Q$ 			  \hspace{0.22cm}\tcp*[l]{\fs{7}{9}selection strategy (e.g., 4DS or another strategy)}\\
}
\medskip
\Output{$\mathbf{G}$ \hspace{0.31cm}\tcp*[l]{\fs{7}{9}actively trained classifier (e.g., SVM or another classifier) in the final PAL step}}
\medskip
}
\setcounter{AlgoLine}{0}
\fs{9}{11}
	$i \leftarrow 0$\;
	\Repeat{$|\textit{L}_{i}| = n$}{
		$i \leftarrow i+1$\;
		\eIf{initialization round}{
	  select query set $\mathit{S}_1$ $\subset \mathit{U}_0$ according to density based strategy using $\mathbf{M}_0$ (cf. Alg.~\ref{alg:densitybasedstrategy})\;
  }{
	select query set $\mathit{S}_i$ $\subseteq \mathit{U}_{i-1}$ according to selection strategy $\cal Q$ using $\mathbf{G}_{i-1}$\;
  }
   		build $\textit{S}_{\textit{i, labeled}}$ asking $\cal O$ for class labels of $\mathit{S}_i$ \;
	  	build $\mathit{L}_{i}$ by adding set $\textit{S}_{\textit{i, labeled}}$ to training set $\mathit{L}_{i-1}$\;
	  	build $\mathit{U}_{i}$ by removing set $\textit{S}_{\textit{i, labeled}}$ from pool $\mathit{U}_{i-1}$\;
		build $\mathbf{M}_{\textit{i}}$ by revising $\mathbf{M}_{\textit{i}-1}$ based on $\mathit{L}_{\textit{i}}$ with the \textit{transductive learner} (see \cite{RCS14})\;
		train $\mathbf{G}_i$ using $\textit{L}_i $ and $ \textit{M}_{\textit{i}}$\;
		}
\Return{$\mathbf{G}_i$}

\end{algorithm}

\begin{algorithm}[htb!]
\caption{Density-Based Selection Strategy}
\label{alg:densitybasedstrategy}
\DontPrintSemicolon

\LinesNumberedHidden
{
\medskip
\fs{9}{11}
\SetKwInOut{Input}{input}
\SetKwInOut{Output}{output}
\Input{\hspace{0.1cm}$m$ 				    \hspace{0.25cm}\tcp*[l]{\fs{7}{9}size of the query set $\mathit{S}$}\\
		\hspace{0.1cm}$U_0$ 			    \hspace{0.10cm}\tcp*[l]{\fs{7}{9}initially unlabeled pool with $|U_0| \geq m$}\\		
		\hspace{0.1cm}$\mathbf{M}_{0}$	\hspace{0.03cm}\tcp*[l]{\fs{7}{9}mixture density model with $J$ components}\\
		}
\medskip
\Output{$\mathit{S}$ \hspace{0.4cm}\tcp*[l]{\fs{7}{9}query set for labeling}}
\medskip
}
\LinesNumbered
\setcounter{AlgoLine}{0}
\fs{9}{11}
	$\textit{S} \leftarrow \emptyset$ \tcp*[l]{\fs{7}{9} initialize the query set}
	$\textit{V} \leftarrow \emptyset$ \tcp*[l]{\fs{7}{9} index set of the visited model components}
	\Repeat{$|\textit{S}| = m$}{
	r $\leftarrow$ new integer random value $\in \left[0,J\right]$\;
\uIf{$r \notin V$}{
	$V \leftarrow V \cup \{r\}$ \tcp*[l]{\fs{7}{9} mark component $r$ as visited by adding $r$ to $V$}
   	build a ranking of all samples $\mathbf{x} \in \textit{U}_{0}$ according to their likelihood values $p(\mathbf{x}|r)$\;
   	select the sample $\mathbf{x}'$ randomly from the top $10 \%$ highest ranked samples\;
   	$S \leftarrow S \cup \{\mathbf{x}'\}$ \tcp*[l]{\fs{7}{9} add sample $\mathbf{x}'$ to the query set}

   } \ElseIf {$|V|= J$}{
   $\textit{V} \leftarrow \emptyset$ \tcp*[l]{\fs{7}{9} reset the index set of the visited components}
   } 
  }
\Return{$\mathit{S}$}
\end{algorithm}

\section{Simulation Experiments}
\label{sec:results}
In this section we evaluate the experiments performed on 20 data sets and compare our new approach to integrate structure information into the AL training of discriminative classifiers such as SVM to AL approaches that capture this information in a different way (LapSVM) or neglect it. First, this section describes the setup of the experiments. Second, we visualize the behavior of an actively trained SVM that recognizes structure information with help of the RWM kernel in comparison to an SVM with RBF kernel that does not use this information. For this, we use two data sets with two-dimensional input spaces and apply uncertainty sampling (US) in order to outline the impact of the different kernels on the AL behavior of the SVM. Third, simulation experiments are performed on 20 benchmark data sets to compare our new approach to related techniques numerically and in some more detail. 

\subsection{Setup of the Experiments}
\label{sec:setup}
In this section, we describe the classifier paradigms and the selection strategies that are used in our simulation experiments. We sketch the main characteristics of the data sets and define our evaluation criteria.

\subsubsection{Classifiers}
\label{subsec:classifiers}
 In our experiments we compare actively trained SVM with different data-dependent kernels -- RWM, GMM, and LAP kernels -- and data-independent kernels, such as the RBF kernel. For the SVM with LAP kernel (also called LapSVM or Laplacian SVM), we ported the MATLAB implementation of Melacci \cite{Melacci12} to Java and adapted it to cope with multi-class problems.  

To find good estimates (in case of the RWM and GMM kernels) for the hyper-parameters of the VI algorithm (training of the mixture density models capturing structure information in unlabeled data) we used an exhaustive search on the unlabeled training data. To rate a considered set of VI parameters we applied an interestingness measure, called \textit{representativity} \cite{FKS11}. It measures the dissimilarity of the mixture density model trained with VI and a density estimate resulting from a non-parametric Parzen window estimation. As dissimilarity measure we used the symmetric Kullback-Leiber divergence instead of the Hellinger distance mentioned in~\cite{FKS11}.

 To get good parametrization results regarding the SVM (with corresponding kernel) we performed for each fold of the (outer) 5-fold cross-validation an inner 4-fold cross-validation on the labeled set $L_{1}$ (after execution of the initialization round). That is, the parameters of the SVM are determined only once during the AL process, whereby with the inner cross-validation the set $L_{1}$ is split into a validation set $L_{val}$ (one quarter) and a training set $L_{train}$ (three quarters). 
 To rate a considered parameter combination we determined the classification performance of the SVM by considering $L_{val}$ and $U_{1}$ (the expected error) simultaneously.
 However, at the beginning of the AL process we used the whole training set $L=U_0 = L_{val} \cup L_{train} \cup U_{1}$ (i.e., without class assignments) for capturing structure information. That is, all samples in the set $L$ are used to determine the Laplacian graph in case of the LAP kernel and to determine the mixture model in case of the RWM and GMM kernels. Overall, the test data (set $T$) of the (outer) cross-validation is never used for any optimization or parametrization purposes.

\begin{figure}[htbp!]
	\begin{center}
	\subfigure[Outer cross-validation (one fold).]{\label{fig:subfig1stCrossValidation}\includegraphics[width=0.46\textwidth]{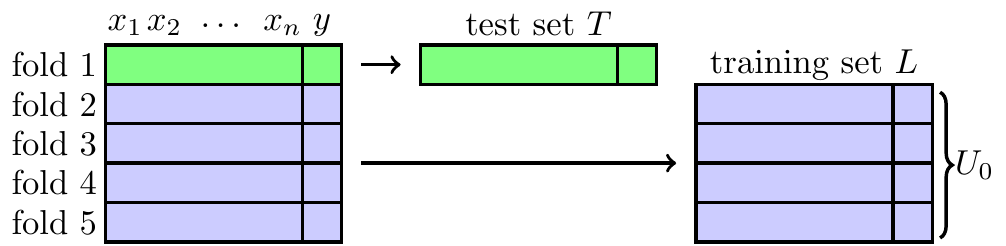}}
	\quad	
	\subfigure[Inner cross-validation (one fold).]{\label{fig:subfig2ndCrossValidation}\includegraphics[width=0.47\textwidth]{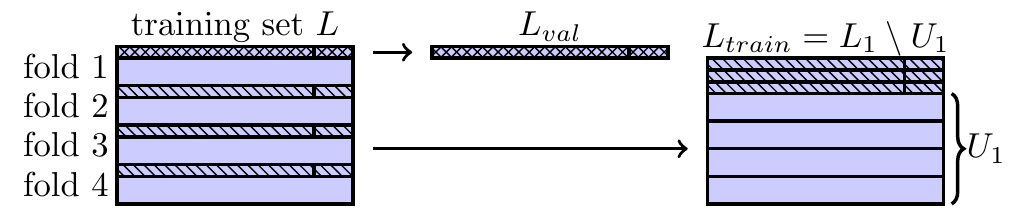}}
	\caption{Disjoint subsets of one fold of the (outer) 5-fold cross-validation. $L_{val} \cup L_{train}$ correspond to a non-randomly chosen subset from the ``training'' folds (training set $L$). The remaining samples of $L$ build the subset $U$.}
	\label{fig:crossValidation}
	\end{center}
\end{figure} 

The penalty parameter $C = 10^i$ and the kernel width $\gamma = 10^i$ were varied for $i \in \{-3,-2, \dots, 2\}$, the four additional parameters of the LAP kernel $\gamma_{I} =10^{\textit{j}}$ and $\gamma_{A} = 10^{\textit{j}}$ for $\textit{j} \in \{-5,-6, \dots, 2\}$, the neighborhood size was fixed to $k=7$ and the degree to $p=1$. To account for information from categorical input dimensions we adapted all kernels in the same way as the RWM kernel (described in Section \ref{subsec:exploitstructure}). Therefore, to find the best values ​​of $\alpha$ (weighting factor of continuous input dimensions) and $\beta$ (weighting factor of discrete input dimensions) we varied $ \alpha$ and $\beta $ from $0$ to $1$ in step sizes of $0.1$ (for the data sets Australian, Credit A, Credit G, Heart, and Pima that have categorical attributes, cf.~Table \ref{tab:DataSetsBasic}). 

\subsubsection{Selection Strategies}
\label{subsec:strategies}
For selecting informative samples we used three different selection strategies in our suite of AL experiments: \textit{random sampling} (Random), \textit{uncertainty sampling} (US) and our \textit{4DS} (selection strategy based on \textit{distance}, \textit{density} (data), \textit{diversity}, and \textit{distribution} (class) information). The latter strategy takes structure information into account for sample selection, whereas the other ones do not consider this information: Random sampling chooses samples randomly with uniform selection probability. It was shown in~\cite{Cawlay11} that this trivial approach outperforms some existing, at the first glance more sophisticated selection strategies. The US strategy \cite{TK02} queries the sample for which the SVM is currently most uncertain concerning its class assignment, i.e.,~the sample with the smallest distance to the current decision boundary. 4DS \cite{RS13} considers the \textit{distance} of samples to the decision boundary, too, and additionally the \textit{density} in regions where samples are selected. Furthermore it indirectly considers the unknown \textit{class distribution} of the samples by utilizing the responsibilities of the model components for these samples and the \textit{diversity} of samples in the query set that are chosen for labeling. The combination of the four measures in 4DS is self-optimizing since the weights for the first three measures depend on the performance of the actively trained classifier and only the weight parameter $\lambda$ for the last measure (diversity) has to be chosen by the user. Here, we vary $\lambda$ from zero to one in steps of $0.1$ and choose the best $\lambda$ according to a $5$-fold cross-validation on the training data.

In addition it should be mentioned that for a fair comparison the classifiers will be retrained after each query of one sample for the strategies Random and US and after each query of five samples for the strategy 4DS. Doing so, 4DS has to select samples based on ``outdated'' information in each query, thus the comparison is certainly not biased towards 4DS. Moreover, in \cite{RS13} it has been shown that 4DS outperforms, based on various evaluation criteria, related selection strategies such as ITDS (information theoretic diversity sampling), DWUS (density weighted uncertainty sampling), DUAL (dual strategy for active learning), PBAC (prototype based active learning), and 3DS (a technique we proposed earlier in~\cite{RS11}) by training a generative classifier (\CMMsha) actively. 

\subsubsection{Data Sets}
\label{subsec:datasets}

For our experiments, we use the MNIST data set \cite{MNIST15} and 20 benchmark data sets: 14 real-world data sets (Australian, Credit A, Credit G, Ecoli, Glass, Heart, Iris, Page Blocks, Pima, Seeds, Vehicle, Vowel, Wine, and Yeast) from the UCI Machine Learning Repository \cite{AN07}, two real-world (Phoneme and Satimage) and two artificial data sets (Clouds and Concentric) from the UCL Machine Learning Group \cite{UCL14}, and two artificial data sets, Ripley suggested in \cite{Ripley96} and Two Moons suggested in \cite{Melacci12}. In order to obtain meaningful results regarding the performance of our new approach, we consider three requirements for the selection of the data sets: First, the majority of the data sets should come from real life applications. Second, the data sets should have very different numbers of classes. And third, some of the data sets should have unbalanced class distributions. The description of the benchmark data sets and the MINST data set is summarized in Table \ref{tab:DataSetsBasic}.

\begin{table*}[htbp!]
\centering
\scriptsize
\renewcommand{\arraystretch}{0.8}
\renewcommand{\tabcolsep}{14pt}
\caption[Data set information]{General data set information.}
\label{tab:DataSetsBasic}
\begin{tabular}{l l l l l l }
\toprule
\multirow{3}{*}{Data Set} & \multicolumn{5}{c}{Description}\\
\cmidrule[0.5pt]{2-6}
&  Number\ of & Continuous & Categorical & Number\ of & Class \\
&  Samples &  Attributes\ & Attributes & Classes & Distribution (in \%)\\
\midrule
Australian 	&690		&6		&8		&2	& 55.5,44.5\\
Clouds 			&5000		&2		&--		&2	& 52.2,50.0\\
Concentric    &2500      &2     &–-    & 2& 36.8, 63.2\\
Credit A	   &690		   &6		&9		&2	& 44.5,55.5\\
Credit G 		&1000		&7		&13		&2	&70.0,30.0\\
Ecoli			&336		&7		&--		&8 &42.6,22.9,15.5,10.4,5.9,1.5,0.6,0.6\\
Glass			&214		&9		&--		&6&32.7,35.5,7.9,6.1,4.2,13.6\\
Heart			&270		&6		&7		&2 & 44.4,55.6\\
Iris         & 150      &4      & --     & 3 & 33.3,33.3,33.3 \\
Page Blocks     &5473   &10     & --    & 5 &89.8,6.0,0.5,1.6,2.1\\
Phoneme			&5404		&5		&--		&2	& 70.7,29.3\\
Pima 			&768		&--		&8		&2	& 65.0,35.0\\
Ripley			&1250		&2		&--		&2	&50.0,50.0\\
Satimage 		&6345		&5		&--		&6	&24.1,11.1,20.3,9.7,11.1,23.7\\
Seeds         & 210       & 7    & -- & 3 & 33.3,33.3,33.3 \\
Two Moons 		&800		&2	    &--		&2	&50,50\\
Vehicle			&846		&18		&--		&4	&23.5,25.7,25.8,25.0\\
Vowel			&528		&10		&--		&11&9.1,9.1,9.1,9.1,9.1,9.1,9.1,9.1,9.1,9.1,9.1\\
Wine 			&178		&13		&--		&3&33.1,39.8,26.9\\
Yeast & 1484 & 8 & -- & 10 & 16.4,28.1,31.2,2.9,2.3,3.4,10.1,2.0,1.3,0.3 \\
MNIST & 70000 & 784 & -- & 10 & 10.0,10.0,10.0,10.0,10.0,10.0,10.0,10.0,10.0,10.0\\
\bottomrule
\end{tabular}
\end{table*}

In our experiments regarding the 20 benchmark data sets, we performed a $z$-score normalization for all data sets and conducted a stratified 5-fold cross-validation evaluation, as sketched in Fig.~\ref{fig:crossValidation}. In each round of the outer cross-validation, one subset is kept out as \textit{test set}~$T$. Of course, $T$ is not considered for any parametrization purposes. The other four subsets build the pool~$U_{0}$ of unlabeled samples (cf.~Fig.~\ref{fig:subfig1stCrossValidation}), such that our AL approach starts with an empty training set~$L_{0}$. Consequently, in the first query round an SVM classifier is not given and any ``knowledge'' cannot be considered for sample selection. Therefore, in the first AL round (initialization) we select a number of samples equal to $4\times \text{the number of classes}$ with help of a density based approach that uses only structure information captured in the mixture density model. Based on the class information for these samples, that build the set~$L_{1}$, the parameters of the SVM are determined with a grid search technique. This means as mentioned before that the SVM are parameterized only once during the AL process (e.g.,~for a two class problem only eight labeled samples are used for parametrization purposes). Moreover, the data splits $U_1$ and $L_1$ are chosen identically for all actively trained classifier paradigms. We decided to actively select not more than 500 samples from each data set, apart from Ecoli (270), Glass (171), Heart (216), Iris (120), Seeds (168), and Wine (142) because of their limited size. Thus, this number of labeled samples is taken as stopping criterion for the AL process.

For the experiment regarding the MNIST dataset, we additionally reduced the size of the input dimensions from $784$ to $34$ dimensions by applying a principle component analysis (PCA) and conducted the stratified 5-fold cross-validation, as described before, only on the corresponding 60000 training samples. 
Consequently, the 10000 test samples are never used for any parametrization or optimization purposes. 

\subsubsection{Evaluation Criteria}
\label{subsec:criteria}

To assess our results numerically we used three evaluation measures: \textit{ranked performance} (RP), \textit{data utilization rate} (DUR), and  \textit{area under the learning curve} (AULC) (cf.~\cite{CKS06}).

The first measure, RP, ranks the actively trained paradigms based on a non-parametric statistical Friedman test~\cite{Friedman40}. Basis for a rank is the classification accuracy on test data measured at the PAL step for which the performance on the training data is optimal. This step is not necessarily  the last PAL step when the maximum number of actively selected samples -- 140, 170, 215, 420, or, 500, respectively (see above) -- is reached. Based on these test accuracies, the Friedman test ranks -- considering a given significance value $\alpha$ -- $S$ classifiers for each of $N$ data sets separately, in the sense that the best performing classifier (highest accuracy) gets the lowest rank, a rank of 1, and the worst classifier (lowest accuracy) the highest rank, a rank of $S$. In case of ties, the Friedman test assigns averaged ranks. Let $r_{i}^{\textit{j}}$ be the rank of of the $i$-th classifier on the $\textit{j}$-th data set, then the Friedman test compares the classifiers based on the averaged ranks $R_\textit{j} = \frac{1}{N}\sum_{i=1}^{S}{r_{i}^{\textit{j}}}$. Under the null hypothesis, which claims that all classifiers are equivalent in their performance and hence their averaged ranks $R_\textit{j}$ should be equal, the Friedman statistic is distributed according to the $\chi_{F}^{2}$ distribution with $S-1$ degrees of freedom~\cite{JS11}. The Friedman test rejects the null hypothesis if Friedman's $\chi_{F}^{2}$ is greater than the $p$-value of the $\chi_{F}^{2}$ distribution. If the null hypothesis can be rejected we proceed with the Nemenyi test~\cite{Nemenyi63} as post hoc test in order to show which classifier performs significantly different. Here, the performance differences of two classifiers are significant if the corresponding average ranks differ by at least the critical difference $\text{CD} = q_{\alpha} \sqrt{\sfrac{S(S+1)}{6N}}$ where the critical value $q_{\alpha}$ is based on the Studentized range statistic divided by $\sqrt{2}$. The results of the Nemenyi test can be visualized with help of critical difference plots~\cite{Demsar06}. In these plots, non-significantly different classifiers are connected in groups (their rank difference is smaller than $\text{CD}$). To summarize the ranked performances over all data sets, the average ranks and the numbers of wins are also determined for each strategy as described in~\cite{CKS06}. The \textit{number of wins} is the number of data sets for which a paradigm performs best. Wins can be ``shared'' when different classifiers perform comparably on the same data set. That is, a good paradigm yields a low average rank and a large number of wins.

The second measure, DUR, determines the fraction of samples that must be labeled to achieve good classification results. To measure this, we first define the \textit{target accuracy} as the average accuracy achieved by a baseline strategy over five folds using between $80\%$ and $100\%$ of the maximum number of actively selected samples (see above). Here, we use the strategy US to train an SVM with RBF kernel actively as baseline. The (DUR) (cf.~\cite{CKS06}) is then the minimum number of samples needed by each of the other strategies to reach the target accuracy divided by the number of samples needed by US to train an SVM with RBF kernel actively. Simulations where a strategy does not reach the target accuracy are reported explicitly. The DUR indicates how efficiently a selection strategy uses the data, but it does not reflect detailled performance changes up to the point when the target accuracy is reached. To summarize over all data sets we again determine the mean and number of wins for each strategy. A good strategy yields a low mean DUR (in particular, it should be lower than one) and a large number of wins.

The third measure is the AULC \cite{CKS06}, again measured against the baseline strategy, which is the difference between the area under the learning curve for a given strategy and that of baseline strategy SVM with RBF kernel actively trained with US. A negative value indicates that this strategy (on average over five folds) performs worse than the baseline strategy. The AULC also is calculated for the maximum number of actively selected samples. This measure is, in contrast to the previous ones, sensitive to performance changes throughout the AL process. For example, if two strategies reach the target accuracy with the same number of samples (i.e., the same DUR), one might have a higher AULC if performance improvements occur in an earlier phase of the PAL process, for instance. To summarize over all data sets, we again determine the mean as well as the number of wins. A good strategy should have a high AULC (in particular, it should be positive) and a large number of wins.

\subsection{Behavior of actively trained SVM using Structure Information}
\label{sec:behavior}

\begin{figure}[tbp!]
\begin{center}
\subfigure[iteration $i=0$.]{\label{fig:RWM_cloudsA}\includegraphics[width=0.318\textwidth]{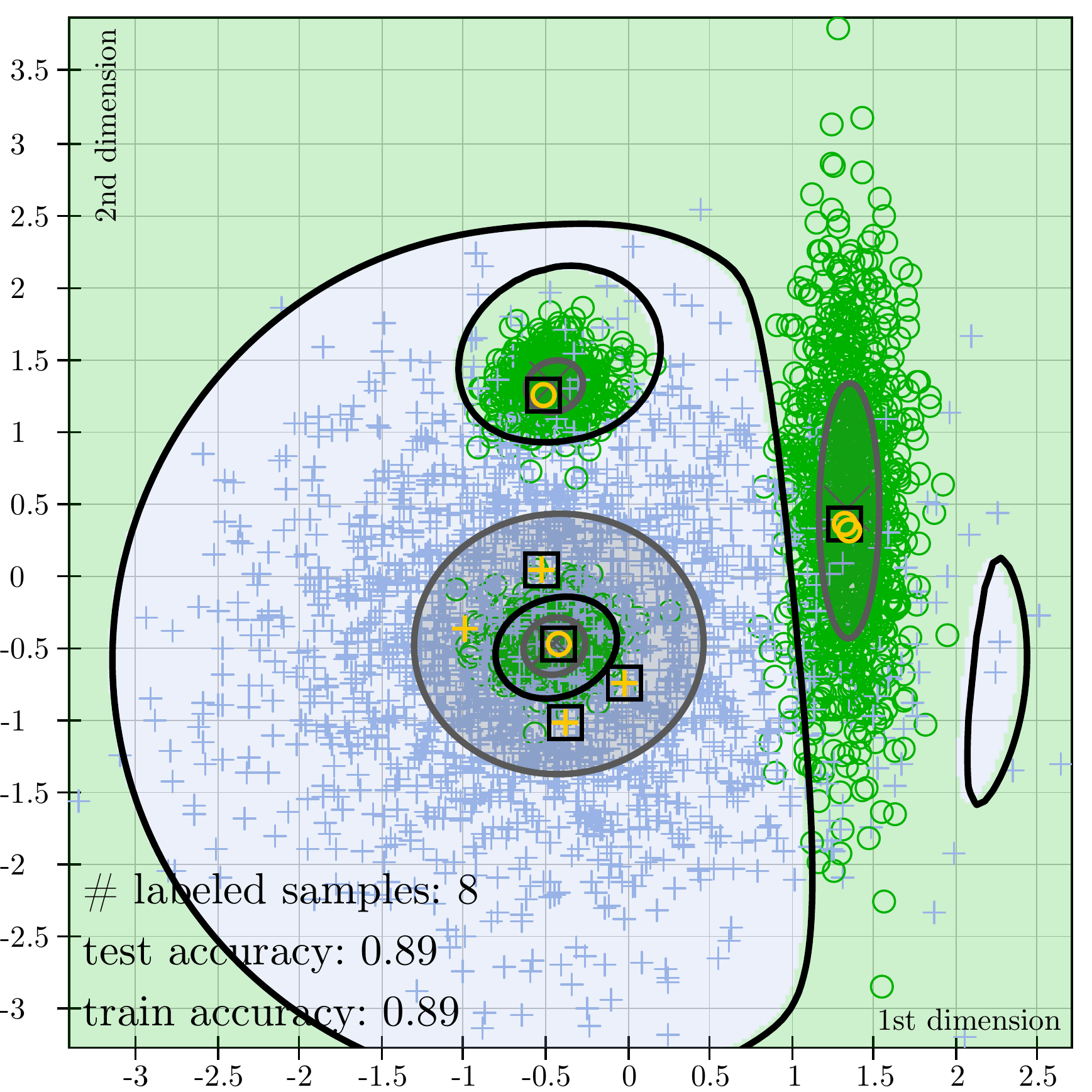}}\quad
\subfigure[iteration $i=12$.]{\label{fig:RWM_cloudsB}\includegraphics[width=0.318\textwidth]{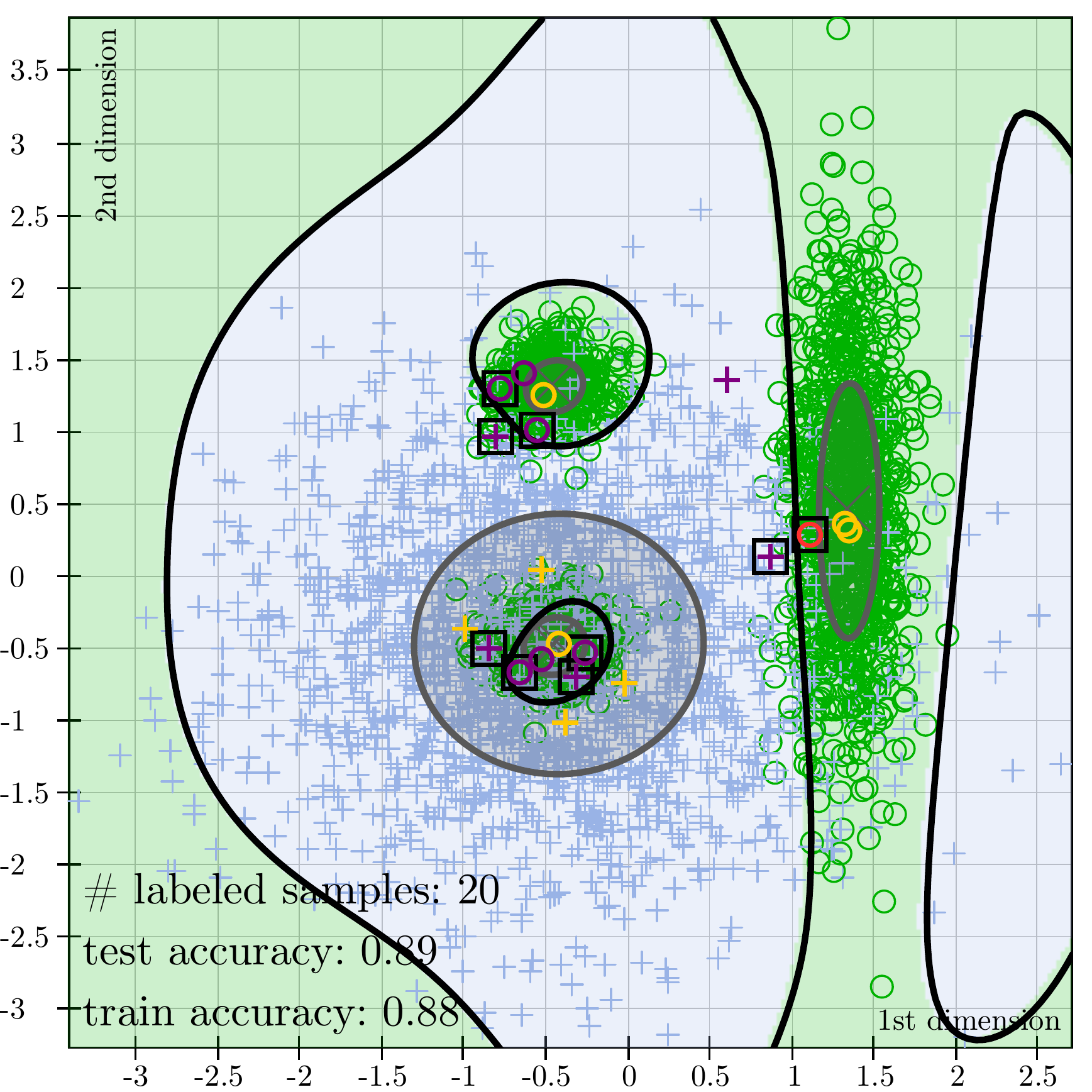}}\quad
\subfigure[iteration $i=32$.]{\label{fig:RWM_cloudsC}\includegraphics[width=0.318\textwidth]{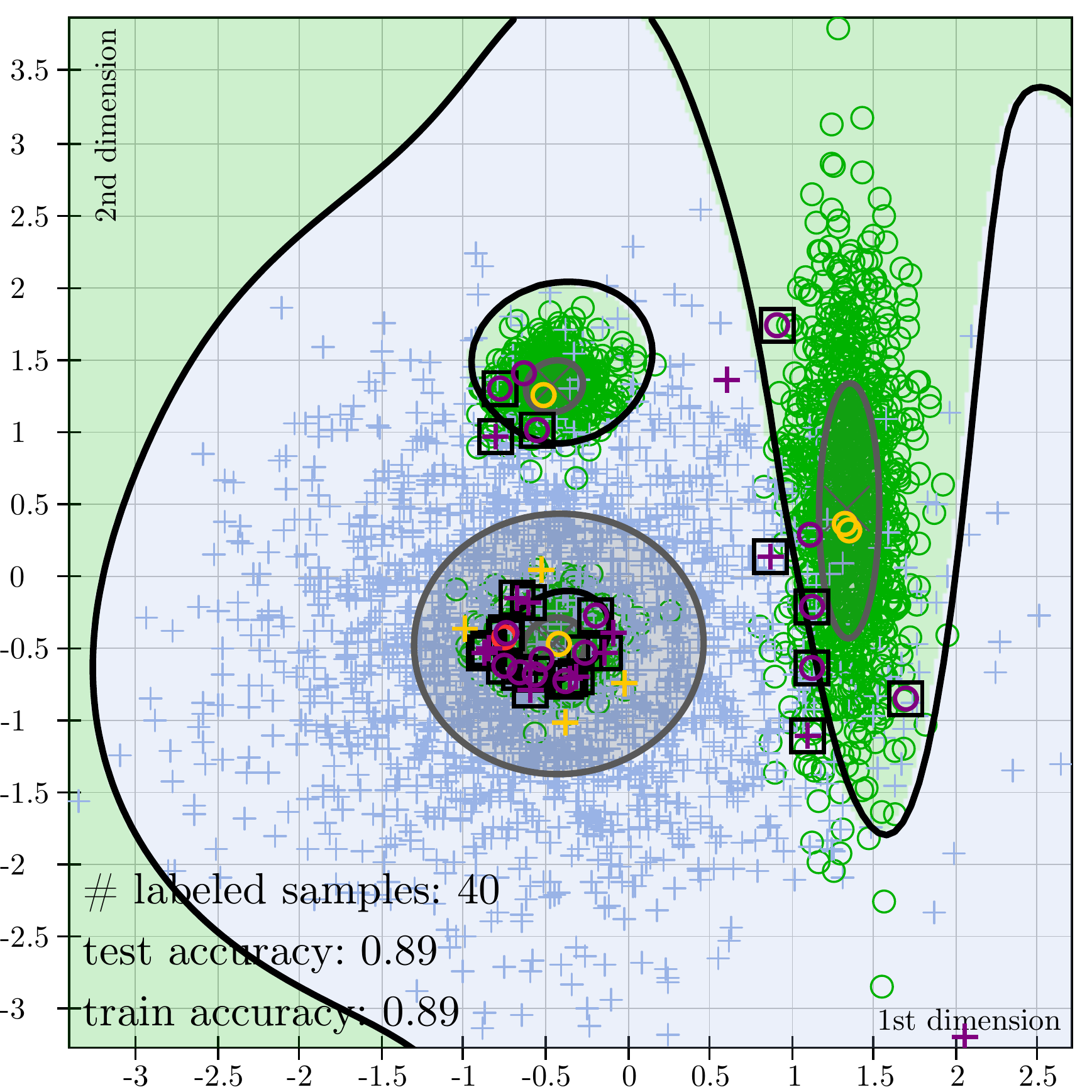}}\vfill
\subfigure[iteration $i=62$.]{\label{fig:RWM_cloudsD}\includegraphics[width=0.318\textwidth]{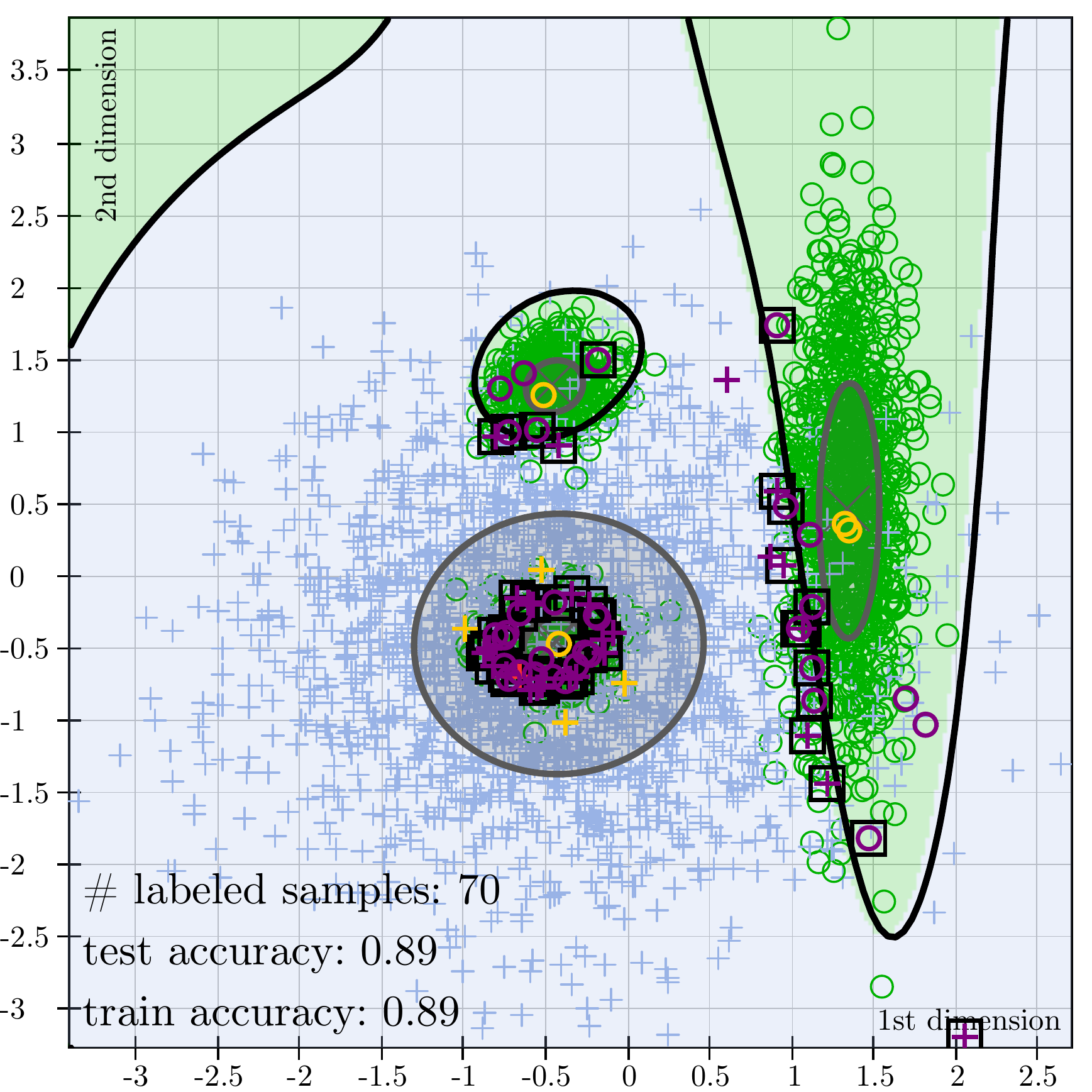}}\quad
\subfigure[iteration $i=112$.]{\label{fig:RWM_cloudsE}\includegraphics[width=0.318\textwidth]{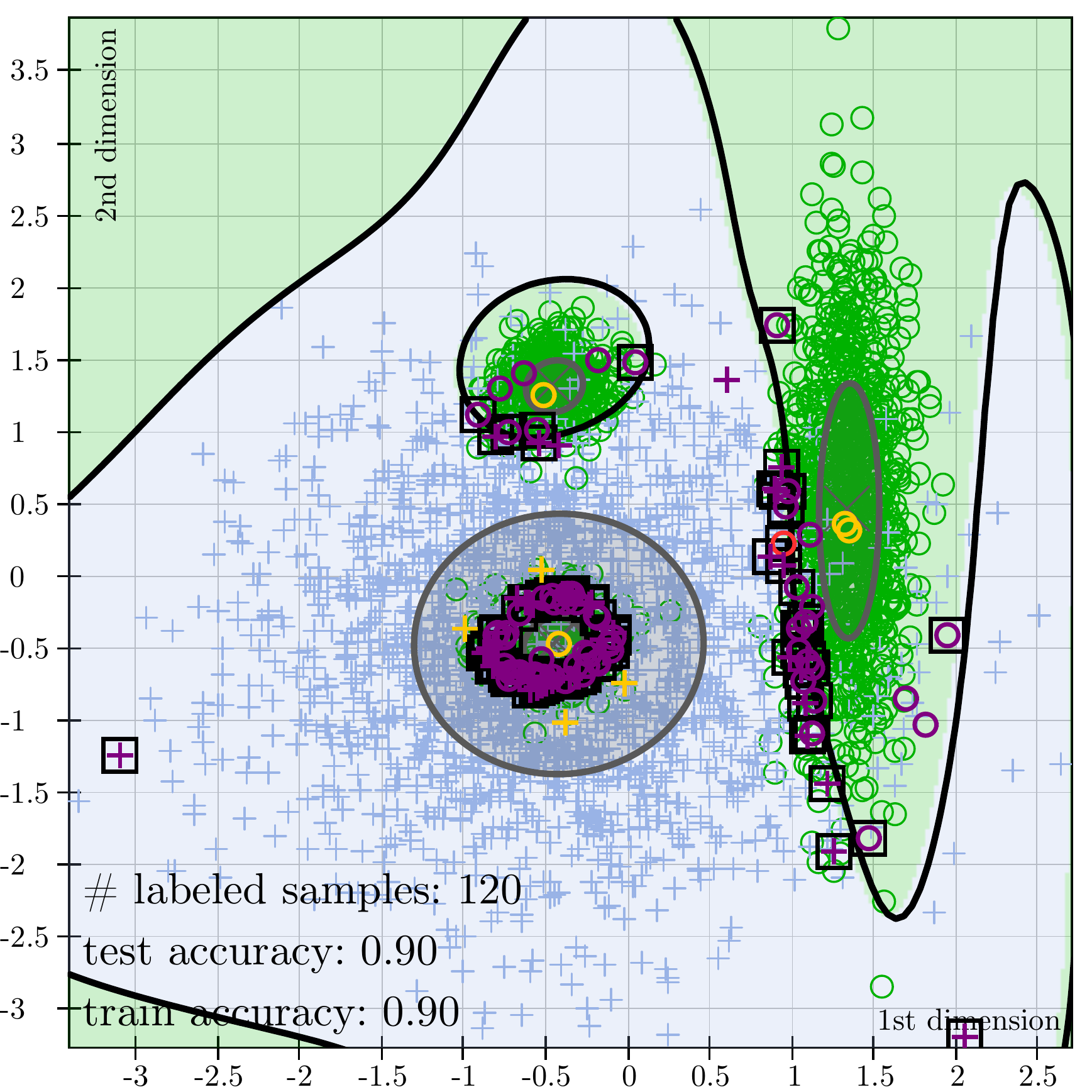}}\quad
\subfigure[classification accuracy vs. learning cycle.]{\label{fig:RWM_cloudsF}\includegraphics[width=0.318\textwidth]{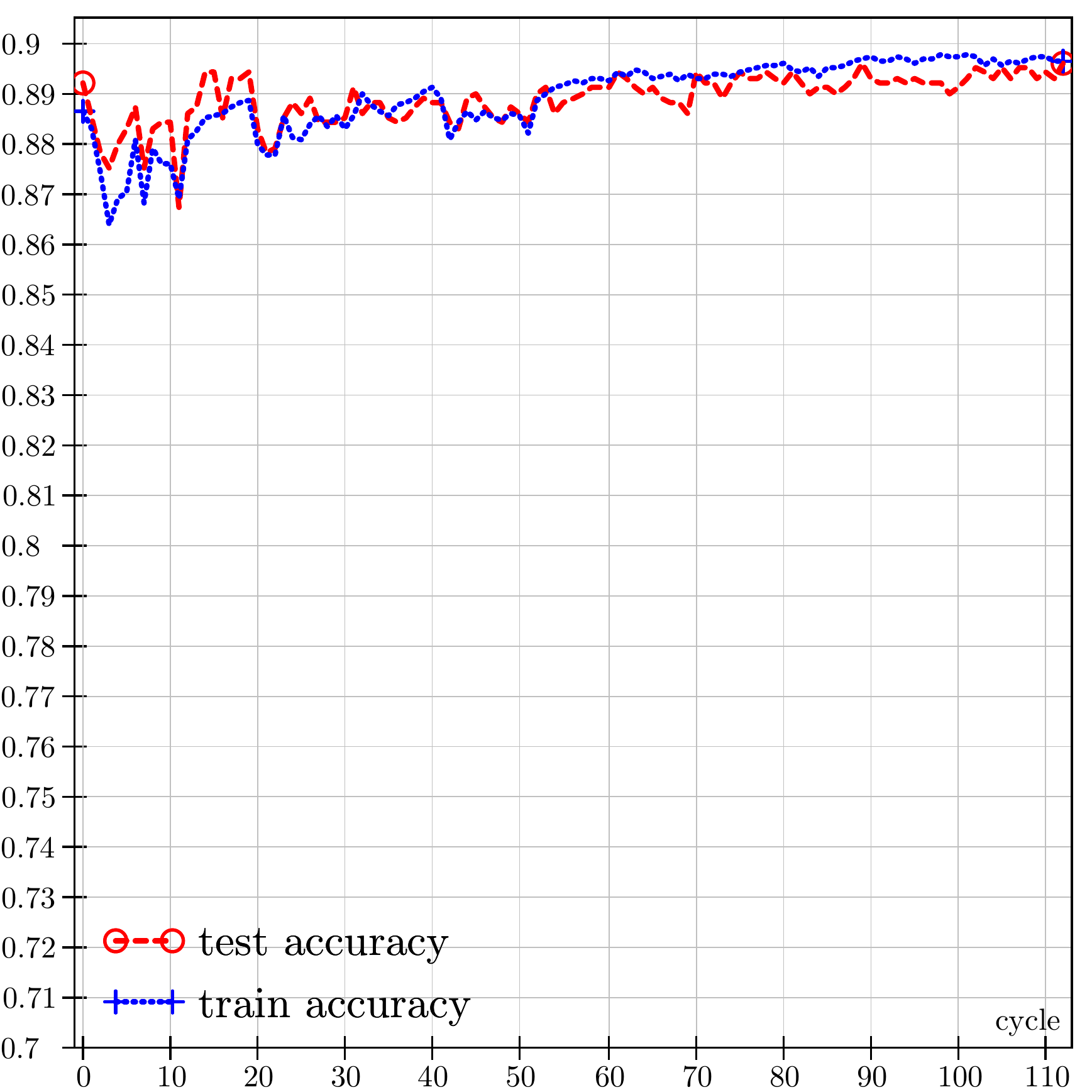}}
\caption{SVM with RWM kernel actively trained with US on the Clouds data set. In each AL round one sample was queried. Eight samples (orange colored) are initially (iteration $i=1$) chosen by a density based strategy. The labeled samples are colored red if they are selected with US in the current query round, and violet otherwise. The samples marked with an rectangle correspond to the support vectors of the SVM.}
\label{fig:RWM_clouds}
\end{center}
\end{figure}

\begin{figure}[tbp!]
\begin{center}
\subfigure[iteration $i=0$.]{\label{fig:RBF_cloudsA}\includegraphics[width=0.318\textwidth]{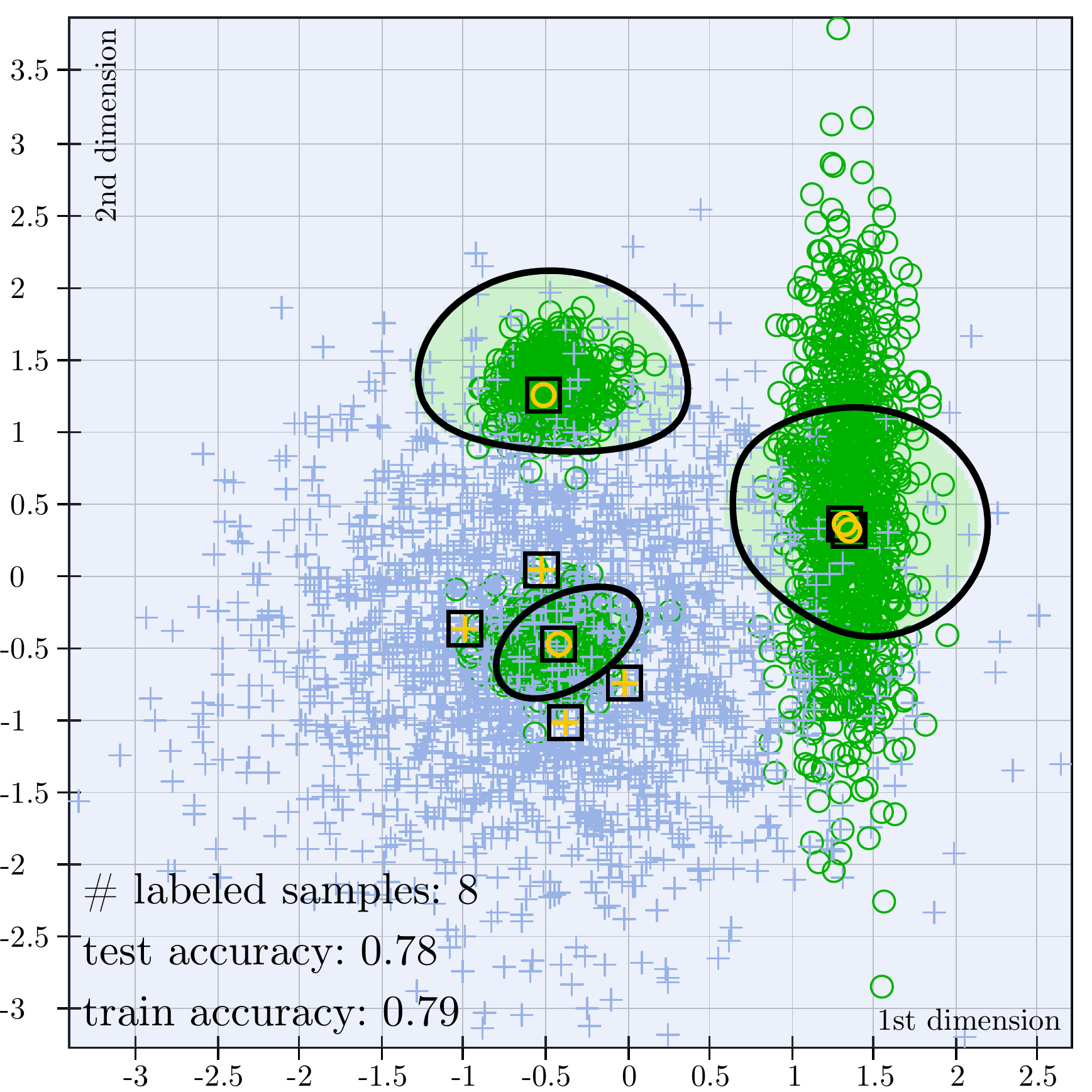}}\quad
\subfigure[iteration $i=12$.]{\label{fig:RBF_cloudsB}\includegraphics[width=0.318\textwidth]{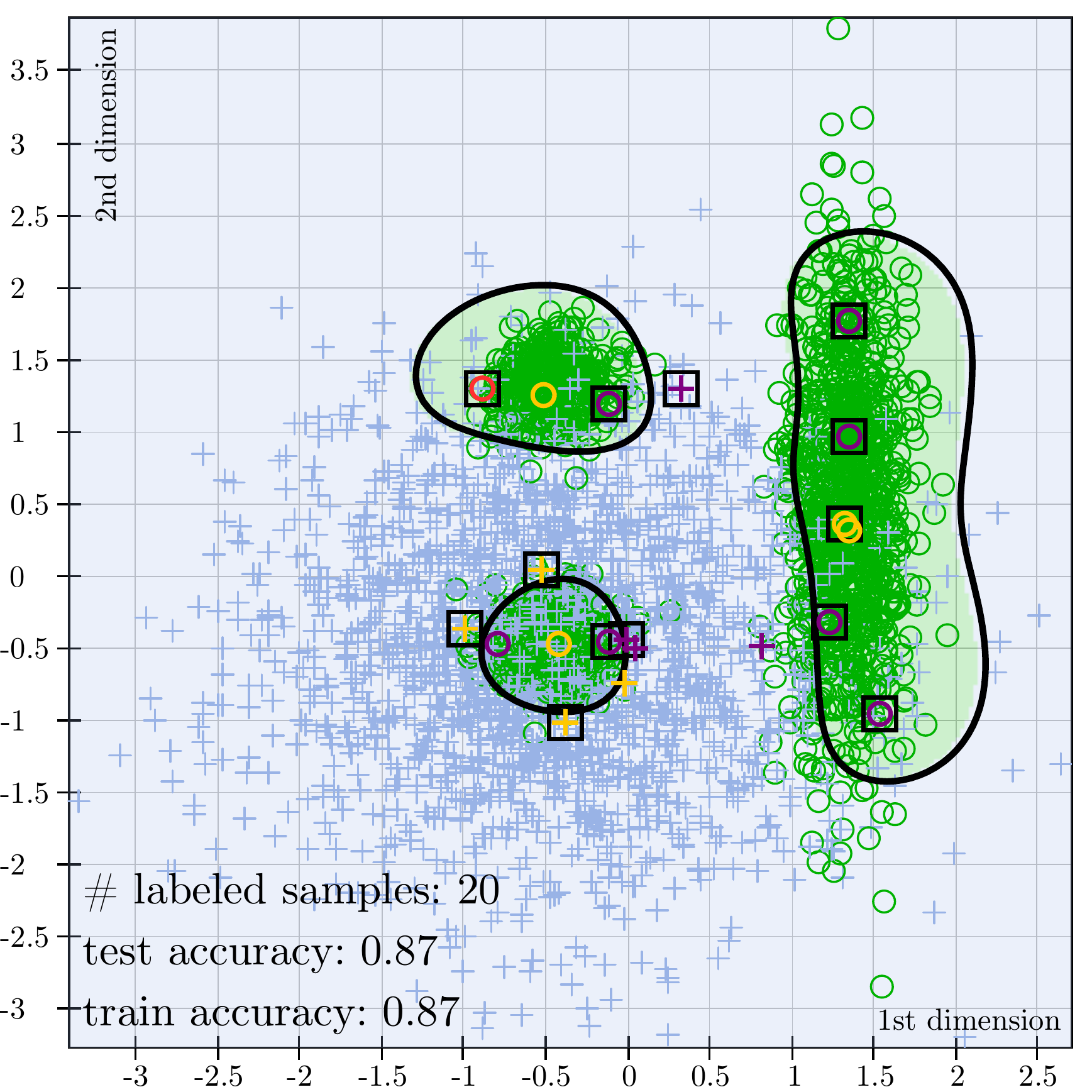}}\quad
\subfigure[iteration $i=32$.]{\label{fig:RBF_cloudsC}\includegraphics[width=0.318\textwidth]{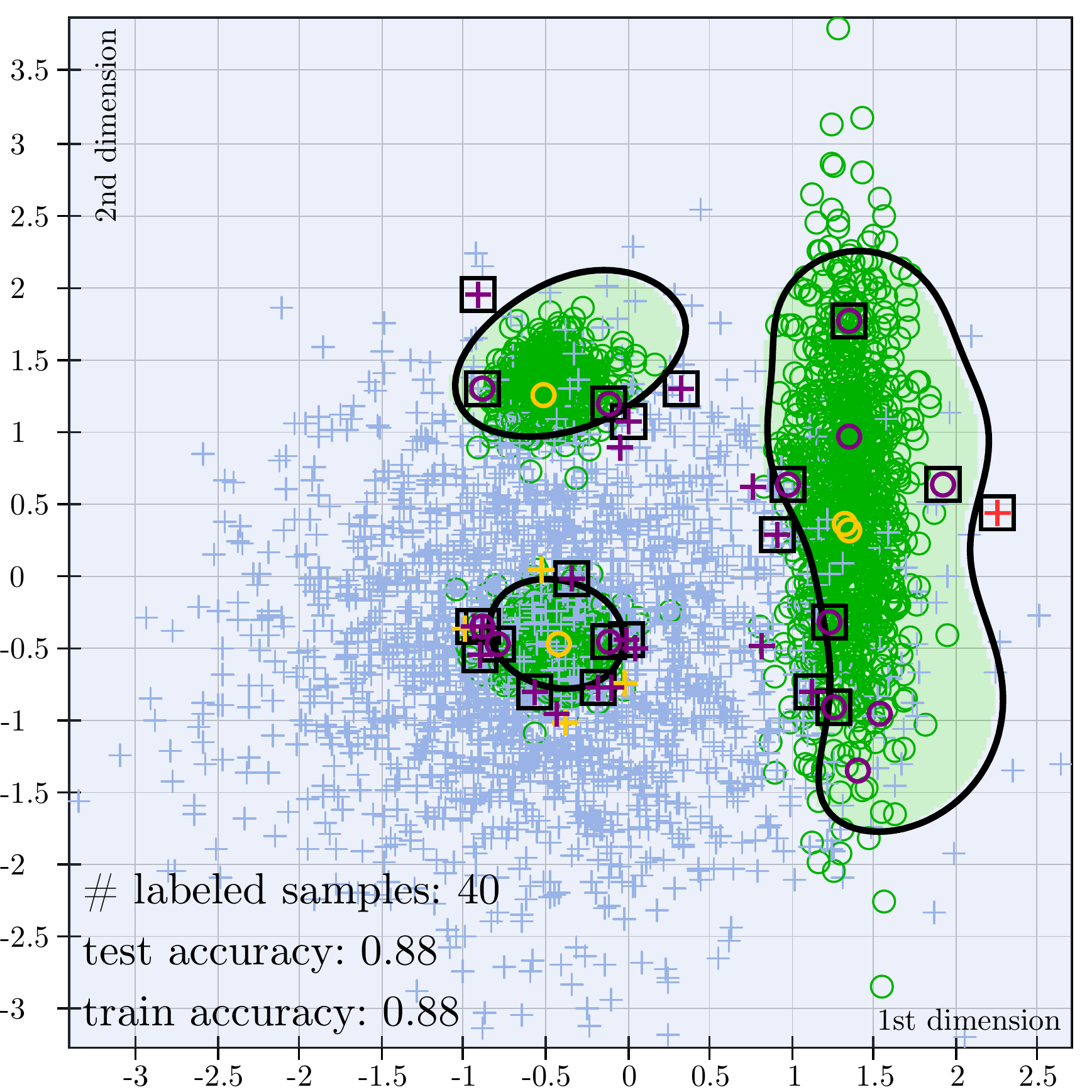}}\vfill
\subfigure[iteration $i=62$.]{\label{fig:RBF_cloudsD}\includegraphics[width=0.318\textwidth]{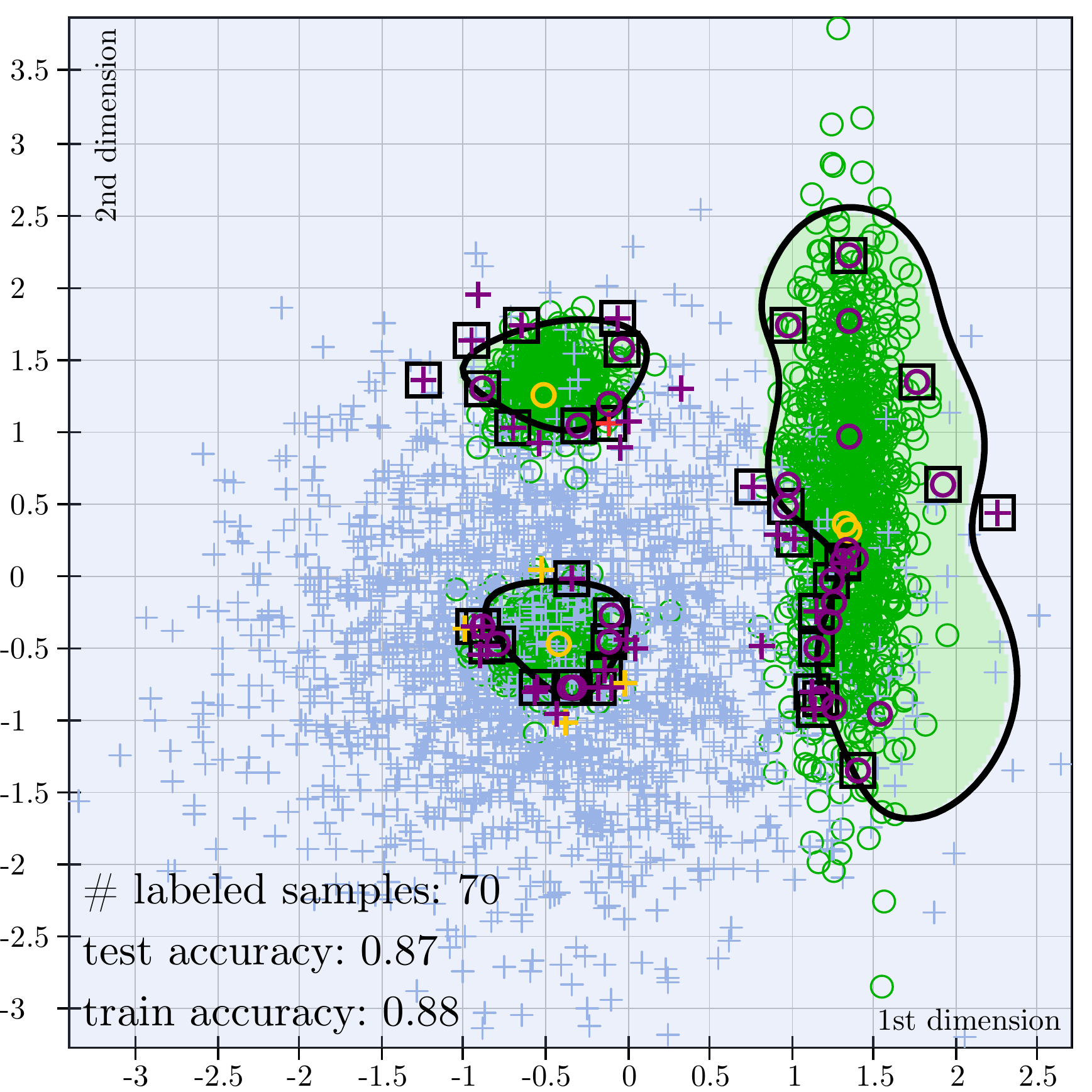}}\quad
\subfigure[iteration $i=112$.]{\label{fig:RBF_cloudsE}\includegraphics[width=0.318\textwidth]{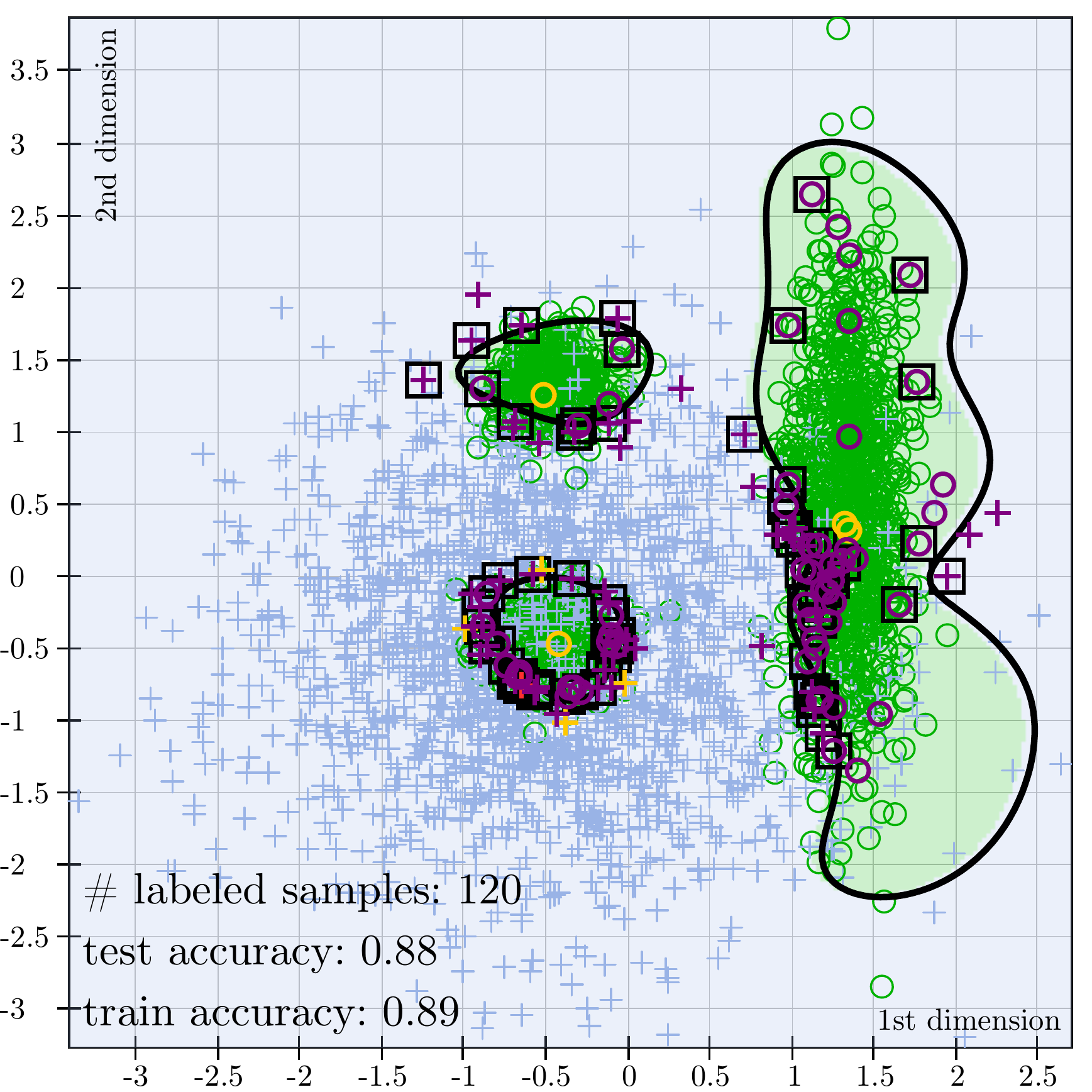}}\quad
\subfigure[classification accuracy vs. learning cycle.]{\label{fig:RBF_cloudsF}\includegraphics[width=0.318\textwidth]{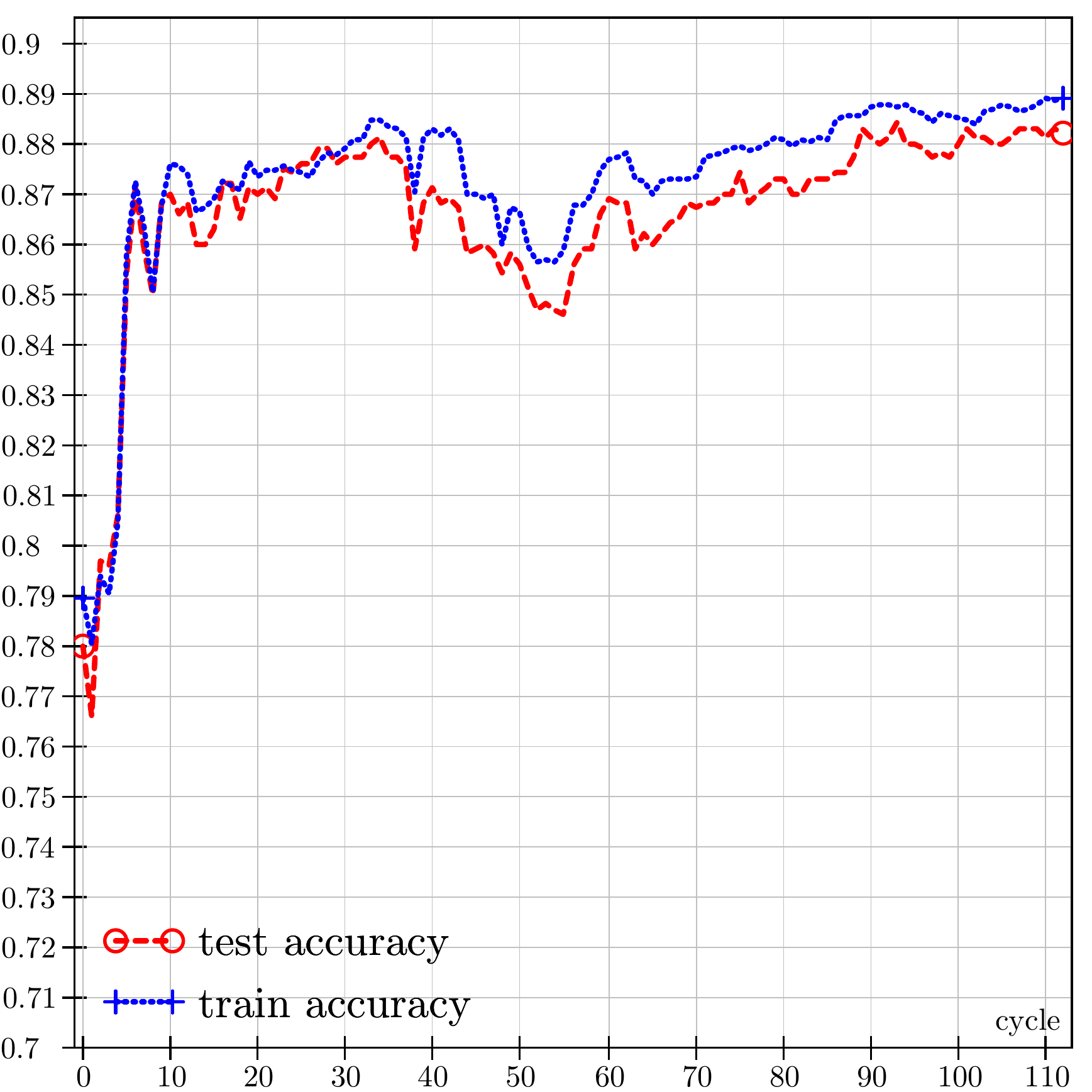}}
\caption{SVM with RBF kernel actively trained with US on the Clouds data set. In each AL round one sample was queried. Eight samples (orange colored) are initially (iteration $i=1$) chosen by a density based strategy. The labeled samples are colored red if they are selected with US in the current query round, and violet otherwise. The samples marked with an rectangle correspond to the support vectors of the SVM.}
\label{fig:RBF_clouds}
\end{center}
\end{figure}

In this section we compare the behavior of an actively trained SVM that recognizes structure information with help of the RWM kernel to that of an SVM with RBF kernel that does not use this information. For visualization purposes, we use a 
two-dimensional, artificially generated data set, called Clouds, which was taken from the UCL Machine Learning Group \cite{UCL14} (for more information see Section \ref{subsec:datasets}). 
On Clouds data set we conducted a $z$-score normalization and split the data set in a $5$-fold cross validation into training and test sets. For a fair comparison we trained SVM with the considered kernels on identically chosen data splits (training and test).
The Figs.~\ref{fig:RWM_clouds} and \ref{fig:RBF_clouds} show the active training process at different iterations only for the first cross-validation fold. Here, each of the parts (a)--(e) display the actively trained SVM after the execution of the initialization round and after the query for $20$, $40$, $70$, and $120$ samples. 
In part (f) of the figures the development of the classification performance regarding the test and training data is pictured, respectively. During the initialization round~($i=1$) of the AL process eight samples are selected with the density based strategy (colored orange). We stop the AL process of the SVM after the selection of $120$ samples. 
As we only want to show the impact of the different kernels on the AL behavior, we applied the selection strategy US with the evaluation measure \textit{smallest distance} (cf.~\cite{TK02}) to select informative samples. For this reason, the SVM is retrained in each query round~$i>1$ after the selection of one sample, that is labeled by an oracle and added to the training set~$\textit{L}_i$.

The parameters of the SVM with RBF kernel and RWM kernel are determined with a second (inner) $4$-fold cross-validation, that only uses the eight labeled samples of the initialization round. Here we applied a grid search, where we varied the penalty parameter~$C=10^{i}$ and the kernel width $\gamma =10^{i}$ with $i \in \{-3,-2,\dots,2\}$. We estimate the density model that the RWM kernel uses for capturing structure information in an unsupervised way with VI, whose hyper-parameters are determined with grid search (on the pool $\textit{U}_0$), too.

Fig.~\ref{fig:RWM_cloudsA} shows that the SVM with RWM kernel yields $89\%$ accuracy on the test data after the execution of the initialization round of our PAL process. For this, less than $1\%$ of the available training data has to be labeled. It can be seen that the mixture model of the RWM kernel (gray colored ellipses) models the ``true'' data distribution quite well. Thus, the resulting decision boundary of SVM with RWM kernel (trained only with the labeled samples in $L_1$) approximates the location of the ``true'' decision boundary between the two classes of the Clouds data set very well.
Fig.~\ref{fig:RBF_cloudsF} illustrates the corresponding learning curves of the SVM with RWM kernel on the training and test data. It can be seen that the SVM yields only slight improvements regarding the classification accuracy on test data in additional query rounds. 
After the selection of $120$ samples (iteration $i=112$) the SVM with RWM kernel reach a test accuracy of nearly $90\%$ which is the same accuracy that a non-actively trained SVM with RBF kernel yields based on $4000$ labeled training samples. 
In addition, the ellipses (with gray colored background) in Figs.~\ref{fig:RWM_cloudsA}--\ref{fig:RWM_cloudsE} show that the mixture model in the $i$-th iteration of the PAL process was not adapted based on the class information of labeled samples $\mathbf{x} \in L_i$ for $i \in \left\{0,1,\dots, 112 \right\}$. 
This is because, on the one hand the processes of the Clouds data set generate normally distributed samples so that the VI algorithm uses the ``right'' number of components to model the four data ``generating'' processes and on the other hand, based on class information that becomes available during the PAL process, no components are recognized as ``disputed'' (see Section~\ref{subsec:refinement}).

In Fig.~\ref{fig:RBF_clouds}, the PAL process of an SVM with RBF kernel is depicted. Already, after the execution of the initialization round, cf.~Fig.~\ref{fig:RBF_cloudsA}, a significant difference between the two kernels can be observed. An SVM with RBF kernel initially yields only an accuracy of about $78\%$ on the test data (the SVM with RWM kernel $89\%$) and the SVM with RBF kernel ``underestimates'' the variance of the process that is located on the right hand side of the two dimensional input space. This is due to the fact that an SVM with RBF kernel (in contrast to an SVM with RWM kernel) compares two given samples with help of the Euclidean distance, and does not use any structure information from the unlabeled samples to assess the similarity between samples. Figs.~\ref{fig:RBF_cloudsB}--\ref{fig:RBF_cloudsE} illustrate the fact, that the SVM with RBF kernels model the ``true'' the decision boundary between the two classes the better, the more actively queried (and labeled) samples are considered for training. Consequently, the classification accuracy of the SVM with RBF kernel increases about significantly during the PAL process, such that the SVM yields an accuracy of $88\%$ on the test data after the active selection of $120$ training samples, cf.~Fig.~\ref{fig:RBF_cloudsF}. However, the SVM with RWM kernel achieves the same test accuracy based on the initial training set consisting of only eight labeled samples (iteration $i=1$). 

This short, preliminary study shows that an SVM that uses structure information with help of a data-dependent kernel (here: the RWM kernel) reaches a superior AL behavior as an SVM (with RBF kernel), that neglects this information. This means that considering structure information an SVM may achieve higher classification accuracies using training sets with a smaller number of labeled samples.

\subsection{Comparison based on $20$ Benchmark Data Sets}
\label{sec:benchmarkDataSets}

To evaluate the benefit of using structure information for active SVM training in more detail, we conduct experiments with $20$ publicly available benchmark data sets. Thus, we are able to come to statistically significant conclusions concerning our new approach. During our AL process, structure information can be considered in different ways: First, this information can only be used by the selection strategy to query the most informative samples in each AL cycle. Second, it can also be taken into account directly for the SVM training (i.e.,~for finding the corresponding support vectors), e.g.,~with data-dependent kernels (RWM, GMM, or LAP kernels). And third, these possibilities to consider structure information can be combined. Consequently, the following questions shall be answered by the our experiments: Does it make any sense to consider structure information for training an SVM actively? And how should this information be integrated into this AL process?  
 
 The following experiments compare the AL performance of SVM with RWM, GMM, RBF, and LAP kernels. For a comprehensive picture, we take into account an SVM with RBF kernel as state-of-the-art method and an SVM with LAP kernel (LapSVM) as best practice approach for an SVM based on a data-dependent kernel. Both paradigms are trained actively with Uncertainty Sampling (US). 
 
 For a visual assessment of the AL behavior of the SVM (with the corresponding kernels) we generate learning curves. These curves outline the classification performance depending  on the number of queries, i.e., iterations of the PAL algorithm (mean accuracy on the five test sets in the cross-validation versus size of $L_{i}$ after each query $i$).
 
 In the following, we compare the performance of the actively trained classifiers with help of the three evaluation criteria RP, DUR, and AULC (cf.~Section~\ref{subsec:criteria}).   
 
 \begin{table}[tbp!]
	\caption{Ranked performance (RP) for the active training of different paradigms. The classification accuracies (on the test data) of SVM with RWM, GMM, LAP, and RBF kernels are given in percentage. For each data set the highest classification accuracies are printed bold.} 
	\label{tab:RP_comparison}
	\begin{center}
		\renewcommand{\arraystretch}{0.8}
		\renewcommand{\tabcolsep}{22.5pt}
		\scriptsize{
			\begin{tabular}{l c c c c c c}
				\toprule
				\multirow{2}{*}{Data Set} & \multicolumn{1}{c}{RWM kernel}  & \multicolumn{1}{c}{GMM kernel}  & \multicolumn{1}{c}{LAP kernel}  & \multicolumn{1}{c}{RBF kernel}  & \multicolumn{1}{c}{RBF kernel} \\
				& \multicolumn{1}{c}{4DS}  & \multicolumn{1}{c}{4DS}  & \multicolumn{1}{c}{US}  & \multicolumn{1}{c}{4DS}  & \multicolumn{1}{c}{US} \\
				\midrule
				Australian & 85.65 & 84.93 & \bf{86.67} & 84.93 & 84.06 \\
				Clouds & \bf{88.92} & 82.82 & 83.96 & 81.08 & 77.20 \\
				Concentric & 99.64 & 99.28 & \bf{99.68} & 99.56 & 99.52 \\
				Credit A & \bf{85.65} & 85.36 & 85.36 & 85.07 & 84.35 \\
				Credit G & 72.40 & 72.00 & \bf{75.50} & 72.00 & 71.20 \\
				Ecoli & 85.15 & 85.44 & 73.90 & \bf{86.64} & 85.73 \\
				Glass & \bf{71.01} & 68.70 & 48.57 & 66.82 & 65.89 \\
				Heart & 84.81 & 84.44 & 77.41 & \bf{85.19} & 82.96 \\
				Iris & \bf{98.00} & 97.33 & 96.00 & \bf{98.00} & 96.67 \\
				Page Blocks & \bf{94.52} & 90.39 & 93.20 & 94.35 & 93.06 \\
				Phoneme & 80.66 & 79.40 & \bf{83.57} & 78.98 & 80.50 \\
				Pima & 75.00 & 75.78 & 75.78 & 75.00 & \bf{76.04} \\
				Ripley & \bf{90.40} & 89.68 & 89.12 & 89.04 & 88.96 \\
				Satimage & \bf{86.33} & 86.09 & 82.19 & 85.30 & 75.39 \\
				Seeds & \bf{97.62} & 95.71 & 90.48 & 92.86 & 91.43 \\
				Two Moons & \bf{100.00} & 97.25 & \bf{100.00} & 95.75 & 95.50 \\
				Vehicle & 76.84 & 79.54 & 69.62 & \bf{81.32} & 80.02 \\
				Vowel & \bf{93.23} & 71.92 & 86.57 & 80.91 & 77.98 \\
				Wine & \bf{98.32} & 97.76 & 97.19 & 97.21 & 97.21 \\
				Yeast & \bf{58.08} & 57.95 & 40.62 & 57.62 & 56.94 \\
				[0.5em]
				Mean & 86.11 & 84.09 & 81.77 & 84.38 & 83.03 \\
				[0.5em]
				Rank & 1.750 & 3.075 & 3.200 & 3.050 & 3.925 \\
				Wins & 11.0 & 0.0 & 4.5 & 3.5 & 1.0 \\
				\bottomrule
			\end{tabular}
		}
	\end{center}
\end{table}

\begin{figure}[htbp!]
\begin{center}
\includegraphics[width=0.6\textwidth]{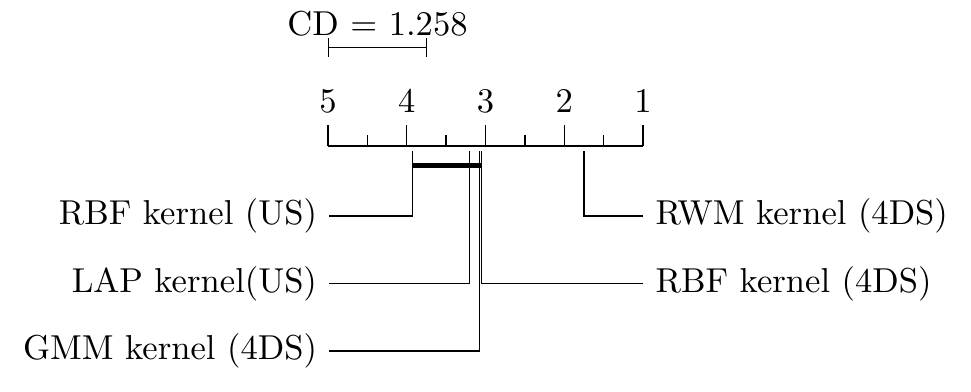}
\caption{CD plot of the evaluation measure RP (with $\alpha=0.1$).}
\label{fig:RP_comparison}
\end{center}
\end{figure}

To compare the PAL processes of the five classifiers statistically, we applied the first evaluation criterion RP. This criterion uses the test accuracy of each classifier $\mathbf{G}_i$, that achieves the highest accuracy on the training data in the $i$-th query round (iteration $i$ of the AL process). The corresponding results for test data are given in Table~\ref{tab:RP_comparison}. Here, the best results (classifiers that the Friedman test assigns the smallest rank numbers) are highlighted in boldface. With five classifiers and 20 data sets, Friedman's $\chi^{2}_{F}$ is distributed according to a $\chi^{2}_{F}$ distribution with $5 - 1$ degrees of freedom. The critical value of $\chi^{2}_{F}(4)$ for $\alpha=0.1$ is $7.779$ and, thus, smaller than Friedman's $\chi^{2}_{F} = 19.73$, so we can reject the null hypothesis.
With the Nemenyi test, we compute the critical difference $\text{CD}= 2.516 \sqrt{\sfrac{5 \cdot 6}{6 \cdot 20}} = 1.258$ to investigate which actively trained classifier performs significantly different. The corresponding CD plot is shown in Fig.~\ref{fig:RP_comparison}. The last two lines in Table~\ref{tab:RP_comparison} show that the SVM with RWM kernel, which was trained actively with 4DS, performs better than all other classifiers on $11$ data sets (wins) and it also receives the smallest averaged rank of $1.750$. The second best rank is achieved by the SVM with RBF kernel, also trained actively with 4DS. However, regarding the number of wins the SVM with LAP kernel (actively trained with US) yields the second best result of $4.5$ wins. A look at the CD plot confirms that the SVM with RWM kernel yields significantly better results (on the test data) than the other classifiers, as only the SVM with RWM kernel is contained in the winner group. Furthermore, the SVM with GMM, RBF, and LAP kernels are located in the same group, i.e., they do not yield significantly different results. 

How large is the fraction of samples that have to be labeled to achieve good classification results? The second evaluation criterion, DUR, tries to answer this question. Table~\ref{tab:DUR_comparison} contains the number of samples and the data utilization ratio needed to achieve a given target accuracy defined by the baseline approach (SVM with RBF kernel actively trained with US). Here, the DUR of an efficient AL process  should be clearly lower than one which is the DUR of the baseline approach. If we consider the last two lines of Table~\ref{tab:DUR_comparison}, it can be seen that only the SVM with data-dependent kernel RWM achieves a mean DUR ($0.947$) smaller than one. The SVM with GMM and RBF kernels (both actively trained with 4DS) achieve worse results regarding the mean DUR (highest) and wins (smallest). The SVM with RBF kernel, trained with US, achieves the second highest number of wins (five) and the second smallest mean DUR ($1.000$).
 
\begin{table}[tbp!]
	\caption{Data utilization ratio (DUR) of the active training of SVM with RWM, GMM, LAP, and RBF kernels. The minimum number of training samples needed to achieve the target accuracy appears in parentheses under the data utilization ratio. The symbol ``$>$'' indicates that the target accuracy (given in $\%$) could not be reached with the maximum number of samples allowed to be labeled actively. The results that receive a win (smallest number of labeled samples) are printed bold.}
	\label{tab:DUR_comparison}
	\begin{center}
		\renewcommand{\arraystretch}{0.8}
		\renewcommand{\tabcolsep}{16.0pt}
		\scriptsize{
			\begin{tabular}{l c c c c c c}
				\toprule
				\multirow{2}{*}{Data Set} & \multicolumn{1}{c}{RWM kernel}  & \multicolumn{1}{c}{GMM kernel}  & \multicolumn{1}{c}{LAP kernel}  & \multicolumn{1}{c}{RBF kernel}  & \multicolumn{1}{c}{RBF kernel} & \multicolumn{1}{c}{\multirow{2}{*}{Target Accuracy}}\\
				& \multicolumn{1}{c}{4DS}  & \multicolumn{1}{c}{4DS}  & \multicolumn{1}{c}{US}  & \multicolumn{1}{c}{4DS}  & \multicolumn{1}{c}{US} &\\
				\midrule
				Australian &  \bf{0.556} & 0.926 & 0.667 & 0.926 & 1.000 & 86.75 \\ 
				&  \bf{(45)} &  {(75)} &  {(54)} &  {(75)} &  {(81)} &  \\ 
				
				Clouds &  \bf{0.055} & 0.738 & 0.293 & 1.299 & 1.000 & 76.19 \\ 
				&   {\bf{(9)}} &  {(121)} &  {(48)} &  {(213)} &  {(164)} &  \\ 
				
				Concentric & 2.472 & 1.782 & 1.070 & 1.577 &  \bf{1.000} & 99.77 \\ 
				&  {(351)} &  {(253)} &  {(152)} &  {(224)} &   {\bf{(142)}} &  \\ 
				
				Credit A & 2.341 & 1.611 & 1.443 & 1.509 &  \bf{1.000} & 90.40 \\ 
				&  {(391)} &  {(269)} &  {(241)} &  {(252)} &   {\bf{(167)}} &  \\ 
				
				Credit G & 1.341 & 1.271 &  \bf{0.839} & 1.271 & 1.000 & 83.26 \\ 
				&  {(401)} &  {(380)} &   {\bf{(251)}} &  {(380)} &  {(299)} &  \\ 
				
				Ecoli & 0.860 & 0.682 & 1.464 &  \bf{0.531} & 1.000 & 90.30 \\ 
				&  {(154)} &  {(122)} &  {(262)} &   {\bf{(95)}} &  {(179)} &  \\ 
				
				Glass &  \bf{0.507} & 0.923 & 1.190 & 0.951 & 1.000 & 76.57 \\ 
				&   {\bf{(72)}} &  {(131)} &  {(169)} &  {(135)} &  {(142)} &  \\ 
				
				Heart &  \bf{0.970} & 1.297 & 1.158 & 1.267 & 1.000 & 89.65 \\ 
				&   {\bf{(98)}} &  {(131)} &  {(117)} &  {(128)} &  {(101)} &  \\ 
				
				Iris & 1.086 & 1.657 & 1.114 & 1.143 &  \bf{1.000} & 98.17 \\ 
				&  {(38)} &  {(58)} &  {(39)} &  {(40)} &   {\bf{(35)}} &  \\ 
				
				Page Blocks &  \bf{0.771} & 1.568 & 0.864 & 0.807 & 1.000 & 92.80 \\ 
				&   {\bf{(216)}} &  {(439)} &  {(242)} &  {(226)} &  {(280)} &  \\ 
				
				Phoneme &  \bf{0.845} & 1.019 & 1.070 & 1.009 & 1.000 & 81.53 \\ 
				&   {\bf{(267)}} &  {(322)} &  {(338)} &  {(319)} &  {(316)} &  \\ 
				
				Pima & 1.460 & 2.012 & 2.075 & 1.820 &  \bf{1.000} & 80.58 \\ 
				&  {(235)} &  {(324)} &  {(334)} &  {(293)} &   {\bf{(161)}} &  \\ 
				
				Ripley &  \bf{0.167} & 0.606 & 1.606 & 0.909 & 1.000 & 88.59 \\ 
				&   {\bf{(11)}} &  {(40)} &  {(106)} &  {(60)} &  {(66)} &  \\ 
				
				Satimage &  \bf{0.147} &  \bf{0.147} & 0.742 & 0.196 & 1.000 & 74.55 \\ 
				&   {\bf{(24)}} &   {\bf{(24)}} &  {(121)} &  {(32)} &  {(163)} &  \\ 
				
				Seeds & 0.541 &  \bf{0.486} & 3.378 & 1.270 & 1.000 & 94.52 \\ 
				&  {(20)} &   {\bf{(18)}} &  {(125)} &  {(47)} &  {(37)} &  \\ 
				
				Two Moons & 0.260 & 4.500 &  \bf{0.180} & 3.440 & 1.000 & 95.69 \\ 
				&  {(13)} &  {(225)} &   {\bf{(9)}} &  {(172)} &  {(50)} &  \\ 
				
				Vehicle &  \bf{0.850} & 1.024 & 1.181 & 0.940 & 1.000 & 90.38 \\ 
				&   {\bf{(357)}} &  {(430)} &  {(496)} &  {(395)} &  {(420)} &  \\ 
				
				Vowel &  \bf{0.335} & 0.537 & 0.532 & 0.757 & 1.000 & 80.04 \\ 
				&   {\bf{(146)}} &  {(234)} &  {(232)} &  {(330)} &  {(436)} &  \\ 
				
				Wine & 2.867 & 1.633 & 1.167 & 1.867 &  \bf{1.000} & 99.86 \\ 
				&  {(86)} &  {(49)} &  {(35)} &  {(56)} &   {\bf{(30)}} &  \\ 
				
				Yeast &  \bf{0.503} & 0.531 & $>$1.748 & 0.517 & 1.000 & 56.81 \\ 
				&   {\bf{(144)}} &  {(152)} &  {($>$500)} &  {(148)} &  {(286)} &  \\[0.5em]
				Mean  & 0.947 & 1.248 & 1.189 & 1.200 & 1.000\\
				Wins  & 10.5 & 1.5 & 2.0 & 1.0 & 5.0\\
				\bottomrule 
			\end{tabular}
		}
	\end{center}
\end{table}

The third criterion, AULC, evaluates the learning speed of the actively trained classifiers. This is done by referring to the baseline approach by subtracting the area under the learning curve of the baseline classifier from the one of a considered classifier. Consequently, an efficient PAL process should always reach a positive AULC. Table~\ref{tab:AULC_comparison} presents the corresponding results. Here, the SVM with RWM kernel, actively trained with 4DS, yields on $12$ of the $20$ data sets a positive AULC and on nine data sets a win. Consequently, the SVM with RWM kernel achieves the highest positive AULC of $2.978$ on average. In addition, the SVM with GMM and RBF kernels (actively trained with 4DS) achieve positive AULC values, too. With respect to the AULC criterion it can be seen that the SVM with LAP kernel (trained with US) yields some very good but also a lot of bad results. Overall, the SVM with LAP kernel yields the worst mean AULC of $-2.094$ despite a number of seven wins.

\begin{table}[tbp!]
	\caption{Area under the learning curve (AULC) of the active training of SVM with RWM, GMM, LAP, and RBF kernels. The results that receive a win  (highest AULC) are printed bold.}
	\label{tab:AULC_comparison}
	\begin{center}
		\renewcommand{\arraystretch}{0.8}
		\renewcommand{\tabcolsep}{22.5pt}
		\scriptsize{
			\begin{tabular}{l c c c c c c}
				\toprule
				\multirow{2}{*}{Data Set} & \multicolumn{1}{c}{RWM kernel}  & \multicolumn{1}{c}{GMM kernel}  & \multicolumn{1}{c}{LAP kernel}  & \multicolumn{1}{c}{RBF kernel}  & \multicolumn{1}{c}{RBF kernel} \\
				& \multicolumn{1}{c}{4DS}  & \multicolumn{1}{c}{4DS}  & \multicolumn{1}{c}{US}  & \multicolumn{1}{c}{4DS}  & \multicolumn{1}{c}{US} \\
				\midrule
				Australian & 1.081 & -0.002 &  \bf{1.851} & -0.002 & 0.000 \\ 
				
				Clouds &  \bf{12.374} & 0.764 & 7.132 & -1.358 & 0.000 \\ 
				
				Concentric & -0.363 & -0.338 &  \bf{0.595} & -0.210 & 0.000 \\ 
				
				Credit A & -2.252 &  \bf{0.179} & -1.203 & -1.178 & 0.000 \\ 
				
				Credit G & -2.337 & -2.337 &  \bf{3.339} & -2.337 & 0.000 \\ 
				
				Ecoli & 0.547 & 0.469 & -7.333 &  \bf{0.915} & 0.000 \\ 
				
				Glass &  \bf{4.164} & -1.196 & -8.635 & 0.174 & 0.000 \\ 
				
				Heart & -0.496 & -0.098 &  \bf{1.779} & -0.243 & 0.000 \\ 
				
				Iris &  \bf{0.102} & 0.003 & 0.047 & 0.017 & 0.000 \\ 
				
				Page Blocks & -3.357 & 0.281 & -25.556 &  \bf{0.944} & 0.000 \\ 
				
				Phoneme & -0.476 & -3.101 &  \bf{1.856} & -1.280 & 0.000 \\ 
				
				Pima & -0.110 & -5.786 &  \bf{0.807} & -0.746 & 0.000 \\ 
				
				Ripley &  \bf{2.183} & 1.061 & -0.489 & 0.004 & 0.000 \\ 
				
				Satimage &  \bf{11.423} & 11.413 & 4.119 & 10.096 & 0.000 \\ 
				
				Seeds & 0.725 &  \bf{0.800} & -3.394 & 0.008 & 0.000 \\ 
				
				Two Moons &  \bf{4.389} & 0.811 & 4.387 & -0.583 & 0.000 \\ 
				
				Vehicle &  \bf{4.785} & 2.642 & -7.687 & 2.616 & 0.000 \\ 
				
				Vowel &  \bf{18.577} & 3.363 & 10.608 & 3.693 & 0.000 \\ 
				
				Wine & -0.216 & -0.077 &  \bf{0.056} & -0.138 & 0.000 \\ 
				
				Yeast &  \bf{8.821} & 6.364 & -24.153 & 6.031 & 0.000 \\[0.5em]
				Mean  & 2.978 & 0.761 & -2.094 & 0.821 & 0.000\\
				Wins  & 9.0 & 2.0 & 7.0 & 2.0 & 0.0\\
				\bottomrule
			\end{tabular}
		}
	\end{center}
\end{table}
 
\begin{figure}[tbp!]
\begin{center}
\subfigure[Concentric data set.]{\includegraphics[width=0.318\textwidth]{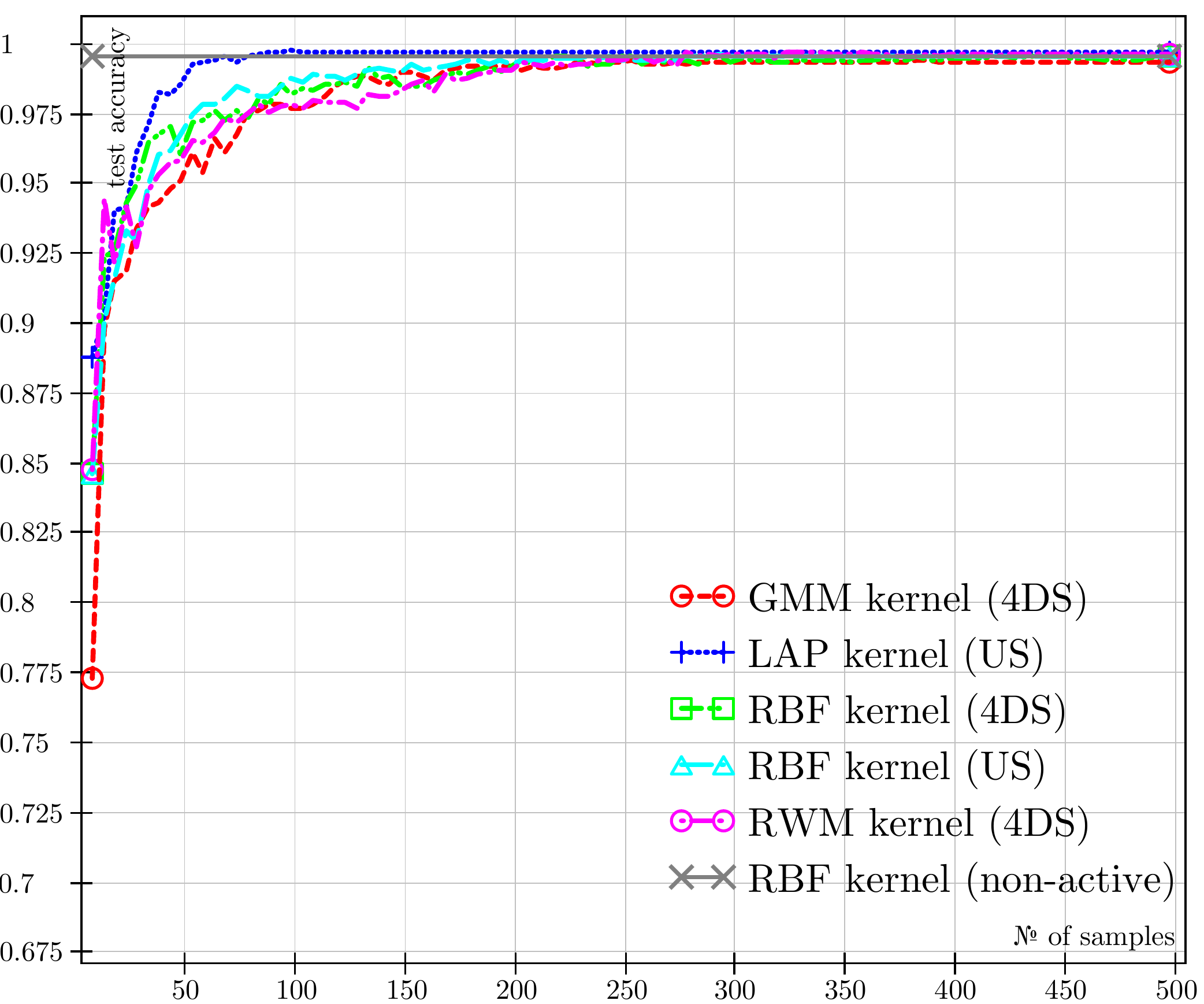}}\quad
\subfigure[Credit A  data set.]{\includegraphics[width=0.318\textwidth]{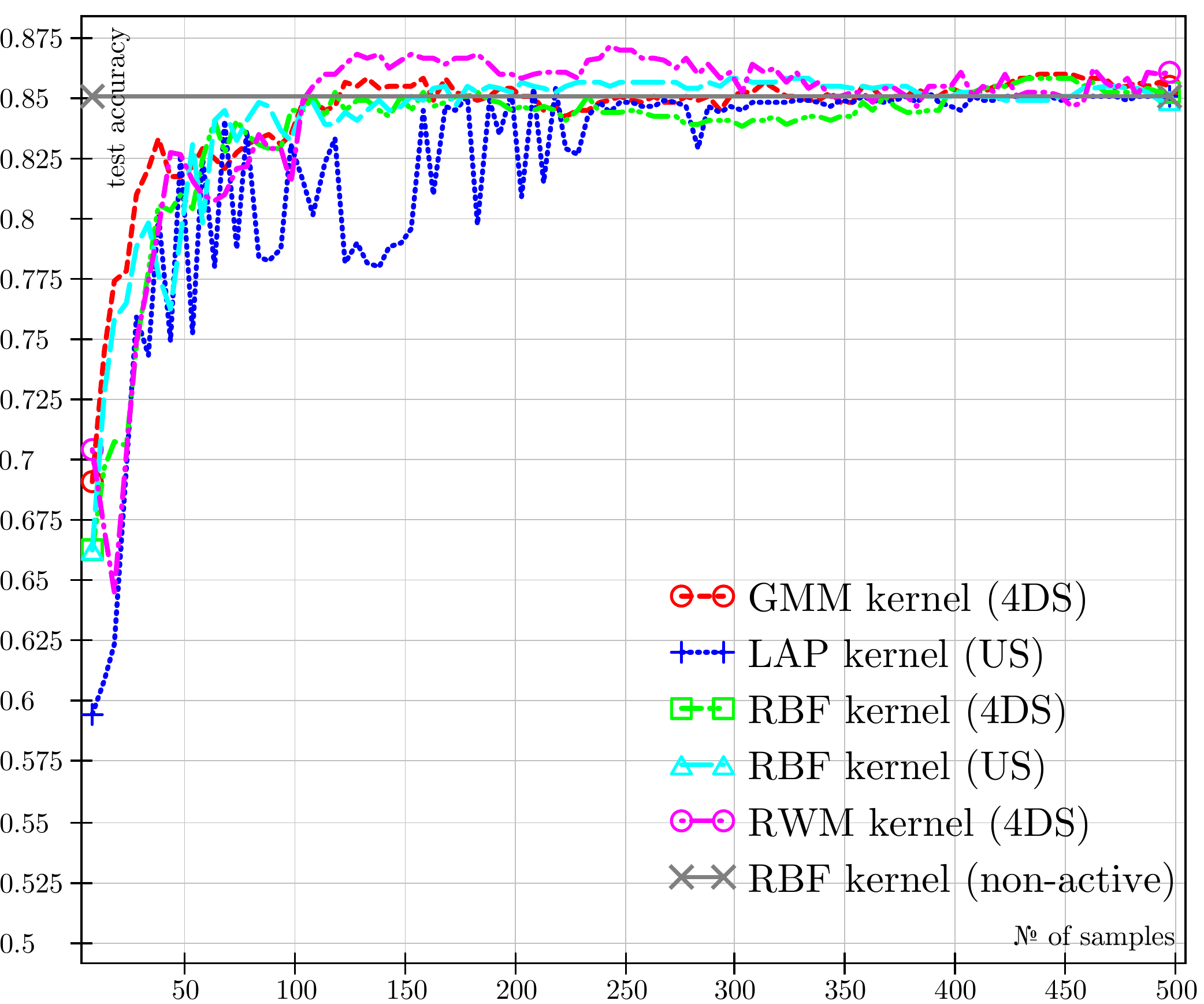}}\quad
\subfigure[Glass data set. ]{\includegraphics[width=0.318\textwidth]{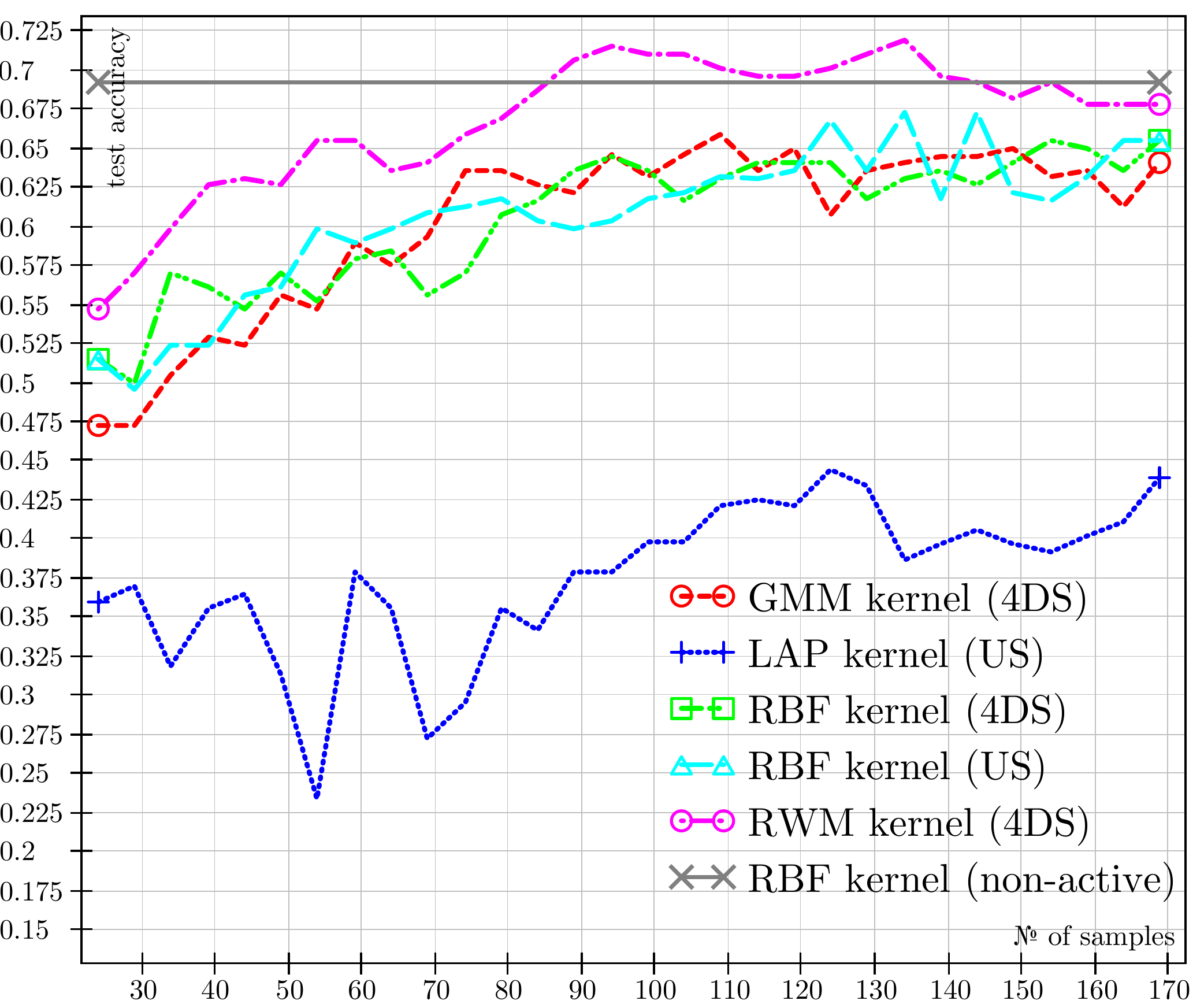}}\vfill
\subfigure[Satimage  data set.]{\includegraphics[width=0.318\textwidth]{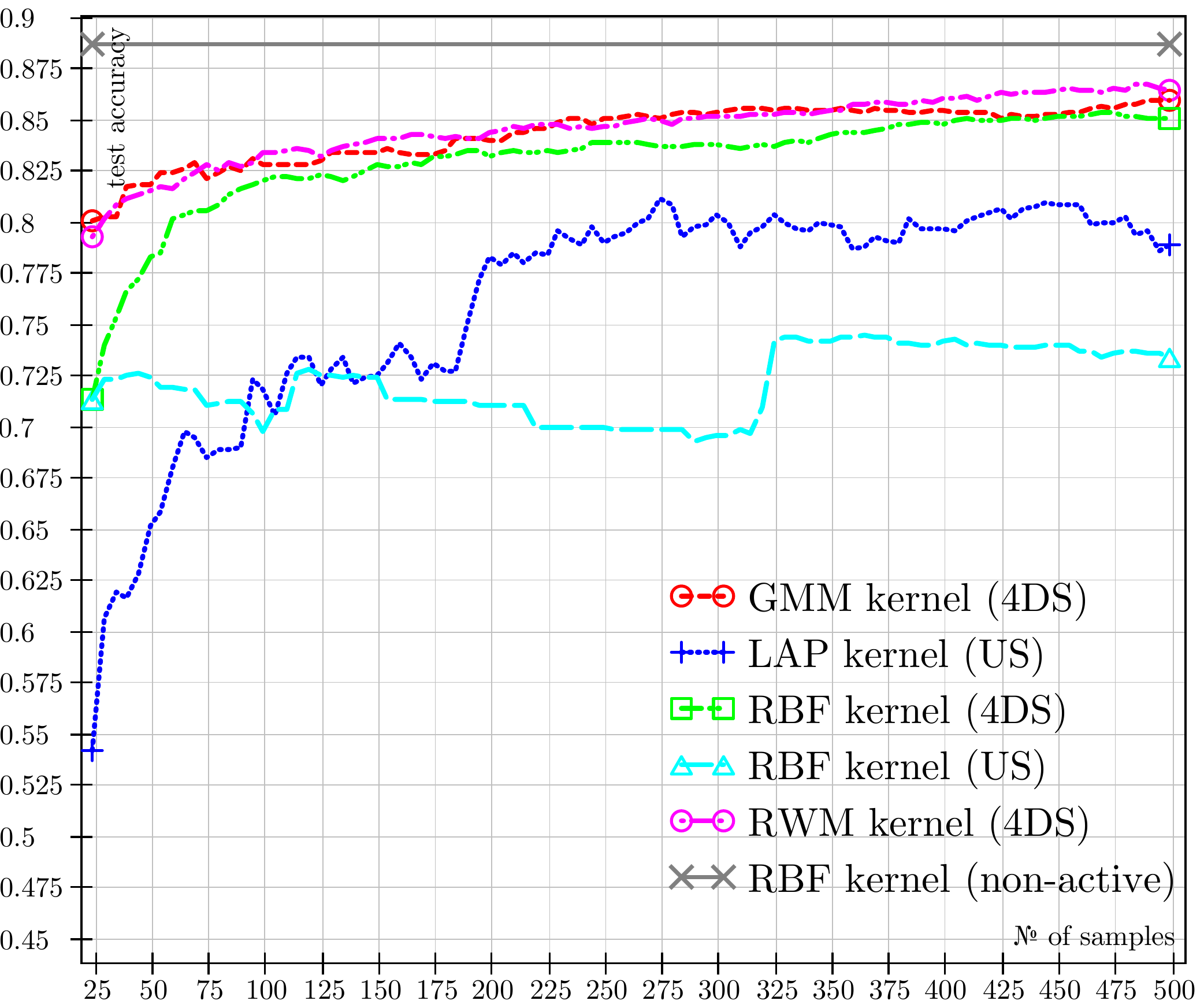}}\quad
\subfigure[Seeds data set.]{\includegraphics[width=0.318\textwidth]{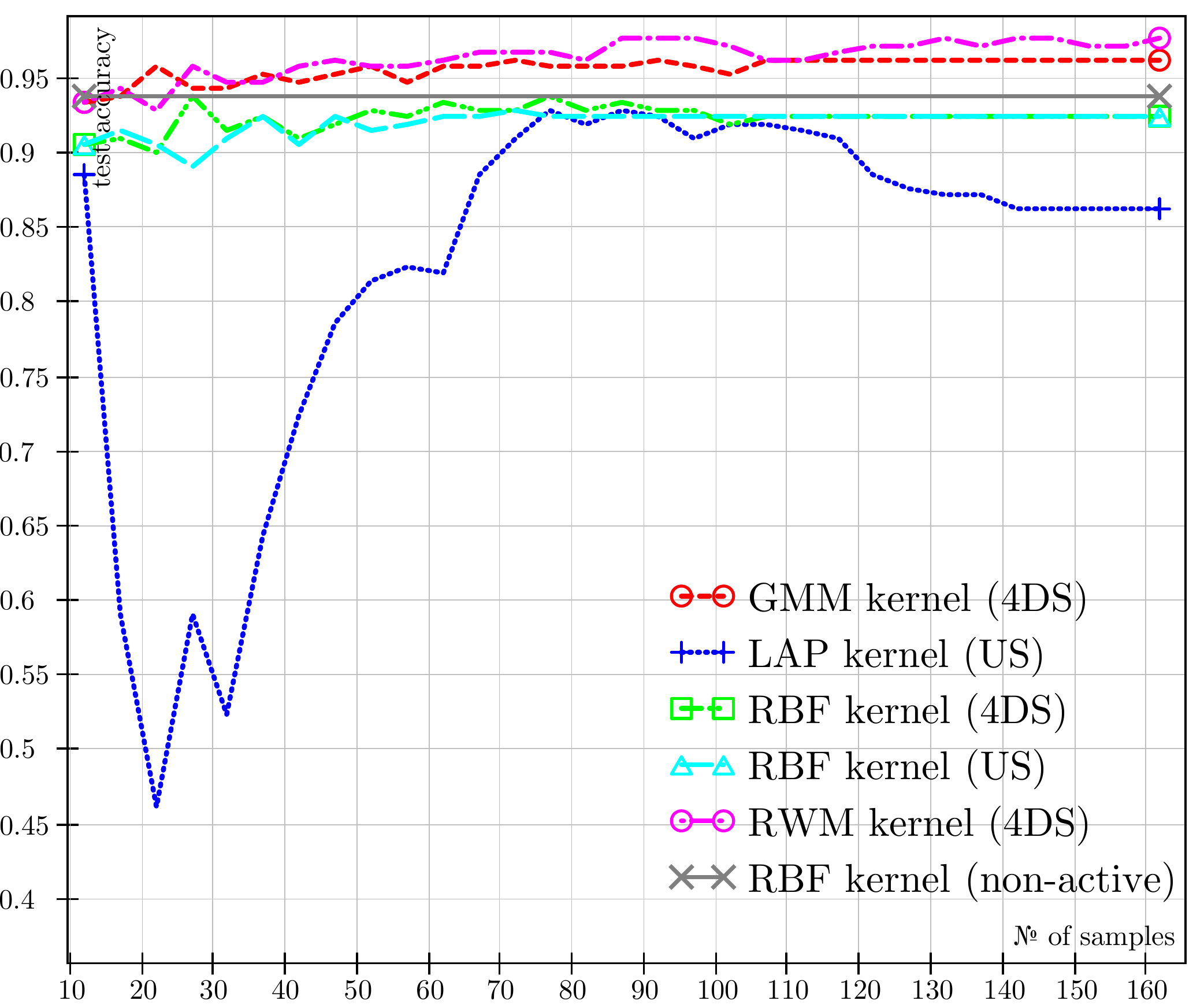}}\quad
\subfigure[Vowel data set.]{\includegraphics[width=0.318\textwidth]{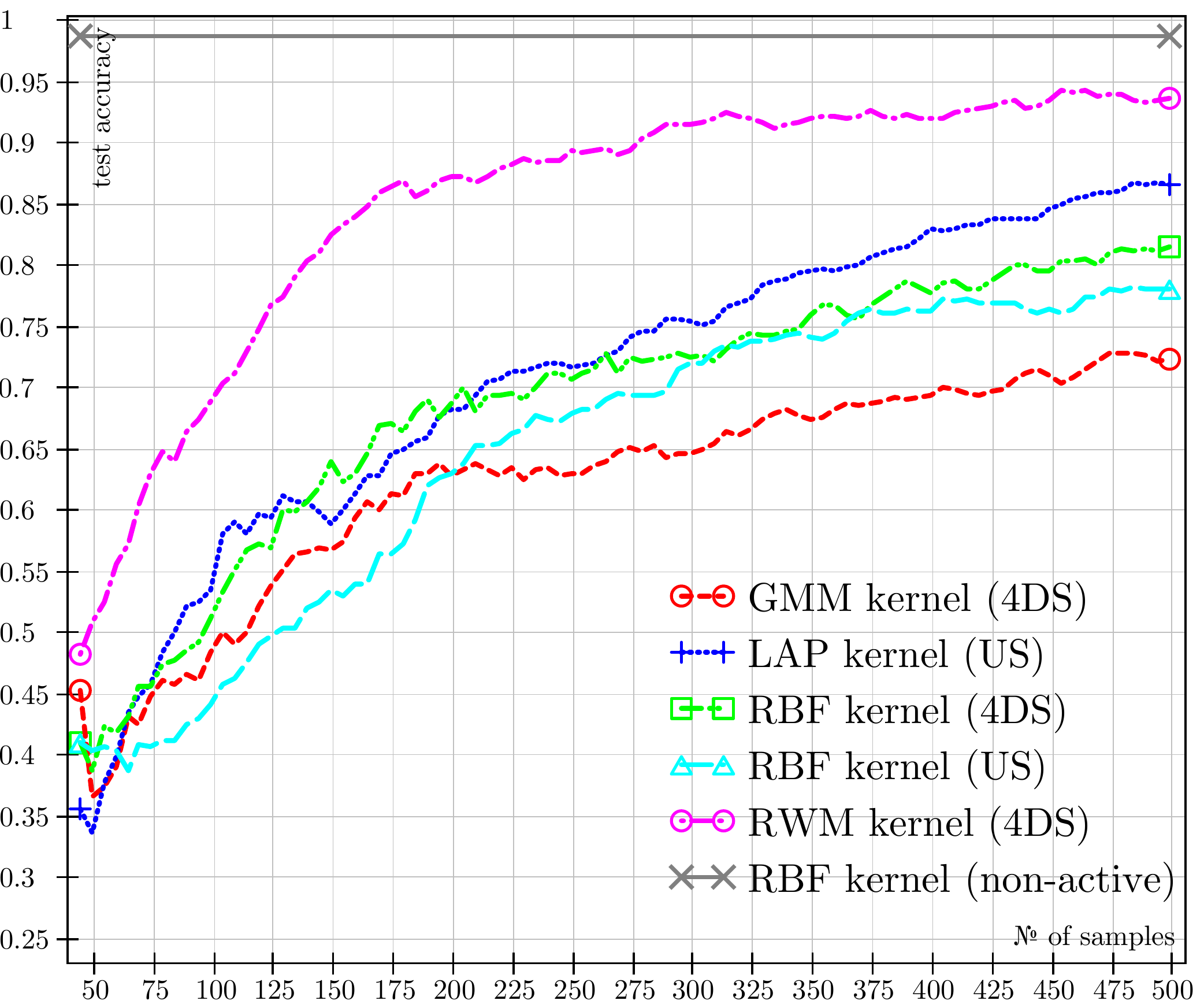}}
\caption{Learning curves (averaged test accuracy over five cross-validation folds versus the number of actively selected samples) of the active training of SVM with RWM, GMM, LAP, RBF kernels for the data sets Concentric, Credit A, Glass, Satimage, Seeds, and Vowel.}
\label{fig:kernel_learning_curves}
\end{center}
\end{figure}

In the following we illustrate the AL behavior of our new approach for six of the data sets: Concentric, Credit~A, Glass, Satimage, Seeds, and Vowel. Fig~\ref{fig:kernel_learning_curves} shows the learning curves (classification performance on test data, averaged over five folds) for the SVM with LAP and RBF kernels (trained with US), and the SVM with RWM, GMM, and RBF kernels, trained with 4DS, respectively. First, we can see that typically the actively trained SVM with data-dependent kernels reach the classification accuracy of non-actively trained SVM with RBF kernel faster than the actively trained SVM with a data-independent kernel (here: RBF kernel). This fact speaks for a synergistic effect between AL and SSL. Second, it can be seen, that the integration of structure information into the PAL process of an SVM with the new data-dependent kernel RWM and the new selection strategy 4DS leads to a rapid and steep increase of the classification accuracy on many of the investigated data sets (cf.~Credit A, Glass, and Vowel). Third, the data sets Concentric and Glass show that active training of the SVM with LAP kernel combined with the selection strategy US sometimes may lead to very bad results. Fourth, there are data sets (e.g.,~Satimage or Vowel) on which actively trained classifiers may yield higher performances if we allow for a active selection of more than 500 samples. 
 
\subsection{Comparison based on the MNIST Data Set}
\label{sec:MNISTDataSet}
To demonstrate the adaptability of our approach the real-world MNIST~\cite{MNIST15} handwritten digit data set is considered.
It is a widely-used benchmark for classification problems in the context of supervised learning, since it incorporates a training set of $60000$ gray-scaled images of handwritten digits from $0$ to $9$ and a test set of additional $10000$ images. 

In this case study we compare the performance of our new approach (SVM with RWM kernel actively trained with 4DS) to that of an SVM with RBF kernel actively trained with US. For reducing the computational and temporal efforts, we captured the structure information only once (before the AL-process starts) in an unsupervised fashion with VI and used only default parameter values for the two different kernels (RBF kernel: $C=1.0$ and $\gamma=0.1$; RWM kernel: $C=1.0$ and $\gamma=0.01$). In addition, we performed a PCA and decided to use the first $34$ principal components, as they cumulatively comprise about $90\%$ of total variance.

In order to assess the results scored by our new AL approach, we first look at some publications that use the MNIST data set, too. To the best of our knowledge, there are only a few publications that attempt to solve this multi-class problem by means of AL. Some of them use only two classes from MNIST to simulate binary classification problems. In~\cite{Beygelzimer2008}, only the images representing the digits $3$ and $5$ are regarded. Additionally, the dimensions are reduced from 784 to 25 by performing a PCA, whereas the size of training and test set are decreased to $1000$ samples each. The same binary classification problem $3$ vs.~$5$ is addressed in~\cite{Ganti2013} and~\cite{Orhan2015}. The latter considers only $2000$ images which are randomly divided into two portions for training and test purposes. In~\cite{Mazzoni2006}, a subset of $1000$ images, $100$ for each digit, are used to solve the binary classification problem $1$ vs.~$7$. Only $200$ images are meaningful, whereas the remaining $800$ are irrelevant, as they are neither $1$ nor $7$. Further, the classification of digit $8$ vs.~all other digits is addressed in~\cite{Bordes2005}. Similarly, subsets of different sizes, $1250$, $2500$, and $5000$ are used in~\cite{Nguyen2004} pursuing the goal of separating the images of a given digit against the other. The classification performance is determined by averaging the sum of the false negatives and false positives over the two classes relative to the size of the subsets.

\begin{figure}[tbp!]
	\begin{center}
		\includegraphics[width=0.85\textwidth]{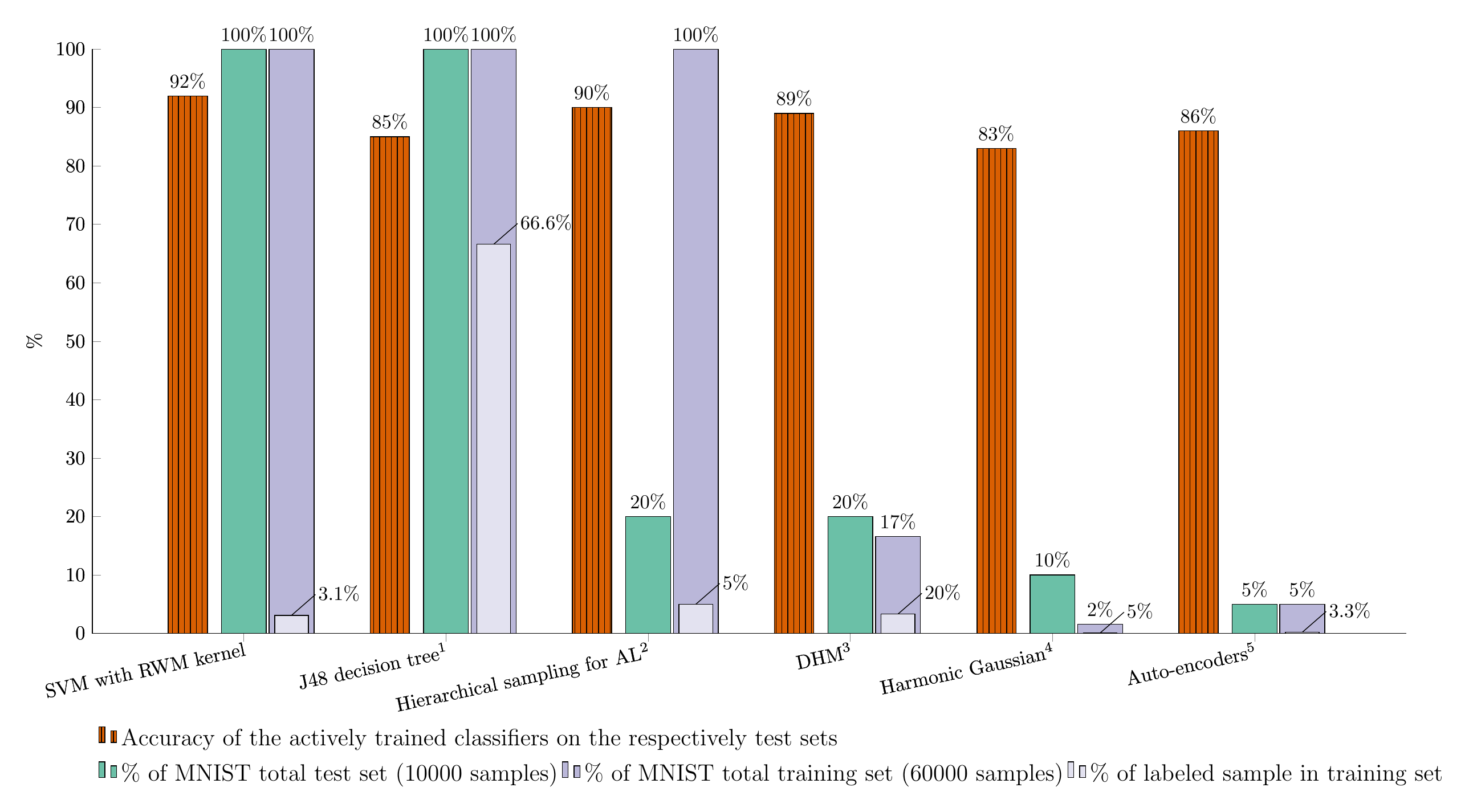}
		\caption{Overview of different AL paradigms using MNIST data set. The accuracy for different techniques on the test set is shown if present, otherwise on the training set. The algorithms are described in detail in the corresponding publications: \textsuperscript{1}\cite{Beygelzimer2011},~\textsuperscript{2}\cite{Dasgupta2008},~\textsuperscript{3}\cite{Dasgupta2011},~\textsuperscript{4}\cite{Ji2012},~\textsuperscript{5}\cite{Lefakis2007}.
		Note: The algorithms are trained and tested on different subsets of the MNIST data set.}
		\label{fig:mnist_comparison}
	\end{center}
\end{figure}

Research has also been conducted with AL paradigms that consider the MNIST data set as a multi-class problem. For example, after actively acquiring labels for $10000$ training images, a classification accuracy of about $90\%$ is reached on $2000$ test images in~\cite{Dasgupta2011}. An analog division of images in training and test data is carried out in~\cite{Dasgupta2008} and an accuracy of about $90\%$ is reached. The first $1000$ images from MNIST training set and the first $1000$ from the MNIST test set are extracted for experimental purposes in~\cite{Ji2012}. The proposed AL method achieves on the test set a classification accuracy of less than $90\%$. In~\cite{Lefakis2007}, a set of $3000$ images are uniformly drawn from the MNIST training set and the proposed technique is tested on $500$ images from the MNIST test set. A classification accuracy of about $85\%$ is reported after acquiring labels for 100 samples.
Experimental results based on the entire MNIST data set are reported in~\cite{Beygelzimer2011} and an overall accuracy of about $85\%$ is reached after labeling $40000$ images. An overview of paradigms that try to solve the multi-class problem with all $10$ classes is presented in Fig.~\ref{fig:mnist_comparison}. The percent of labeled samples is relative to the size of the corresponding training set, e.g.,~in case of Harmonic Gaussian $5\%$ out of $2\%$ of the MNIST total training set has been used to achieve an accuracy of $83\%$ on $20\%$ of the MNIST total test set. Since different subsets of the suggested training set and/or of the test set  are used, it is difficult to compare our paradigm to the results presented in other publications. 
Moreover, the size of the initially labeled set varies strongly (sometimes only learning curves are presented) across the published experimental results, which makes a comparison even harder.

\begin{figure}[tbp!]
	\begin{center}
		\subfigure[$100$ labeled images ($0.2\%$ of training set)]{\includegraphics[width=0.318\textwidth]{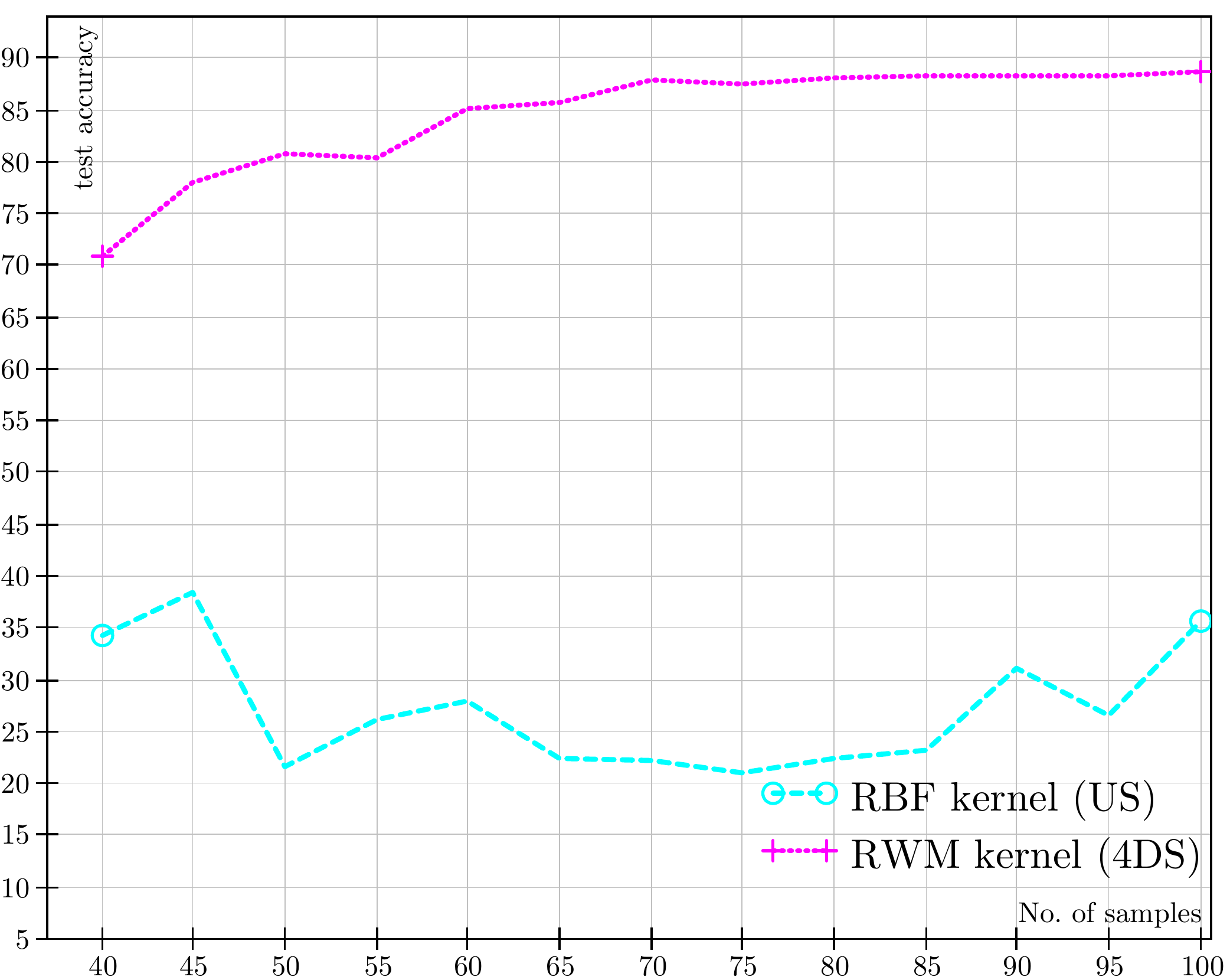}}\quad
		\subfigure[$500$ labeled images ($0.8\%$ of training set)]{\includegraphics[width=0.318\textwidth]{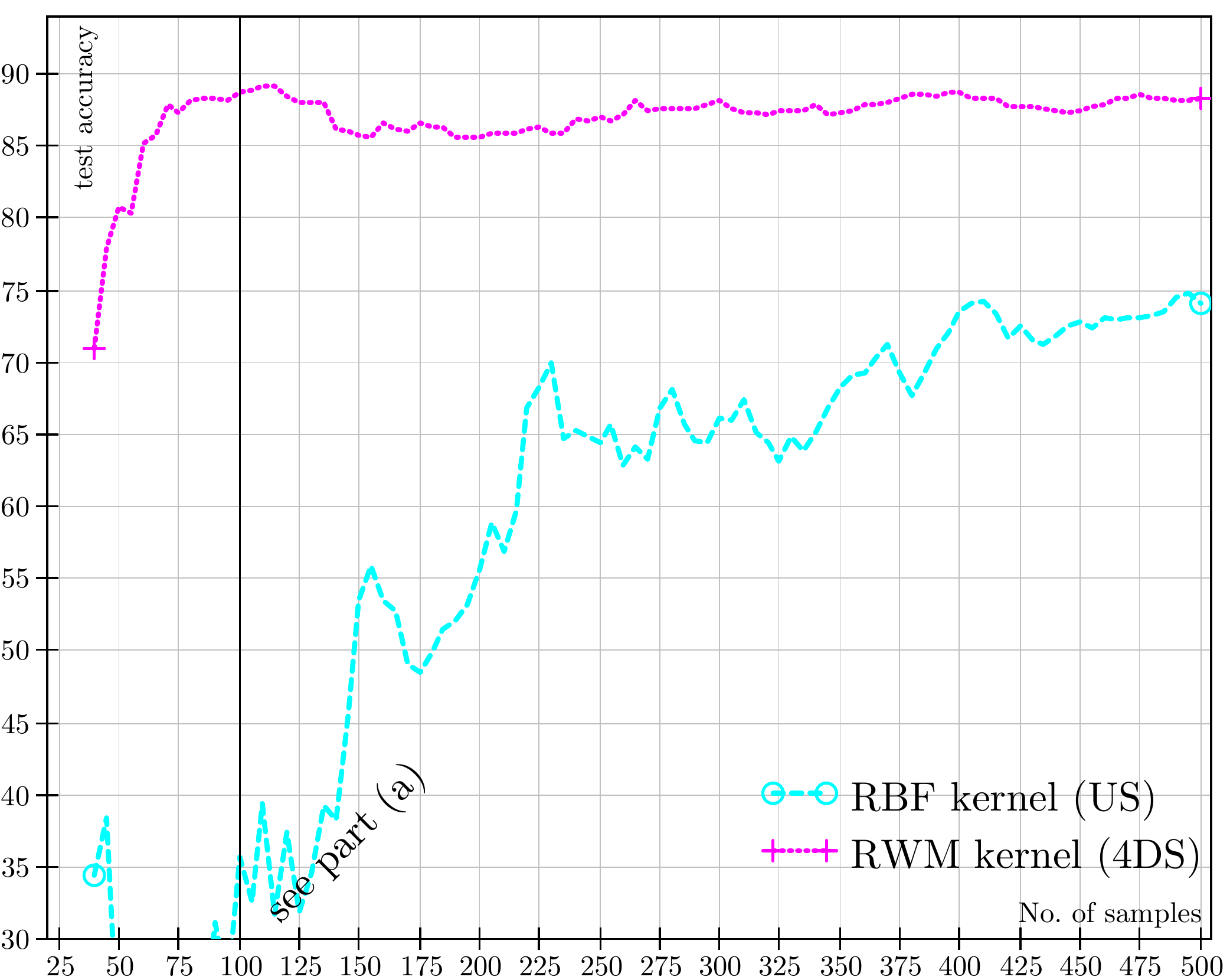}}\quad
		\subfigure[$2500$ labeled images ($4.2\%$ of training set)]{\includegraphics[width=0.318\textwidth]{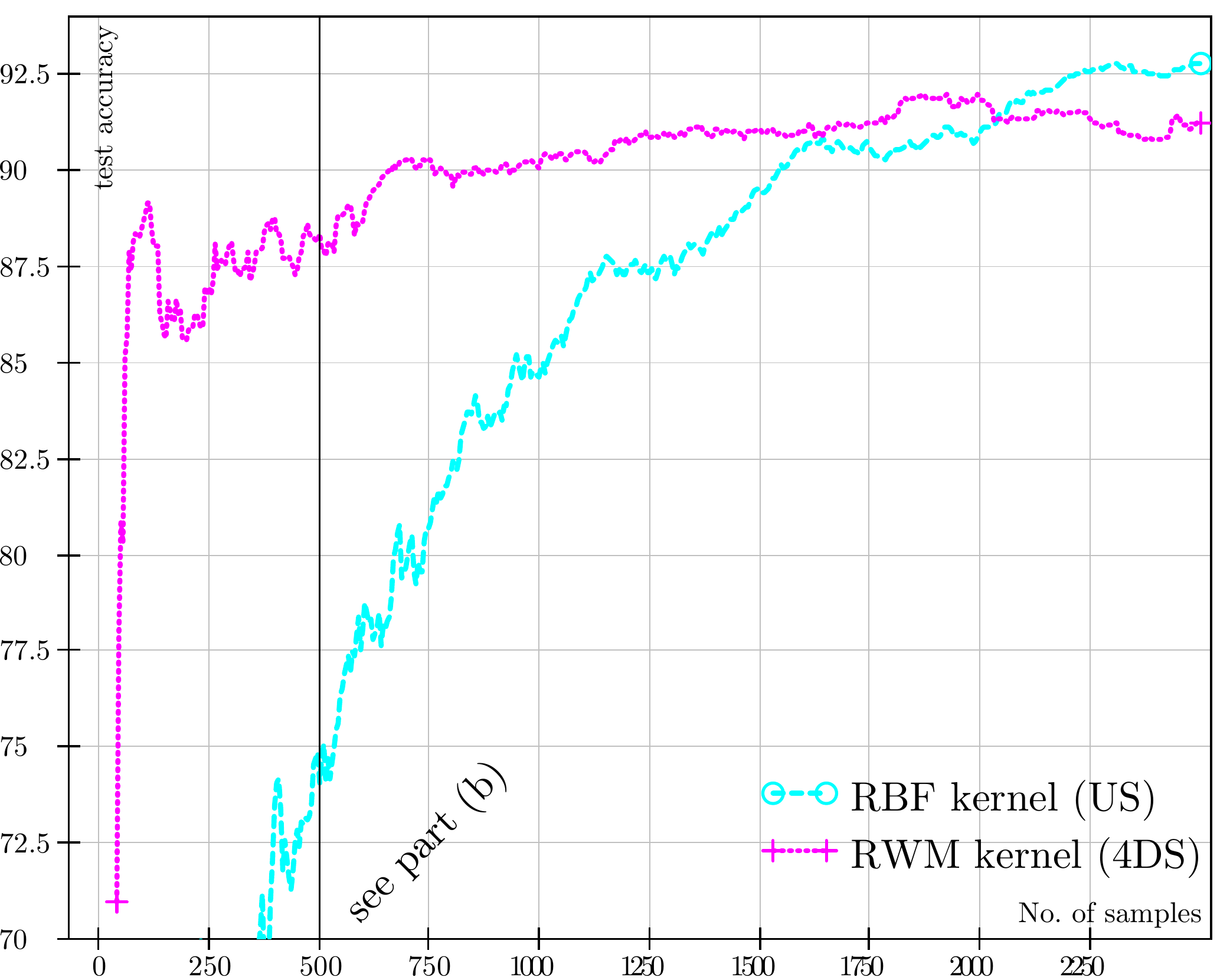}}
		\caption{Learning curves (test accuracy versus the number of actively selected samples) of the active training of SVM with RBF and RWM kernels for the MNIST data set (after 100, 500, and 2500 labeled samples).}
		\label{fig:mnist_learning_curves}
	\end{center}
\end{figure}

The results of the experiments conducted on the MNIST data set are shown in Fig.~\ref{fig:mnist_learning_curves}, as it depicts the performance of the active training of SVM with RBF and RWM kernels on the MNIST test set. We started our AL process without any label information and selected $40$ samples (it represents $0.06\%$ of the training set) by means of the proposed density-based selection strategy (cf.,~Alg.~\ref{alg:densitybasedstrategy}) in the first query round. The labeled samples are the same for both, SVM with RBF and RWM kernels. In case of SVM with RBF kernel and the US selection strategy we select one sample in each query round, whereas in case of SVM with the RWM kernel five samples were chosen. It can easily be observed that the proposed paradigm achieves a steep rise in classification accuracy in the initial phase of the AL and is able to continuously maintain a good performance throughout the whole AL process. Moreover, by regarding only the initial $40$ samples it reaches an accuracy of about $70\%$, which is twice the accuracy of SVM with RBF kernel ($35\%$). When the AL process advances, the SVM with RBF kernel is able to reach the same accuracy as SVM with RWM kernel (after about $1500$ actively selected samples) and performs slightly better at the end.

In summary, it is quite difficult to compare the performance of our approach on the MNIST data set to those in the existing AL literature, due to substantially different partitioning in training set, test set, and initially labeled set, and their respective setup for parameter tuning, as shown in Fig.~\ref{fig:mnist_comparison}. Nevertheless, the behavior exhibited by our approach (see Figs.~\ref{fig:kernel_learning_curves} and~\ref{fig:mnist_learning_curves}) underpins our claim that by capturing and exploiting structure information contained in unlabeled data, the performance of AL can be significantly increased if only a very small number of labeled samples is regarded, which is one of the main goals of AL. 
Moreover, we expect that the results of our new approach can be further improved, if we on the one hand side refine the density model based on the available class information (cf.~Section~\ref{subsec:refinement}) and on the other hand use parameter estimation methods to find good values for the kernel parameters~(cf.~Section~\ref{subsec:classifiers}).

\section{Conclusion and Outlook}
\label{sec:conclusion}

In this article we proposed and evaluated an effective and efficient approach that fuses generative and discriminative modeling techniques for an AL processes. The proposed approach takes advantage of structure information in data which is given by the spatial arrangement of the (un)labeled data in the input space. Consequently, we have shown how structure information captured by means of probabilistic models can be iteratively improved at runtime and exploited to improve the performance of the AL process. Furthermore, we proposed a new density-based selection strategy for selecting the samples in the initial query round.

The key advantages of the proposed paradigm can be summarized as follows:
\begin{itemize}
	\item Due to the new density-based selection strategy, the AL process is able to start without any initially labeled samples.
	\item Better performance (classification accuracy) is reached in an early stage of the active learning, as it exploits the structure information contained in the unlabeled data.
\end{itemize}

In our future work we want to address the following questions:
As the AL process advances and more labeled samples become available, is it possible to find better parameters that will improve the performance? How computationally intensive will it be? Is it feasible and viable to smoothly switch from SVM with RWM kernel to SVM with RBF kernel when enough label information is accessible? Finally, we need to deeply investigate how to parametrize the RWM kernel in case of high dimensional data sets.

\biboptions{sort&compress}
\bibliographystyle{elsart-num-sort}
\bibliography{bib/bibliography,bib/mnistActive}

\begin{thebibliography}{10}
\expandafter\ifx\csname url\endcsname\relax
  \def\url#1{\texttt{#1}}\fi
\expandafter\ifx\csname urlprefix\endcsname\relax\def\urlprefix{URL }\fi

\bibitem{AN07}
A.~Asuncion, D.~Newman, {UCI} {M}achine {L}earning {R}epository,
  \url{http://archive.ics.uci.edu/ml/} (last access 26/02/2016).

\bibitem{BNS06}
M.~Belkin, P.~Niyogi, V.~Sindhwani, Manifold regularization: A geometric
  framework for learning from labeled and unlabeled examples, Journal of
  Machine Learning Research 7 (2006) 2399--2434.

\bibitem{Beygelzimer2008}
A.~Beygelzimer, S.~Dasgupta, J.~Langford, {Importance Weighted Active
  Learning}, in: Proceedings of the 26th Annual International Conference on
  Machine Learning (ICML'09), Montreal, QC, 2009, pp. 49--56.

\bibitem{Beygelzimer2011}
A.~Beygelzimer, D.~Hsu, {Efficient active learning}, in: Proceedings of the
  28th International Conference on Machine Learning (ICML'11), Workshops,
  Bellevue, WA, 2011.

\bibitem{Bis06}
C.~M. Bishop, Pattern Recognition and Machine Learning, Springer, New York, NY,
  2006.

\bibitem{Bordes2005}
A.~Bordes, Å.~Ertekin, J.~Weston, L.~Bottou, {Fast Kernel Classifiers with
  Online and Active Learning}, Journal of Machine Learning Research 6 (2005)
  1579--1619.

\bibitem{Cawlay11}
C.~G. Cawlay, Baseline methods for active learning, in: JMLR: Workshop and
  Conference Proceedings 16, Sardinia, Italy, 2011, pp. 47 -- 57.

\bibitem{libsvm}
C.-C. Chang, C.-J. Lin, {LIBSVM}: A library for support vector machines, ACM
  Transactions on Intelligent Systems and Technology 2 (2011) 27:1--27:27.

\bibitem{CSZ06}
O.~Chapelle, B.~Sch{\"o}lkopf, A.~Zien (eds.), Semi-Supervised Learning, MIT
  Press, Cambridge, MA, 2006.

\bibitem{CSK08}
O.~Chapelle, V.~Sindhwani, S.~S. Keerthi, Optimization techniques for
  semi-supervised support vector machines, Journal of Machine Learning Research
  9 (2008) 203--233.

\bibitem{CM05}
A.~Culotta, A.~McCallum, Reducing labeling effort for structured prediction
  tasks, in: Proceedings of the 20th National Conference on Artificial
  Intelligence (AAAI'05), Pittsburgh, PA, 2005, pp. 746--751.

\bibitem{CKS06}
M.~Culver, D.~Kun, S.~Scott, Active learning to maximize area under the {ROC}
  curve, in: Proceedings of the Sixth International Conference on Data Mining
  (ICDM '06), Hong Kong, China, 2006, pp. 149--158.

\bibitem{CC10}
L.~Cunhe, W.~Chenggang, A new semi-supervised support vector machine learning
  algorithm based on active learning, in: Proceedings of the Second
  International Conference on Future Computer and Communication (ICFCC'10),
  Wuhan, China, 2010, pp. 638 -- 641.

\bibitem{DE95}
I.~Dagan, S.~P. Engelson, Committee-based sampling for training probabilistic
  classifiers, in: Proceedings of the Twelfth International Conference on
  Machine Learning (ICML'95), Katalonien, Spanien, 1995, pp. 150--157.

\bibitem{DRH06}
C.~K. Dagli, S.~Rajaram, T.~S. Huang, Utilizing information theoretic diversity
  for {SVM} active learning, in: Proceedings of the 18th International
  Conference on Pattern Recognition (ICPR'06), Hong Kong, China, 2006, pp.
  506--511.

\bibitem{Dasgupta2011}
S.~Dasgupta, {Two faces of active learning}, Theoretical Computer Science
  412~(19) (2011) 1767--1781.

\bibitem{Dasgupta2008}
S.~Dasgupta, D.~Hsu, {Hierarchical sampling for active learning}, Helsinki,
  Finnland, 2008, pp. 208--215.

\bibitem{Demsar06}
J.~Dem\v{s}ar, Statistical comparisons of classifiers over multiple data sets,
  Journal of Machine Learning Research 7 (2006) 1--30.

\bibitem{DHS01}
R.~O. Duda, P.~E. Hart, D.~G. Stork, Pattern Classification, John Wiley \&
  Sons, Chichester, NY, 2001.

\bibitem{FGQZ11}
M.~Fan, N.Gu, H.~Qiao, B.~Zhang, Sparse regularization for semi-supervised
  classification, Pattern Recognition 44~(8) (2011) 1777--1784.

\bibitem{FKS11}
D.~Fisch, E.~Kalkowski, B.~Sick, S.~J. Ovaska, In your interest - objective
  interestingness measures for a generative classifier, in: Proceedings of the
  third International Conference on Agents and Artificial Intelligence (ICAART
  '11), Rome, Italy, 2011, pp. 414--423.

\bibitem{FS09}
D.~Fisch, B.~Sick, Training of radial basis function classifiers with resilient
  propagation and variational {B}ayesian inference, in: International Joint
  Conference on Neural Networks (IJCNN '09), Atlanta, GA, 2009, pp. 838--847.

\bibitem{FHYA12}
C.~Fook, M.~Hariharan, S.~Yaacob, A.~Adom, A review: Malay speech recognition
  and audio visual speech recognition, in: International Conference on
  Biomedical Engineering, 2012, pp. 479--484.

\bibitem{Friedman40}
M.~Friedman, A comparison of alternative tests of significance for the problem
  of $m$ rankings, The Annals of Mathematical Statistics 11~(1) (1940) 86--92.

\bibitem{Ganti2013}
R.~Ganti, A.~Gray, {Building bridges: Viewing active learning from the
  multi-armed bandit lens}, in: Proceedings of the Twenty-Ninth Conference on
  Uncertainty in Artificial Intelligence (UAI'13), Bellevue, WA, 2013.

\bibitem{GCDL11}
I.~Guyon, G.~Cawley, G.~Dror, V.~Lemaire, Results of the active learning
  challenge, in: Journal of Machine Learning Research: Workshop and Conference
  Proceedings 16, Sardinien, Italien, 2011, pp. 19 -- 45.

\bibitem{HSS03}
A.~Hofmann, C.~Schmitz, B.~Sick, Intrusion detection in computer networks with
  neural and fuzzy classifiers, in: O.~Kaynak, E.~Alpaydin, E.~Oja, L.~Xu
  (eds.), Artificial Neural Networks and Neural Information Processing
  (ICANN/ICONIP), vol. 2714 of LNCS, Springer-Verlag Berlin Heidelberg, 2003,
  pp. 316--324.

\bibitem{HJZL09}
S.~C.~H. Hoi, R.~Jin, J.~Zhu, M.~R. Lyu, Semi-supervised {SVM} batch mode
  active learning with applications to image retrieval, ACM Transactions on
  Information Systems (TOIS) 27~(3) (2009) 1--29.

\bibitem{HND10}
R.~Hu, B.~M. Namee, S.~J. Delany, Off to a good start: Using clustering to
  select the initial training set in active learning, in: Proceedings of the
  23rd International Florida Artificial Intelligence Research Society
  Conference (AAAI'10), Daytona Beach, FL, 2010, pp. 26--31.

\bibitem{JS11}
N.~Japkowicz, M.~Shah, Evaluating Learning Algorithms: A Classification
  Perspective, Cambridge University Press, New York, NY, USA, 2011.

\bibitem{Ji2012}
M.~Ji, J.~Han, {A Variance Minimization Criterion to Active Learning on
  Graphs}, in: Proceedings of the 15th International Conference on Artificial
  Intelligence and Statistics (AISTATS '12), vol.~22, La Palma, Canary Islands,
  2012, pp. 556--564.

\bibitem{Joachims99}
T.~Joachims, Transductive inference for text classification using support
  vector machines, in: Proceedings of the Sixteenth International Conference on
  Machine Learning, ICML '99, San Francisco, CA, 1999, pp. 200--209.

\bibitem{JH09}
J.~Jun, I.~Horace, Active learning with {SVM}, in: J.~Ram\'{o}n, R.~Dopico,
  J.~Dorado, A.~Pazos (eds.), Encyclopedia of Artificial Intelligence, vol.~3,
  IGI Global, Hershey, PA, 2009, pp. 1--7.

\bibitem{KPI14}
J.~Kremer, K.~S. Pedersen, C.~Igel, Active learning with support vector
  machines, Wiley Interdisciplinary Reviews. Data Mining and Knowledge
  Discovery 4~(4) (2014) 313--326.

\bibitem{Lefakis2007}
L.~Lefakis, M.~Wiering, {Semi-Supervised Methods for Handwritten Character
  Recognition using Active Learning}, in: Proceedings of the
  Belgium–Netherlands Conference on Artificial Intelligence, Utrecht,
  Netherlands, 2007, pp. 205--212.

\bibitem{LXQ13}
Y.~Leng, X.~Xu, G.~Qi, Combining active learning and semi-supervised learning
  to construct {SVM} classifier, Knowledge-Based Systems 44~(0) (2013)
  121--131.

\bibitem{LFL12}
C.~Li, C.~Ferng, H.~Lin, Active learning with hinted support vector machine,
  in: Proceedings of the Fourth Asian Conference on Machine Learning (ACML'12),
  Singapur, Singapur, 2012, pp. 221 -- 235.

\bibitem{MME14}
M.~G. Malhat, H.~M. Mousa, A.~B. El-Sisi, Clustering of chemical data sets for
  drug discovery, in: International Conference on Informatics and Systems,
  2014, pp. DEKM--11 -- DEKM--18.

\bibitem{Mazzoni2006}
D.~Mazzoni, K.~L. Wagstaff, M.~C. Burl, {Active Learning with Irrelevant
  Examples}, in: G.~Berlin (ed.), Proceedings of the 17th European Conference
  on Machine Learning (ECML'06), 2006, pp. 695--702.

\bibitem{Melacci12}
S.~Melacci, Manifold regularization: Laplacian svm,
  \url{http://www.dii.unisi.it/~melacci/lapsvmp/} (last access 26/02/2016).

\bibitem{MB11}
S.~Melacci, M.~Belkin, Laplacian support vector machines trained in the primal,
  Journal of Machine Learning Research 12 (2011) 1149--1184.

\bibitem{MNIST15}
MNIST, {MNIST} handwritten digit database,
  \url{http://yann.lecun.com/exdb/mnist/} (last access 26/02/2016).

\bibitem{Nemenyi63}
P.~Nemenyi, Distribution-free Multiple Comparisons, Princeton University,
  Princeton, New Jersey, USA, 1963.

\bibitem{Nguyen2004}
H.~T. Nguyen, a.~Smeulders, {Active learning using pre-clustering}, in:
  Proceedings of the 21st International Conference on Machine Learning
  (ICML'04), Banff, AB, 2004, pp. 623--630.

\bibitem{NCW15}
T.~Ni, F.-L. Chung, S.~Wang, Support vector machine with manifold
  regularization and partially labeling privacy protection, Information
  Sciences 294~(0) (2015) 390--407, innovative Applications of Artificial
  Neural Networks in Engineering.

\bibitem{Orhan2015}
C.~Orhan, {\"{O}}.~Ta\c{s}tan, {ALEVS: Active Learning by Statistical Leverage
  Sampling}, \url{http://arxiv.org/abs/1507.04155} (last access 02/02/2016).

\bibitem{PY10}
S.~J. Pan, Q.~Yang, A survey on transfer learning, IEEE Transactions on
  Knowledge and Data Engineering 22~(10) (2010) 1345--1359.

\bibitem{RSM11}
I.~S. Reddy, S.~Shevade, M.~N. Murty, A fast quasi-newton method for
  semi-supervised {SVM}, Pattern Recognition 44~(10-11) (2011) 2305--2313.

\bibitem{RCS14}
T.~Reitmaier, A.~Calma, B.~Sick, Transductive active learning -- a new
  semi-supervised learning approach based on iteratively refined generative
  models to capture structure in data, Information Sciences 293 (2014)
  275--298.

\bibitem{RS11}
T.~Reitmaier, B.~Sick, Active classifier training with the {3DS} strategy, in:
  IEEE Symposium on Computational Intelligence and Data Mining (CIDM '11),
  France, Paris, 2011, pp. 88--95.

\bibitem{RS13}
T.~Reitmaier, B.~Sick, Let us know your decision: Pool-based active training of
  a generative classifier with the selection strategy {4DS}, Information
  Sciences 230 (2013) 106--131.

\bibitem{RS15}
T.~Reitmaier, B.~Sick, The responsibility weighted {M}ahalanobis kernel for
  semi-supervised training of support vector machines for classification,
  Information Sciences 323 (2015) 179--198.

\bibitem{Ripley96}
B.~Ripley, Pattern recognition and neural networks,
  \url{http://www.stats.ox.ac.uk/pub/PRNN/} (last access 26/02/2016).

\bibitem{SDW01}
T.~Scheffer, C.~Decomain, S.~Wrobel, Active hidden {M}arkov models for
  information extraction, in: Proceedings of the Fourth International
  Conference on Advances in Intelligent Data Analysis (IDA'01), London, UK,
  2001, pp. 309--318.

\bibitem{Scu65}
H.~J. Scudder, Probability of error of some adaptive pattern recognition
  machines, IEEE Transaction on Information Theory (1965) 363--371.

\bibitem{Settles09}
B.~Settles, Active learning literature survey, Computer {S}ciences
  {T}echnischer {B}ericht 1648, University of Wisconsin, Department of Computer
  Science (2009).

\bibitem{Settles11}
B.~Settles, From theories to queries: Active learning in practice, in: Journal
  of Machine Learning Research: Workshop and Conference Proceedings 16,
  Sardinien, Italien, 2011, pp. 1 -- 18.

\bibitem{Sic98}
B.~Sick, Online tool wear monitoring in turning using time-delay neural
  networks, in: Proceedings of the 1998 IEEE International Conference on
  Acoustics, Speech and Signal Processing (ICASSP), vol.~1, 1998, pp. 445--448.

\bibitem{SKC06}
V.~Sindhwani, S.~S. Keerthi, O.~Chapelle, Deterministic annealing for
  semi-supervised kernel machines, in: Proceedings of the 23rd International
  Conference on Machine Learning (ICML'06), Pittsburgh, PA, 2006, pp. 841--848.

\bibitem{SYH11}
M.~Song, H.~Yu, W.-S.~S. Han, Combining active learning and semi-supervised
  learning techniques to extract protein interaction sentences, BMC
  bioinformatics 12~(Suppl 12) (2011) 1--11.

\bibitem{TK02}
S.~Tong, D.~Koller, Support vector machine active learning with applications to
  text classification, Journal of Machine Learning Research 2 (2002) 45--66.

\bibitem{UCL14}
UCL, {UCL/MLG} {E}lena {D}atabase,
  \url{https://www.elen.ucl.ac.be/neural-nets/Research/Projects/ELENA/elena.htm}
  (last access 26/02/2016).

\bibitem{WMZL05}
X.-J. Wang, W.-Y. Ma, L.~Zhang, X.~Li, Multi-graph enabled active learning for
  multimodal web image retrieval, in: Proceedings of the Seventh International
  Workshop on Multimedia Information Retrieval (MIR'05), Singapur, Singapur,
  2005, pp. 65--72.

\bibitem{WYZ11}
Z.~Wang, S.~Yan, C.~Zhang, Active learning with adaptive regularization,
  Pattern Recognition 44~(10--11) (2011) 2375--2383.

\bibitem{WL09}
K.~Q. Weinberger, L.~K. Saul, Distance metric learning for large margin nearest
  neighbor classification, Journal of Machine Learning Research 10 (2009)
  207--244.

\bibitem{XYTXW03}
Z.~Xu, K.~Yu, V.~Tresp, X.~Xu, J.~Wang, Representative sampling for text
  classification using support vector machines, in: Advances in Information
  Retrieval, vol. 2633 of Lecture Notes in Computer Science, Springer, 2003,
  pp. 393--407.

\end{thebibliography}
\end{document}